%% file: main.tex
\documentclass{style/naturep}

\usepackage{amssymb}
\usepackage{amsmath}
\usepackage{graphicx}
\usepackage{algorithmicx}
\usepackage{algorithm}
\usepackage{adjustbox}

\usepackage{graphicx}
\usepackage{caption}
\usepackage[noend]{algpseudocode}
\usepackage{bbm}
\usepackage[margin=0.6in]{geometry}
\usepackage{ragged2e}
\usepackage{setspace}
\usepackage{longtable}
\usepackage{hyperref}
\usepackage{anyfontsize}
\usepackage{setspace}
\usepackage{float}
\usepackage{booktabs}
\usepackage{multirow}
\usepackage{blindtext}
\usepackage{tabularx}
\usepackage{caption}
\usepackage{subcaption}
\usepackage{enumitem, kantlipsum}
\usepackage{xspace}
\usepackage{pifont}
\usepackage{makecell}
\usepackage{graphicx}
\usepackage{amsmath}
\usepackage{amssymb}
\usepackage{booktabs}
\usepackage{float}
\usepackage{placeins}
\usepackage{array}
\usepackage[table,x11names]{xcolor}
\definecolor{maroon}{cmyk}{0,0.87,0.68,0.32}
\definecolor{gray}{rgb}{0.3,0.3,0.3}
\usepackage{lineno}

\newcommand\Heading[1]{
  \noindent\textbf{\Large{#1}}
}

\newcommand\heading[1]{
  \noindent\textbf{\large{#1}}
}

\newcommand\hheading[1]{
  \noindent\textbf{#1}
}

\title{\begin{flushleft}{\begin{spacing}{1}
   \textcolor{black}{A Multimodal Vision Foundation Model for Clinical Dermatology}
\end{spacing}}\end{flushleft}}

\makeatletter
\let\saved@includegraphics\includegraphics
\AtBeginDocument{\let\includegraphics\saved@includegraphics}

\makeatother

\begin{document}
\input{sections/0-cover_page}
\input{sections/0-abstract}
\begin{spacing}{1.35}
\newpage  
\Heading{Introduction}

\input{sections/1-introduction}

\Heading{Results}

\input{sections/2-performance}

\Heading{Discussion}

\input{sections/3-discussion}

\end{spacing}

\begin{spacing}{1.35}
\Heading{Methods}
\input{supplement/0-pretraining}

\input{supplement/1-evaluation}

\input{supplement/2-datasets}

\input{supplement/3-additional}
\end{spacing}


\input{figures/figure}

\clearpage
\setcounter{table}{0}
\renewcommand{\tablename}{Extended Data Table}
\input{tables/table_detail_performance}
\input{tables/model_details}

\clearpage
\begin{nolinenumbers}
\Heading{References}

\begin{spacing}{0.9}
\bibliographystyle{naturemag}
\bibliography{main}
\end{spacing}
\end{nolinenumbers}
\clearpage

\end{document}

%% file: sections/0-cover_page.tex
\maketitle
\vspace{-20mm}
\begin{spacing}{1.4}
\noindent Siyuan Yan$^{1,2}$, Zhen Yu$^{1}$, Clare Primiero$^{3}$, Cristina Vico-Alonso$^{5}$, Zhonghua Wang$^{1}$, Litao Yang$^{1,2}$, Philipp Tschandl$^{4}$, Ming Hu$^{1,2}$, Lie Ju$^{1}$, Gin Tan$^{9}$, Vincent Tang$^{10}$, Aik Beng Ng$^{10}$, David Powell$^{9}$, Paul Bonnington$^{11}$, Simon See$^{10}$, Elisabetta Magnaterra$^{12}$, Peter Ferguson$^{13,14}$, Jennifer Nguyen$^{5}$, Pascale Guitera$^{15}$, Jose Banuls$^{16}$, Monika Janda$^{6}$, Victoria Mar$^{5,8\dagger}$, Harald Kittler$^{4\dagger}$, H. Peter Soyer$^{3,7\dagger}$, Zongyuan Ge$^{1\dagger*}$
\end{spacing}
\vspace{-7mm}
\begin{spacing}{1.4}
\begin{affiliations}
 \item AIM for Health Lab, Faculty of Information Technology, Monash University, Melbourne, Australia  
 \item Faculty of Engineering, Monash University, Melbourne, Australia  
 \item Frazer Institute, The University of Queensland, Dermatology Research Centre, Brisbane, Australia  
 \item Department of Dermatology, Medical University of Vienna, Vienna, Austria  
 \item Victorian Melanoma Service, Alfred Hospital, Melbourne, Australia 
 \item Centre for Health Services Research, Faculty of Medicine, The University of Queensland, Brisbane, Australia
 \item Dermatology Department, Princess Alexandra Hospital, Brisbane, Australia
 \item School of Public Health and Preventive Medicine, Monash University, Melbourne, Australia
 \item eResearch Centre, Monash University, Melbourne, Australia
 \item NVIDIA AI Technology Center, Singapore
 \item The University of Queensland, Brisbane, Australia
 \item Section of Dermatology, Department of Health Sciences, University of Florence, Florence, Italy
 \item Melanoma Institute Australia, North Sydney, New South Wales, Australia
 \item Tissue Pathology and Diagnostic Oncology, Royal Prince Alfred Hospital and NSW Health Pathology, Sydney, New South Wales, Australia
 \item Sydney Melanoma Diagnostic Centre, Royal Prince Alfred Hospital, New South Wales, Australia
 \item Department of Dermatology, Hospital General Universitario de Alicante, ISABIAL, Alicante, Spain
\\$\dagger$These authors contributed equally as senior authors
   \\$^*$Corresponding author: 
 Zongyuan Ge (zongyuan.ge@monash.edu)
 
\end{affiliations}
\end{spacing}

\vspace{-6mm}
\newpage  
\begin{spacing}{1.15}

\noindent \textbf{Diagnosing and treating skin diseases require advanced visual skills across domains and the ability to synthesize information from multiple imaging modalities. While current deep learning models excel at specific tasks like skin cancer diagnosis from dermoscopic images, they struggle to meet the complex, multimodal requirements of clinical practice. Here, we introduce PanDerm, a multimodal dermatology foundation model pretrained through self-supervised learning on over 2 million real-world skin disease images from 11 clinical institutions across 4 imaging modalities. We evaluated PanDerm on 28 diverse benchmarks, including skin cancer screening, risk stratification, differential diagnosis of common and rare skin conditions, lesion segmentation, longitudinal monitoring, and metastasis prediction and prognosis. PanDerm achieved state-of-the-art performance across all evaluated tasks, often outperforming existing models when using only 10\% of labeled data. \textcolor{black}{We conducted three reader studies to assess PanDerm's potential clinical utility. PanDerm outperformed clinicians by 10.2\% in early-stage melanoma detection through longitudinal analysis, improved clinicians' skin cancer diagnostic accuracy by 11\% on dermoscopy images, and enhanced non-dermatologist healthcare providers' differential diagnosis by 16.5\% across 128 skin conditions on clinical photographs. These results demonstrate PanDerm's potential to improve patient care across diverse clinical scenarios and serve as a model for developing multimodal foundation models in other medical specialties, potentially accelerating the integration of AI support in healthcare. The code can be found at https://github.com/SiyuanYan1/PanDerm.}}
\end{spacing}

%% file: sections/1-introduction.tex
\noindent There is a pressing need to fully harness the potential of artificial intelligence (AI) in diagnosing and managing skin diseases. Although deep learning has demonstrated remarkable performance, often matching or surpassing dermatologists, current AI models for dermatology remain limited to isolated tasks, such as diagnosing skin cancer from dermoscopic images \cite{derm3}. These models struggle to integrate various data types and imaging modalities, reducing their utility in \textcolor{black}{different} real-world clinical settings. Dermatology, like internal medicine, is inherently complex, \textcolor{black}{encompassing a broad spectrum of conditions from common dermatoses to life-threatening malignancies, necessitating a comprehensive, patient-centered approach that integrates various clinical workflows.} 

\textcolor{black}{In clinical practice,} diagnosing and treating skin conditions involves a range of tasks, including total body skin cancer detection and risk assessment \cite{tbp1, tbp2, risk1, risk2}, \textcolor{black}{differential diagnosis of hundreds of dermatological conditions such as inflammatory dermatoses and pigmentary disorders} \cite{liu2020deep}, multimodal image analysis \cite{multimodal1,multimodal2}, pathology interpretation \cite{path,path2}, monitoring lesion changes \cite{sequential1,sequential2}, and predicting outcomes \cite{meta1,meta2}. The absence of integrated AI solutions capable of supporting these various workflows currently hampers the practical impact of AI in dermatology. Recent advances in foundation models have emerged as a promising direction to address this challenge \cite{derm_fm, gai}.

Foundation models are large-scale neural networks pretrained on vast, diverse data using self-supervised learning techniques, often leveraging weakly labeled or unlabeled data \cite{ssl1,ssl2,mae}. Built on rich knowledge representations, these models have demonstrated impressive performance across medical fields such as ophthalmology \cite{eyefm}, radiology \cite{radiology_fm}, and pathology \cite{path_fm1,path_fm2,path_fm3,path_fm4}. \textcolor{black}{Through comprehensive pretraining on large and diverse data, these models develop versatile representations that can effectively adapt to various clinical scenarios}, outperforming previous deep learning models in downstream tasks. Their strong feature representations also enable data-efficient applications \cite{fewshot1,fewshot2}, requiring fewer labeled samples, \textcolor{black}{which is particularly crucial for medical domains where expert-annotated data are often limited.}

However, developing effective foundation models for dermatology presents unique challenges. The performance of foundation models is inherently linked to the scale of their parameters and training data \cite{fm2,scale1,scale2}. In general computer vision, foundation models are pretrained on massive datasets like ImageNet \cite{imagenet} or JFT-300M \cite{jft300m} and most existing dermatology AI models still rely on these models for downstream adaptation. Some efforts have focused on self-supervised learning specifically for dermatology using public datasets \cite{autosmim,ham10000} or web-sourced skin images \cite{swavderm}.  However, these approaches are often limited by dataset size, diversity, or the lack of real patient data. \textcolor{black}{Moreover, while recent advances in medical foundation models have shown promise in various specialties, they cannot fully address dermatology's unique needs. Specialty-specific foundation models \cite{path_fm2,radiology_fm,eyefm} typically focus on single imaging modalities, while general biomedical models, despite their broad scope, struggle with domain-specific data scarcity and integrating heterogeneous modalities for comprehensive clinical analysis.}

\begin{figure*}
\centering
\includegraphics[width=\textwidth]{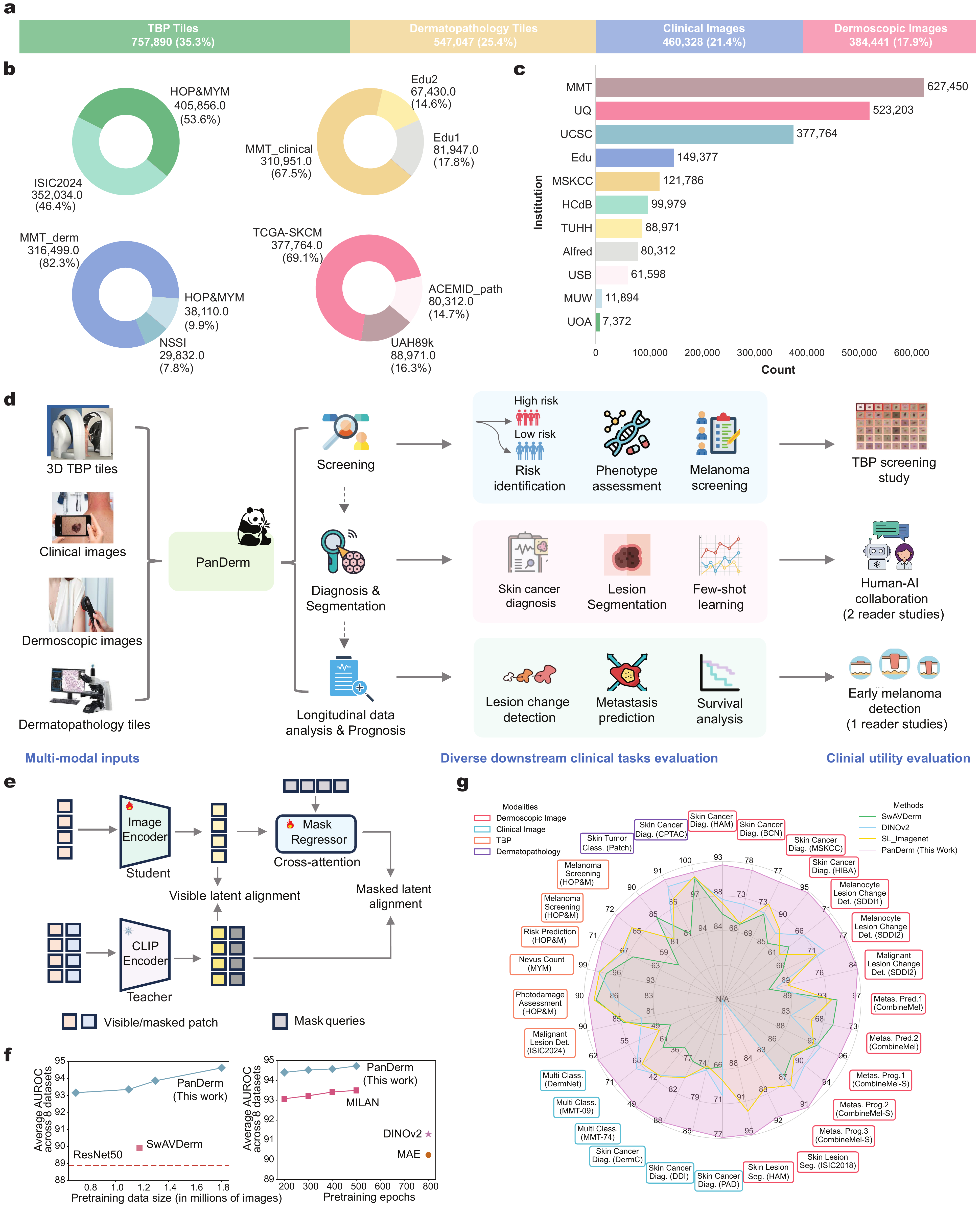}
\caption{\textbf{Overview of this study.} Caption on next page.}
\label{fig1}
\end{figure*}
\begin{figure*}
\caption*{(Previous page.) \textbf{Figure 1: Overview of this study.}
\textbf{a-c.} The pretraining dataset includes 2.1 million images from 11 clinical sources and institutions across 4 modalities. Dataset distribution is shown by modalities (\textbf{a}), data sources (\textbf{b}), and institutions (\textbf{c}). \textbf{d.} PanDerm can interpret various imaging modalities and perform a wide range of dermatology-related clinical tasks. It has been evaluated in a real-world scenario study (TBP-based melanoma screening), and three reader studies. Dermatopathology images: microscopic images of skin biopsy specimens. Clinical images: wide-field images capturing the lesion and surrounding skin. Dermoscopic images: close-up images taken with a dermoscope, showing detailed lesion structures. TBP tiles: lesion crops from macro TBP images. \textbf{e.} PanDerm features a decoupled architecture comprising a ViT-large \cite{vit} encoder, regressor, and CLIP-based teacher model, with pretraining using representation reconstruction and CLIP latent alignment objectives. \textbf{f.} PanDerm's performance across varying pretraining data sizes and epochs, measured by average AUROC on 8 benchmark datasets, with comparative analysis against other strategies. \textbf{g.} PanDerm surpasses existing pretrained models on 28 evaluation datasets across 4 modalities.}
\end{figure*}

Here, we introduce PanDerm, a general-purpose, multimodal dermatology foundation model. \textcolor{black}{Uniquely designed to integrate multiple imaging modalities,} PanDerm is pretrained on over 2 million images sourced from 11 institutions across multiple countries, covering 4 imaging modalities \textcolor{black}{spanning diverse dermatological conditions} (\textbf{Fig~\ref{fig1}a-c}). In the pretraining stage, PanDerm employs \textcolor{black}{the masked latent modeling and} CLIP \cite{clip} feature alignment for self-supervised learning (\textbf{Fig~\ref{fig1}e and Methods}), demonstrating superior data scalability and training efficiency compared to existing self-supervised algorithms (\textbf{Fig~\ref{fig1}f}). \textcolor{black}{The model achieves unified representation learning across total body photography (TBP), clinical, dermoscopic, and dermatopathology images, enabling comprehensive patient analysis throughout diverse clinical workflows (\textbf{Fig~\ref{fig1}d}).}

We systematically evaluate PanDerm across 28 benchmarks (\textbf{Fig~\ref{fig1}g}), covering a diverse array of clinical tasks, including screening, risk stratification, phenotype assessment, naevus counting, longitudinal monitoring, lesion change detection, diagnosis of \textcolor{black}{both common and rare skin conditions}, skin lesion segmentation, as well as recurrence prediction and prognosis. PanDerm achieves state-of-the-art performance on all tasks, often using only 5-10\% of the labeled training data typically required. \textcolor{black}{Through three reader studies, we demonstrate that this unified multimodal approach outperforms clinicians in early-stage melanoma detection, enhances clinicians' diagnostic accuracy in skin cancer diagnosis, and supports non-specialist healthcare providers in the differential diagnosis of diverse skin conditions. These findings highlight the potential of specialty-specific foundation models to advance medical practice by integrating diverse modalities, with broader implications for AI development across healthcare specialties.}

%% file: sections/2-performance.tex
\heading{Ablation studies and comparison with other \textcolor{black}{training} strategies}
\begin{figure*}
\centering
\vspace{-12mm}
\includegraphics[width=\textwidth]
{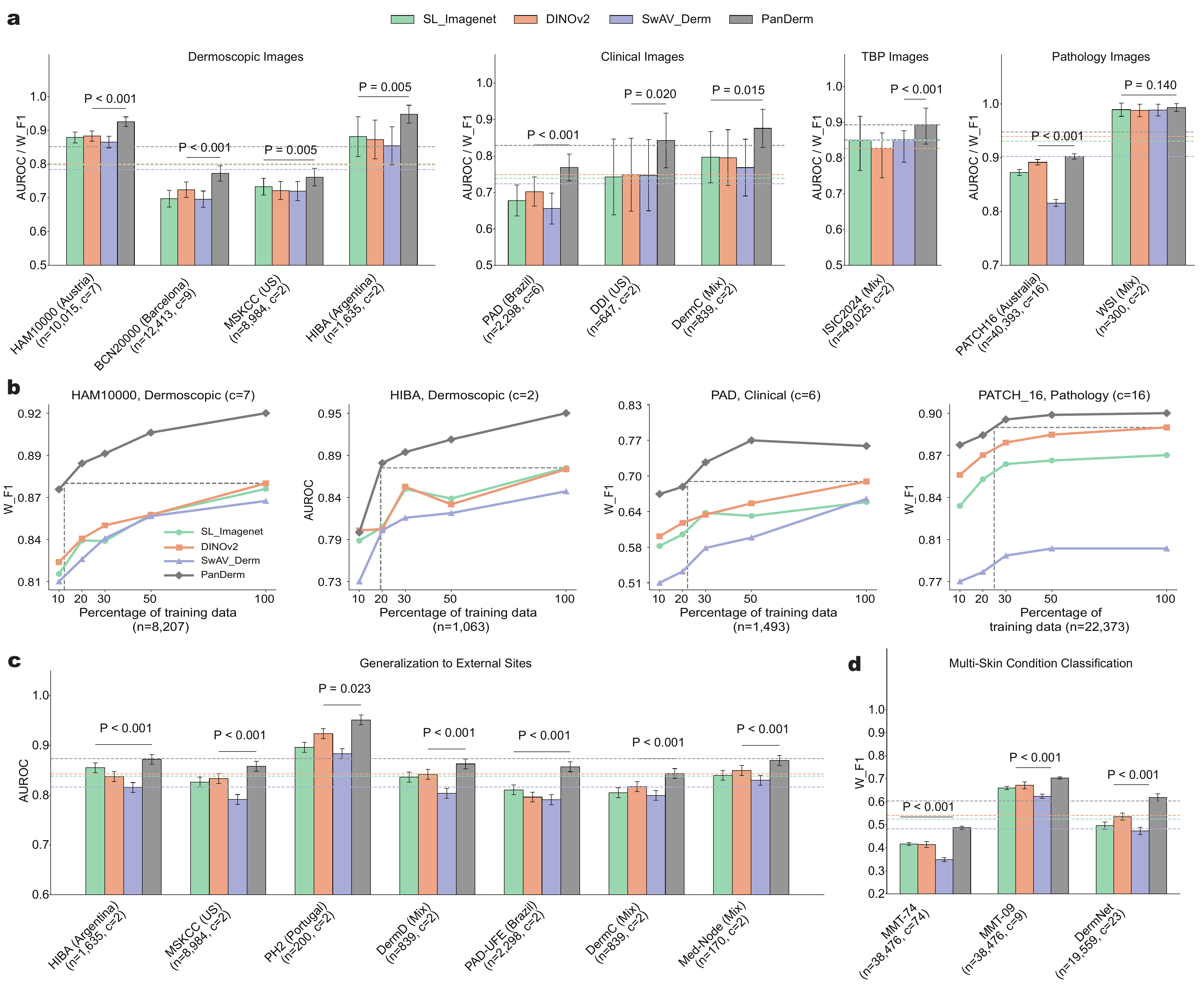}
\caption{\textbf{PanDerm's versatile capacity in diverse diagnosis tasks.} \textbf{a.} Performance comparison of PanDerm versus other pretrained models on 10 pigmented skin lesion datasets across multiple centers and modalities. n: data size, c: class number. Metrics: AUROC for binary class (c=2) and W\_F1 score for multi-class (c$>$2) datasets. Dashed lines indicate average model performance across datasets.
\textbf{b.} Comparison between PanDerm and other pretrained models in label efficiency generalization on four representative datasets, showing performance at various training data percentages. Vertical dash lines indicate the data quantity needed for PanDerm to match existing model performance.
\textbf{c.} External validation for melanoma diagnosis across 7 datasets. \textbf{d.} Performance evaluation of general skin condition classification (up to 74 classes) using clinical images. Error bars in \textbf{a, c, d} show 95\% CIs; bar centers in \textbf{a, c, d} represent mean value; dots in \textbf{b} represent mean value. Estimates were computed using nonparametric bootstrapping with 1000 bootstrap replicates. \textit{P}-values calculated using a two-sided t-test.}
\label{fig2}
\end{figure*}

To evaluate PanDerm's effectiveness, we conducted systematic analyses examining how model performance scales with training data and computational resources (datasets described in \textbf{Extended Data Table~\ref{tab:target_rep}}). First, compared to existing dermatology-specific models, PanDerm showed strong scalability as training data increased from 0.8 to 1.8 million skin images (\textbf{Fig~\ref{fig1}f left}). Notably, it achieved superior performance to SwAVDerm \cite{swavderm}, a leading dermatology self-supervised learning model, using 33\% less training data. When compared to other self-supervised training techniques, PanDerm demonstrated remarkable computational efficiency, requiring only 200 training epochs to achieve the best performance, compared to 500-800 epochs needed by leading methods like MILAN \cite{milan}, DINOv2 \cite{dinov2} and MAE \cite{mae} (\textbf{Fig~\ref{fig1}f right}). Furthermore, PanDerm also surpassed vision-language models such as CLIP \cite{clip}, MONET \cite{monet}, and BiomedCLIP \cite{biomedclip} in benchmark evaluations (\textbf{Extended Data Table~\ref{tab:target_rep}}), while showing emergent capabilities in dermatology similar to those of DINOv2 in natural images, with linear probing performance comparable to full-parameter fine-tuning (\textbf{Extended Data Table~\ref{tab:lp_ft}}). \textcolor{black}{When evaluated against generalist biomedical foundation models, PanDerm demonstrated substantial advantages across different dermatological tasks. Compared to a representative model in this category, BiomedGPT \cite{biomedgpt}, PanDerm showed 20.9\% better AUROC in melanoma detection, 34.7\% higher Weighted F1 in differentiating between skin conditions, and 19.6\% improved weighted F1 in analyzing microscopic skin tissue images (\textbf{Extended Data Table~\ref{tab:GMAI}}). Even using computationally efficient methods, PanDerm maintained its advantages, outperforming both linear-probe and fine-tune versions of BiomedGPT by 14.3\% and 5.1\% respectively in linear probing (\textbf{Extended Data Table~\ref{tab:GAMI2}}).}
Based on these promising results, we expanded our evaluation to compare PanDerm with three representative AI models: SL-Imagenet\cite{imagenet} and DINOv2\cite{dinov2} (both widely-used foundation models pretrained on natural images with a ViT-large \cite{vit} backbone), and SwAVDerm\cite{swavderm} (a self-supervised model pretrained on a large skin image dataset from search engines).

\heading{Diagnostic performance and generalization ability over various datasets}

\noindent
We systematically evaluated PanDerm diagnostic performance across 10 public datasets from 4 imaging modalities and 7 international sites (\textbf{Fig~\ref{fig2}a}). These datasets covered multi-class classification of pigmented neoplastic lesions and binary melanoma diagnosis tasks. PanDerm consistently outperformed all other models, achieving significant improvements on 9 out of 10 datasets, with average gains of 5.1\%, 8.0\%, 4.2\%, and 0.9\% on dermoscopic, clinical, TBP, and pathology datasets, respectively (\textbf{Fig~\ref{fig2}a}). On representative dermoscopy and clinical benchmarks like HAM10000 \cite{ham10000} and PAD \cite{pad}, PanDerm surpassed the next-best models by 4.7\% (\textit{P} $<$ 0.001) and 9.0\% (\textit{P} $<$ 0.001), respectively (\textbf{Fig~\ref{fig2}a}; \textbf{Extended Data Table~\ref{tab1}} and \textbf{Extended Data Fig~\ref{supp_fig3}}). 

PanDerm demonstrated strong performance even with limited training data, achieving comparable results to other models while using only 10\% to 30\% of the labeled images (\textbf{Fig~\ref{fig2}b, Extended Data Table~\ref{tab15}-\ref{tab20}}). Additional results for other tasks are presented in \textbf{Extended Data Fig~\ref{supp_fig4}}. To test PanDerm's generalization applicability, we evaluated its performance on melanoma diagnosis using images from 7 external medical centers, representing patient populations different from the training data. PanDerm demonstrated significant superiority over all pretrained models, achieving higher AUROC scores across all external datasets (\textbf{Fig~\ref{fig2}c}). Notably, it maintained high performance even on clinical photographs that were not used during training, with AUROC gains of 4.0\%, 2.6\%, and 2.1\% on the three external clinical datasets (All \textit{P} $<$ 0.001).

\textcolor{black}{Beyond skin cancer diagnosis, we evaluated PanDerm's ability to diagnose a broader range of skin conditions commonly seen in clinical practice. We tested on three complementary datasets: the public DermNet dataset \cite{Dermnet} covering 23 common conditions and two internal datasets - MMT-09 (9 conditions) and MMT-74 (74 conditions). These datasets comprehensively cover inflammatory diseases, infections, various types of skin tumors, and other frequently encountered skin problems.} As shown in \textbf{Fig~\ref{fig2}d}, PanDerm achieved weighted F1 improvements of 3.2\%, 7.1\%, and 8.2\% on MMT-09, DermNet, and MMT-74, respectively, compared to the next-best models (all \textit{P} $<$ 0.001). PanDerm's advantage grew larger as the number of conditions increased, demonstrating its strong capability to handle complex, multi-disease scenarios. PanDerm also outperformed all other pretrained models on all metrics across the three datasets (all \textit{P} $<$ 0.001; \textbf{Extended Data Table~\ref{tab2}}). In the DermNet dataset, PanDerm exceeded the next-best model's area under the precision-recall curve (AUPR) by 14.7\%. Further details on the experimental setup, datasets, and metrics are provided in \textbf{Methods}.

\heading{Lesion monitoring and change detection in sequential images}

\begin{figure*}
\centering
\vspace{-5mm}
\includegraphics[width=\textwidth]{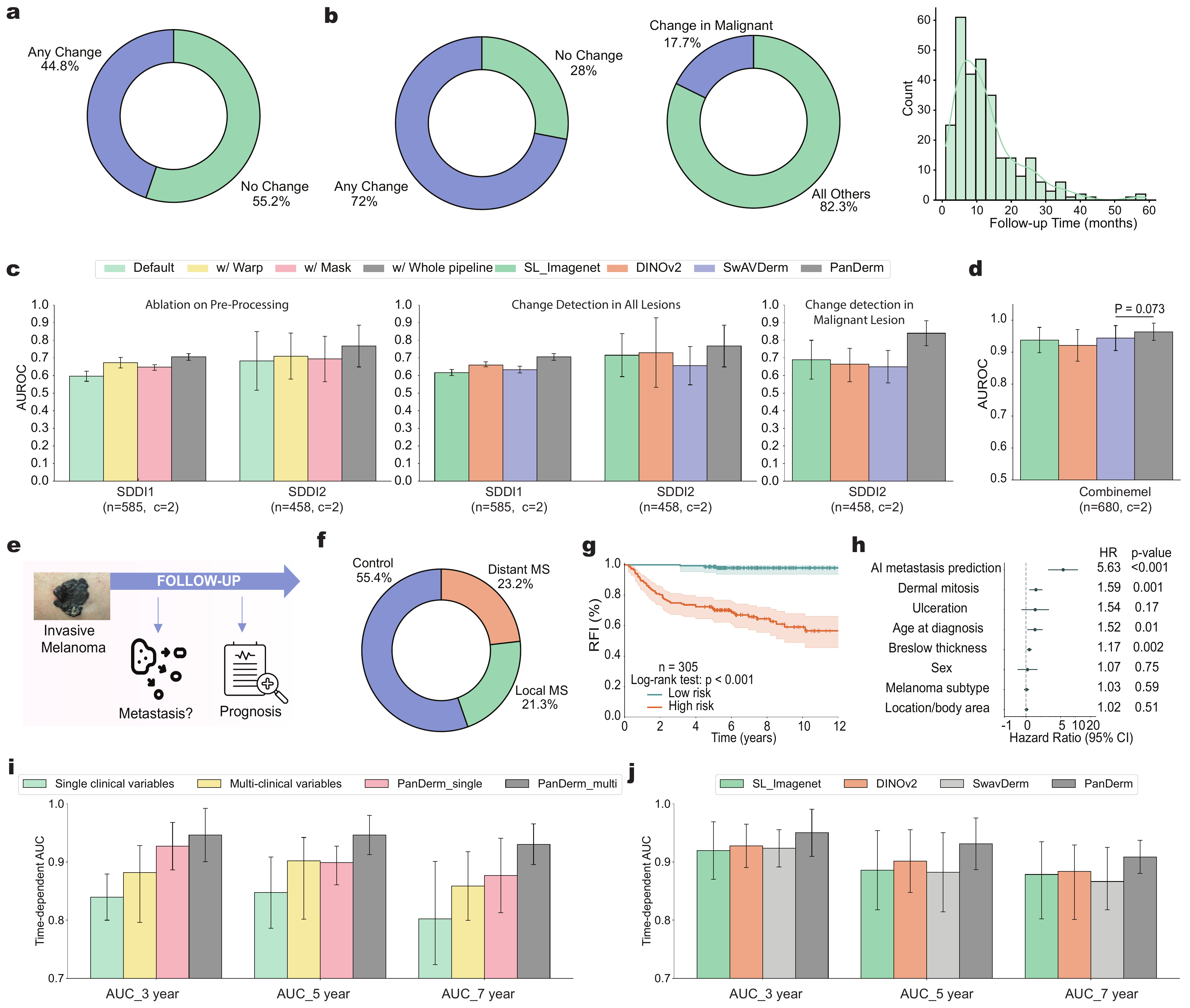}
\caption{\textbf{Short-term lesion change detection and metastasis prognosis results.} \textbf{a.} SDDI1 dataset statistics: ratio of changed lesions, the ratio of changed malignant lesions during follow-up, and follow-up time distribution. \textbf{b.} Ratio of changed lesions in SDDI2 dataset. \textbf{c.} Ablation study on pre-processing methods: Default'' (direct input), w/Warp'' (registration only), w/Mask'' (lesion segmentation), and w/Whole pipeline'' (complete pre-processing as in \textbf{Extended Data Fig~\ref{supp_change_method}}). For change detection, all models were evaluated using the whole pre-processing pipeline. \textbf{d.} Performance of binary metastasis prediction (control vs. metastasis) by AUROC. \textbf{e.} Scheme of PanDerm for melanoma metastasis and prognosis prediction. \textbf{f.} Distribution of metastasis types in Combinemel dataset (MS represents metastasis). \textbf{g.} Kaplan–Meier curves for the recurrence-free interval (RFI) in invasive melanoma patients (CombinMel dataset), stratified by PanDerm prediction scores. \textbf{h.} Forest plots of hazard ratios for PanDerm, stratified groups in invasive melanoma patients. \textbf{i.} Time-dependent AUC of PanDerm vs. clinical variable score combinations. \textbf{j.} Time-dependent AUC comparison of PanDerm and other pretrained models. Error bars in \textbf{c-d} and \textbf{i-j} represent 95\% CIs; bar centers indicate the mean value. Estimates computed with five-fold cross-validation.}
\label{fig3}
\end{figure*}

\noindent
Monitoring suspicious melanocytic lesions over a three-month period is a widely accepted procedure for early melanoma detection, as changes often prompt excision to rule out melanoma, whilst stability can be reassuring \cite{sequential2}. \textcolor{black}{We evaluated PanDerm's ability to detect subtle changes in lesions over time by analyzing pairs of sequential dermoscopic images. To ensure accurate comparison despite variations in imaging conditions, we developed a comprehensive image processing system that standardizes image quality and alignment (\textbf{Extended Data Fig~\ref{supp_change_method}}). This processing system, combined with PanDerm's advanced lesion change detection capabilities \cite{short-term1}, significantly improved change detection accuracy from} 0.596 (95\% CI 0.567-0.624) to 0.706 (95\% CI 0.686-0.725) in SDDI1 (\textbf{Fig~\ref{fig3}a; Fig~\ref{fig3}c left}) (\textit{P} $<$ 0.001) and from 0.683 (95\% CI 0.517-0.894)  to 0.767 (95\% CI 0.649-0.886) in SDDI2 (\textbf{Fig~\ref{fig3}b; \textbf{Fig~\ref{fig3}c left}}) (\textit{P} $<$ 0.001). Using the optimized pipeline for all models, PanDerm achieved AUROC improvements of 4.3\% in SDDI1 (\textit{P} $<$ 0.001) and 3.7\% in SDDI2 over the next-best model (\textbf{Fig~\ref{fig3}c, middle}). For lesions later diagnosed as malignant, PanDerm achieved an AUROC of 0.840 (95\% CI 0.769-0.911), surpassing the next-best model by 15.0\% (\textit{P} $<$ 0.01) (\textbf{Fig~\ref{fig3}c, right}). Further details on lesion change detection method and dataset details are provided in \textbf{Methods} and \textbf{Extended Data Table~\ref{tab3}-\ref{tab5}}.


\heading{Metastasis prediction and prognosis}

\noindent
We explored PanDerm's potential to predict melanoma progression from dermoscopic images, an emerging approach that could provide valuable prognostic information at the time of diagnosis \cite{meta1,meta2,meta3} (\textbf{Fig~\ref{fig3}e}). We evaluated this capability using 680 dermoscopic images from 370 patients with invasive primary melanoma across multiple international centers (\textbf{Fig~\ref{fig3}f}). PanDerm demonstrated exceptional accuracy in distinguishing melanomas likely to metastasize, achieving an AUROC of 0.964 (95\% CI 0.937-0.991), surpassing the next-best model by 2.0\% (\textit{P} = 0.073) (\textbf{Fig~\ref{fig3}d}). It also showed strong capability in differentiating between local and distant metastasis, outperforming existing methods by 2.8\% (\textit{P} $<$ 0.05) in weighted F1 score (\textbf{Extended Data Table~\ref{tab6}}).

To validate PanDerm's clinical utility for patient risk stratification, we conducted survival analyses using Kaplan-Meier analysis and Cox proportional hazards regression. Patients classified as high-risk by PanDerm showed significantly shorter recurrence-free intervals (RFI) compared to those in the low-risk group (HR: 5.63, 95\% CI: 2.87-11.02, \textit{P} $<$ 0.001) (\textbf{Fig~\ref{fig3}g}). When compared alongside standard clinical risk factors - including sex, age, Breslow thickness, ulceration, dermal mitosis, location, and melanoma subtype - PanDerm's predictions emerged as the strongest indicator of recurrence risk in multivariate Cox regression (\textbf{Fig~\ref{fig3}h}). It maintained high predictive accuracy over extended follow-up periods, with time-dependent AUCs of 0.950 (95\% CI 0.910-0.991), 0.931 (95\% CI 0.887-0.976), and 0.909 (95\% CI 0.880-0.937) at 3, 5, and 7 years, exceeding multi-clinical variables by 6.8\%, 2.9\%, and 5.0\%, respectively (\textbf{Fig~\ref{fig3}i}). Combining PanDerm's predictions with clinical factors further improved long-term prognostic accuracy in AUCs at 5 and 7 years. PanDerm also consistently outperformed other AI approaches (\textbf{Fig~\ref{fig3}j}), showing improvements of 2.3\%, 3.0\%, and 2.5\% at 3, 5, and 7 years respectively. Further details are provided in \textbf{Methods}.

\heading{Skin phenotype, risk assessment and malignant lesion screening using TBP}

\begin{figure*}
\centering
\vspace{-5mm}
\includegraphics[width=0.95\textwidth]{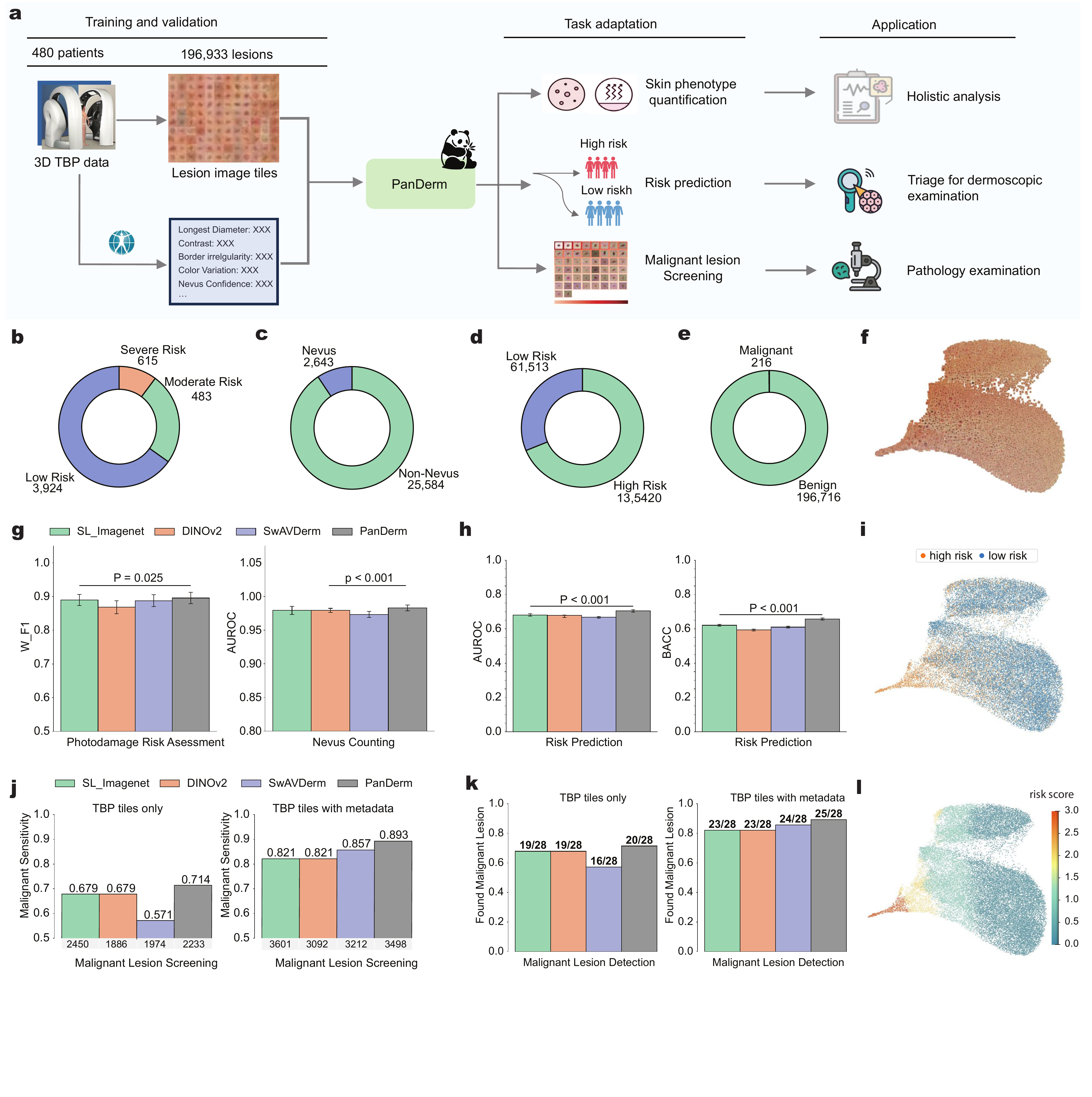}
\vspace{-16mm}
\caption{\textbf{Skin phenotype assessment and melanoma screening using TBP.} \textbf{a.} Illustration of PanDerm processing multimodal TBP data (HOP \& MYM dataset) for skin examination, phenotype score quantification, risk stratification, and malignant lesion screening. \textbf{b-c.} Class distribution of skin phenotype for solar damage risk (\textbf{b}) and nevus count (\textbf{c}) in datasets. \textbf{d-e.} Class distribution of risk groups and malignant lesions. \textbf{g.} Solar damage risk assessment and nevus counting performance by W\_F1 and AUROC. \textbf{h.} Risk prediction performance by AUROC and BACC (Balanced Accuracy). Error bars in \textbf{g-h} show 95\% CIs; bar centers represent mean value. Estimates were computed with nonparametric bootstrapping using 1000 bootstrap replicates. \textit{P}-values calculated with a two-sided t-test. \textbf{j.} Malignant lesion screening performance by sensitivity. Left: using only TBP data. Right: integrating measurement information. The numbers below the bars indicate the recommended suspicious lesion count. \textbf{k.} Number of malignant lesions detected in the test set. \textbf{f.} UMAP plot of PanDerm screening results for test lesions. \textbf{i.} UMAP plot of human screening results for test lesions. \textbf{l.} UMAP plot of PanDerm risk prediction results for test lesions.}
\label{fig4}
\end{figure*}

\noindent
We next evaluated PanDerm's capability in analyzing whole-body imaging (Total Body Photography, TBP) \cite{tbp1,tbp2,isic24} (\textbf{Fig~\ref{fig4}a}). Unlike close-up imaging of individual lesions, TBP enables comprehensive patient-level analysis, particularly for critical melanoma risk factors such as photodamage and nevus count \cite{risk1,risk2,sun1}. In a cohort of 480 patients with 196,933 lesions from Australia, PanDerm achieved a weighted F1 score of 0.896 (95\% CI 0.879-0.913) for photodamage assessment and an AUROC of 0.983 (95\% CI 0.979-0.987) for nevus counting, surpassing all other models (\textit{P} $<$ 0.05 and \textit{P} $<$ 0.001, respectively; \textbf{Fig~\ref{fig4}b}, \textbf{c}, \textbf{g}). Even with limited training data (10\% of the full dataset), PanDerm maintained superior performance (\textbf{Extended Data Fig~\ref{supp_fig4}}). In lesion-specific risk stratification, PanDerm also ranked first with an AUROC of 0.705 (95\% CI 0.698-0.712) and BACC of 0.657 (95\% CI 0.6513-0.663), with all results statistically significant (\textit{P} $<$ 0.001; \textbf{Fig~\ref{fig4}d}, \textbf{h}).

In a clinical validation study, PanDerm effectively identified malignant lesions among a large number of benign ones (216 malignant vs. 197,716 benign lesions) from the HOP \cite{hop} and MYM \cite{mym} cohort (\textbf{Fig~\ref{fig4}e}). Using TBP images alone, PanDerm achieved a sensitivity of 0.893, outperforming the next-best model by 4.2\% (\textbf{Fig~\ref{fig4}j left}). When clinical measurements were available for all models, PanDerm maintained its advantage with a 3.5\% higher sensitivity (\textbf{Fig~\ref{fig4}j right}), reaching a sensitivity of 0.893. Significantly, it detected malignant lesions in 79 out of 80 patients while reducing unnecessary examinations by 60.8\% compared to melanographers (3498 vs. 8913 lesions recommended for detailed examination) (\textbf{Fig~\ref{fig4}j, k}; \textbf{Extended Data Table~\ref{tab10}}).

We observed that PanDerm's analysis approach aligned well with established clinical practice, particularly the "ugly duckling" concept \cite{ud} of identifying atypical lesions through comparison with a patient's other lesions. This was demonstrated through UMAP visualization (\textbf{Fig~\ref{fig4}f}), where PanDerm's feature effectively separated suspicious lesions. The clustering patterns in PanDerm's risk assessment (\textbf{Fig~\ref{fig4}l}) showed correspondence closely with human screening patterns (\textbf{Fig~\ref{fig4}i}), illustrating its exceptional performance in malignant lesion screening.  Additional details are provided in the \textbf{Methods} section, \textbf{Extended Data Table~\ref{tab7}-\ref{tab10}}, and \textbf{Extended Data Fig~\ref{supp_fig5}}.

\heading{Skin lesion segmentation}

\noindent
We evaluated PanDerm's performance on skin lesion segmentation using ISIC2018 \cite{isic2018} and HAM10000 \cite{ham10000} datasets. Compared to existing methods including SL-Imagenet, autoSMIM \cite{autosmim}, and BATFormer \cite{autosmim}, PanDerm achieved significantly higher performance, surpassing the next-best by 3.1\% and 1.9\% in the Jaccard index on both datasets (\textit{P} $<$ 0.001; \textbf{Extended Data Fig~\ref{supp_fig1}a}, \textbf{b}). PanDerm's performance was particularly noteworthy in label-limited scenarios, matching the next-best model while using only 5\% of the training data (104 and 350 images for ISIC2018 and HAM10000, respectively; \textbf{Extended Data Fig~\ref{supp_fig1}c}, \textbf{d}). When compared to MedSAM \cite{medsam}, a medical image segmentation foundation model, PanDerm showed slightly better accuracy (0.5\% improvement, \textit{P} = 0.025 and 0.112; \textbf{Extended Data Table~\ref{tab13}}). This is particularly impressive as PanDerm achieves this performance without specialized training for image segmentation. Additionally, PanDerm offers practical advantages in clinical settings, processing images about 4-5 times faster than MedSAM while using less computational resources (\textbf{Extended Data Table~\ref{tab14}}). Visual examples and detailed performance metrics are provided in \textbf{Extended Data Fig~\ref{supp_fig2}} and \textbf{Table~\ref{tab11}-\ref{tab14}}.

\begin{figure*}
\centering
\vspace{-5mm}
\includegraphics[width=\textwidth]{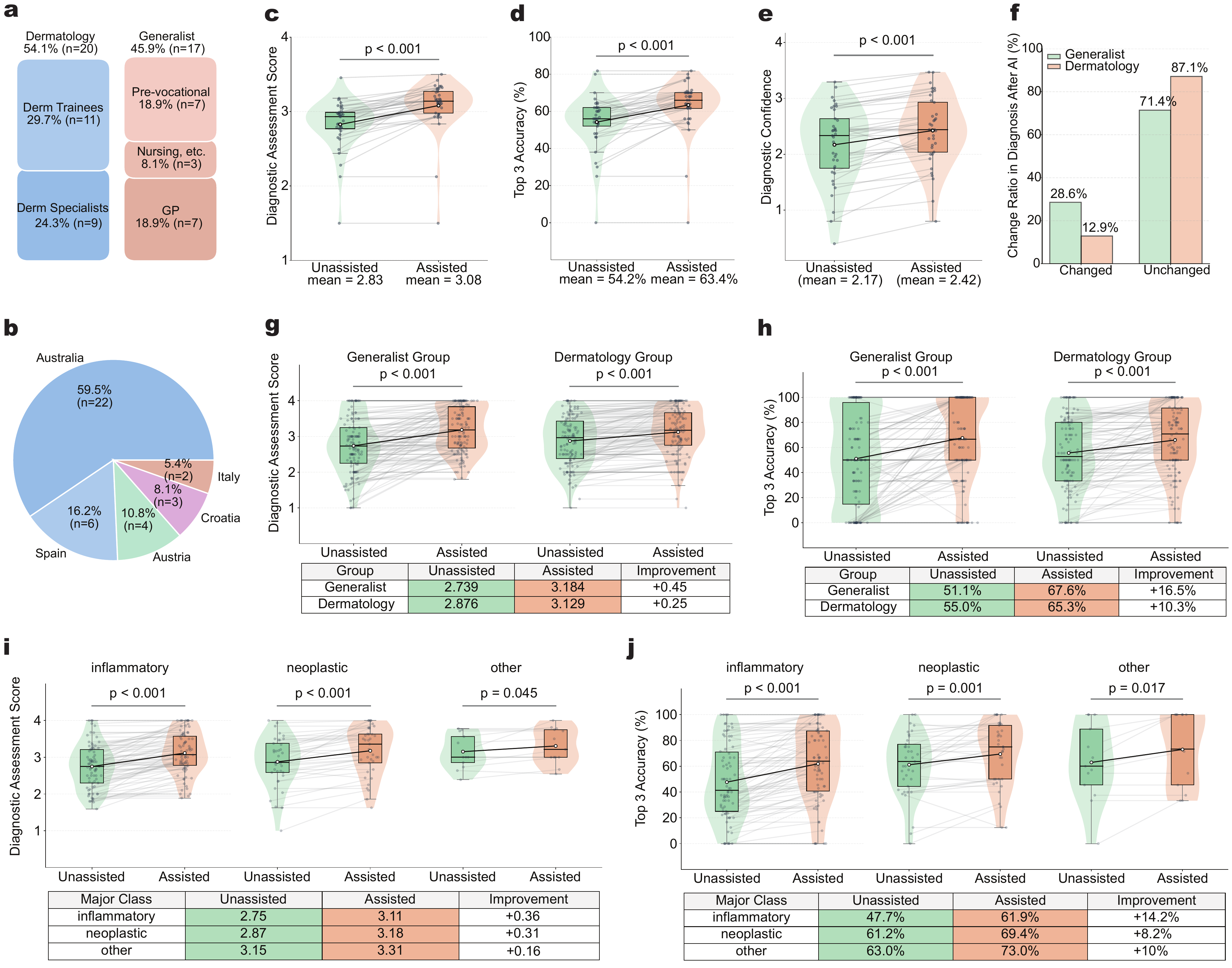}
\caption{\textbf{Performance of PanDerm in Human-AI Collaborative Assessment of 128 Skin Conditions using Clinical Images.}
\textcolor{black}{\textbf{a.} Number of readers (n=37) in this study. Each reviewed up to 50 of 200 cases. Readers were divided into two groups based on specialization: The dermatology group (n=20), including dermatology residents and specialists, and the generalist group (n=17), including pre-vocational trainees, general practitioners, nurses, and clinical trial assistants. \textbf{b.} Geographic distribution of readers. \textbf{c-e.} Comparisons without and with PanDerm support across all readers (n=37) for: (\textbf{c}) Top-1 diagnostic assessment score (1-4), (\textbf{d}) Top-3 diagnostic accuracy, and (\textbf{e}) diagnostic confidence score (1-4). \textit{P}-values in \textbf{c-e} calculated using two-sided paired t-test. \textbf{f.} Change ratio in diagnosis after PanDerm support across specialization groups. \textbf{g-h.} Comparisons without and with PanDerm support stratified by specialization groups for: (\textbf{g}) Top-1 diagnostic assessment score and (\textbf{h}) Top-3 diagnostic accuracy. \textit{P}-values in \textbf{g-h} calculated using two-sided paired t-test (n=126 classes). \textbf{i-j.} Comparisons without and with PanDerm support stratified by major class types for: (\textbf{i}) Top-1 diagnostic assessment score and (\textbf{j}) Top-3 diagnostic accuracy. \textit{P}-values in \textbf{i-j} calculated using two-sided paired t-test (n=126 classes).}}
\label{fig5}
\end{figure*}
\heading{\textcolor{black}{Reader studies}}

\noindent
\textcolor{black}{To assess PanDerm's clinical applicability, we conducted three reader studies evaluating its capabilities across different aspects and modalities of dermatological diagnosis, as follows.}

\noindent
\textbf{Early melanoma detection compared with clinicians.} \textcolor{black}{To examine PanDerm's capability in early melanoma detection, we compared it with 12 human reviewers (7 experienced dermatologists and 5 dermatologist trainees) using sequential dermoscopic images from Alfred Hospital \cite{sddi_alfred}, featuring multiple follow-up images of the same lesions over time.} The study evaluated two key aspects: overall diagnostic accuracy and early melanoma detection capability. In terms of overall accuracy, PanDerm outperformed the average human reviewer by 10.2\% and surpassed the best-performing human by 3.6\%. For early detection, \textcolor{black}{we assessed the time point of the first suspicious changes detected in sequential images relative to clinical diagnosis and biopsy confirmation.} PanDerm demonstrated superior ability in this challenging task, correctly identifying 77.5\% (69 out of 89) of melanoma lesions at the first imaging time point, compared to only 32.6\% (29 correct diagnoses) for human reviewers (\textbf{Extended Data Fig~\ref{supp_fig_r1}}). Individual dots in the histograms represent the earliest correct diagnosis time points for both PanDerm and human reviewers, visualizing the comparative early detection performance.

\noindent
\textbf{Human-AI collaboration \textcolor{black}{for skin cancer diagnosis using dermoscopic images.}} We evaluated PanDerm's impact on clinicians' diagnostic accuracy across seven pigmented lesion classes using dermoscopic images (\textbf{Extended Data Fig~\ref{supp_fig_r2}a}). The study included 41 clinicians with varying levels of competency who evaluated cases both with and without PanDerm's multi-probability prediction support. PanDerm's assistance significantly increased overall diagnostic accuracy from 0.69 (95\% CI 0.65-0.73) to 0.80 (95\% CI 0.76-0.84, \textit{P} $<$ 0.001; \textbf{Extended Data Fig~\ref{supp_fig_r2}b}). Notably, clinicians with lower competency levels showed the greatest improvement, with accuracy gains of 17\% (\textit{P} = 0.0082) for those with low competency and 12\% (\textit{P} $<$ 0.001) for those with medium competency, while highly competent clinicians demonstrated a 6\% improvement (\textit{P} = 0.039; \textbf{Extended Data Fig~\ref{supp_fig_r2}c}, \textbf{Extended Data Table~\ref{tab25}}). Class-specific analysis revealed significant accuracy improvements in 4 of 7 lesion classes (\textit{P} $<$ 0.05; \textbf{Extended Data Fig~\ref{supp_fig_r2}d}, \textbf{Extended Data Table~\ref{tab24}}). For melanoma diagnosis specifically, PanDerm enhanced clinician accuracy from 0.69 (95\% CI 0.64-0.74) to 0.83 (95\% CI 0.79-0.87, \textit{P} $<$ 0.001).

\noindent
\textbf{\textcolor{black}{Human-AI collaboration for differential diagnosis of 128 skin conditions using clinical photos.}} \textcolor{black}{We conducted a comprehensive reader study evaluating PanDerm's diagnostic capabilities across 128 skin conditions using clinical photos. The study included 37 readers from 5 countries (\textbf{Fig~\ref{fig5}b}) and comprised two groups(\textbf{Fig~\ref{fig5}a}): the dermatology group (n=20; 11 dermatology trainees and 9 specialists) and the generalist group (n=17; 7 pre-vocational trainees, 8 GPs, 1 nurse, and 1 clinical trial assistant). This grouping represents the distinction in specialty training backgrounds between dermatology-trained practitioners and those with general medical training. Each reader assessed up to 50 cases from a 200-case pool, providing their top-3 diagnoses both with and without PanDerm's assistance. Each reader evaluated up to 50 cases from a 200-case pool, providing their top-3 diagnoses with and without PanDerm's support. Four experienced dermatologists developed a standardized ontology for condition categorization (\textbf{Extended Data Fig~\ref{supp_fig_ontology_sunburst_plot})}. Performance was assessed primarily using two metrics: a 4-point diagnostic assessment scale for top-1 diagnosis (4: exact ontology match to 1: significant mismatch) and top-3 diagnostic accuracy, with three independent dermatologists scoring and resolving discrepancies through panel review.
PanDerm's assistance significantly improved average top-1 diagnostic scores of all readers from 2.83 to 3.08 (\textit{P} $<$ 0.001; \textbf{Fig~\ref{fig5}c}) and top-3 diagnostic accuracy from 54\% to 63.4\% (\textit{P} $<$ 0.001; \textbf{Fig~\ref{fig5}d}), while increasing readers' diagnostic confidence (2.17 to 2.42, \textit{P} $<$ 0.001; \textbf{Fig~\ref{fig5}e}). The impact was particularly pronounced in the generalist group, showing higher diagnosis modification rates (28.6\% vs 12.9\% in dermatology group; \textbf{Fig~\ref{fig5}f}) and greater improvements in both top-1 diagnostic scores (generalist group: +0.45, dermatology group: +0.25; \textbf{Fig~\ref{fig5}g}) and top-3 accuracy (generalist group: +16.5\%, dermatology group: +10.3\%; \textbf{Fig~\ref{fig5}h}). Analysis by condition classes demonstrated consistent improvements across inflammatory, neoplastic, and other categories (\textit{P} $<$ 0.05; \textbf{Fig~\ref{fig5}i, j}), with inflammatory conditions showing the largest gains (+0.36 in top-1 diagnostic scores, +14.2\% in top-3 accuracy).
Further details on the setup, methodology, readers statistics, and datasets of all three reader studies are provided in \textbf{Methods}, \textbf{Extended Data~\ref{rr1}-\ref{rr3}} and \textbf{Extended Data~\ref{tab_humanAIcomparison}-\ref{tab25}}.}

%% file: sections/3-discussion.tex
\noindent Despite significant advances in AI technology, its application in clinical medicine remains fragmented and underutilized. Current AI systems are often restricted to isolated tasks and are unable to address the diverse demands of medical decision-making. This limits AI's potential in supporting clinicians in disease diagnosis and management. Dermatology, with its complex requirements, including holistic patient assessment, lesion-specific analysis, and potential use of various imaging modalities, serves as an ideal use case for demonstrating AI's capabilities across multiple interconnected clinical tasks. Success in this domain could pave the way for broader adoption of AI models across healthcare.

In this study, we introduce PanDerm, a versatile dermatology foundation model trained through self-supervised learning on over 2 million multimodal dermatological images. Central to PanDerm's development was the curation of a large and diverse image dataset sourced primarily from in-house collections and carefully selected public repositories. This approach contrasts with previous efforts, such as SwAVDerm \cite{swavderm}, which relied on web-sourced skin data, inadvertently incorporating images from commonly used benchmarks like ISIC \cite{isic} and DermNet \cite{Dermnet}, increasing the risk of data leakage and compromising evaluation validity. Our strategy minimizes this risk, ensuring that benchmark evaluations accurately reflect real-world model performance.

\textcolor{black}{To evaluate PanDerm's clinical utility, we conducted validations across 28 benchmark datasets, spanning comprehensive skin cancer assessment and a diverse set of primary care dermatological conditions.} For skin cancer-related assessment, PanDerm outperformed existing models in specialized tasks across various modalities, including risk stratification of lesions, phenotype assessment, detection of lesion changes and malignancy, multi-class cancer diagnosis, lesion segmentation, and metastasis prediction and prognosis. In particular, PanDerm achieved the most results using only 10\% of the task-specific training data typically required by existing models, helping alleviate the scarcity of specialist-labeled data in medical AI. \textcolor{black}{In primary care dermatology settings, PanDerm also outperformed comparative models in diagnosing a diverse set of conditions such as inflammatory diseases, infectious conditions, and frequently encountered dermatoses.} These capabilities stem from its rich knowledge representation, developed through pretraining on varied dermatological image modalities and conditions, leading to consistent and significant performance improvements across tasks and modalities.

\textcolor{black}{Three reader studies supported these benchmark findings, suggesting PanDerm's potential to assist clinical practice across different healthcare settings or specialty training backgrounds. In skin cancer diagnosis, PanDerm demonstrated two main capabilities: it helped improve diagnostic accuracy across clinicians of varying competence levels or specialties when used as a support tool, and notably, showed the ability to identify concerning lesions before clinician detection — potentially facilitating earlier intervention when melanomas are thinner and more amenable to treatment. In general dermatology, PanDerm significantly improved readers' performance in the differential diagnosis of 128 skin conditions, such as inflammatory dermatoses, cutaneous neoplasms, pigmentary disorders, and hair conditions. The improvement appeared more substantial for generalists, such as primary care providers, in evaluating inflammatory conditions — which represent a considerable portion of dermatological consultations in everyday clinical practice. Given the limited access to specialist expertise in primary care settings, where most skin conditions are initially evaluated \cite{global1,global2}, these findings indicate PanDerm's potential to help address the need for dermatological expertise across different healthcare settings through both its technical capabilities and clinical applications.}

The scaling behavior observed in PanDerm's performance, particularly in relation to pretraining dataset size, aligns with recent trends in foundation model development \cite{eyefm,path_fm1,path_fm2,derm1m}. However, achieving this in dermatology required addressing unique challenges in large-scale medical data acquisition and multimodal data integration. Our analysis revealed two key technical insights with broader implications for medical AI development. First, using CLIP \cite{clip} as a teacher model for semantic feature alignment achieved superior training data efficiency (\textbf{Fig~\ref{fig1}f}), outperforming DINOv2 \cite{dinov2}. This is especially relevant in healthcare, where assembling large-scale datasets remains a significant challenge - for context, DINOv2 typically requires 142 million images, a scale impractical in medical domains. Second, our masked feature reconstruction approach proved particularly effective at capturing subtle diagnostic features in skin images compared to methods like MAE \cite{mae}, an essential capability for clinical applications. These advantages enabled PanDerm to improve upon both traditional models such as ResNet50 \cite{resnet} and ViT large \cite{vit}, \textcolor{black}{and recent generalist medical foundation models like BiomedGPT \cite{biomedgpt}. While generalist models advance broader biomedical AI development, their pursuit of breadth across medical domains often limits their effectiveness in specialized areas. Our results suggest that specialty-specific foundation models, designed with consideration of clinical workflows and imaging requirements, may offer more practical solutions. This approach could inform AI development across other medical specialties where multiple imaging modalities play crucial diagnostic roles.}

\textcolor{black}{Despite promising results, we acknowledge several limitations in our evaluation scope and methodology. While our validation covered approximately 200 skin conditions across major categories (e.g., inflammatory diseases, infections, neoplasms, benign growths, pigmented lesions, and vascular anomalies), this represents only a fraction of known dermatological conditions (over 1,000 diagnoses), and is smaller than some previous studies (e.g., Liu et al \cite{liu2020deep} with 445 conditions). Specifically, our coverage of rare genetic disorders, complex systemic diseases, and clinical variants remains limited. Regarding model robustness and fairness, while our benchmark evaluations (\textbf{Extended Data Tables}~\ref{tab21}-\ref{tab22}) demonstrate consistent performance across different settings (anatomical locations, age groups, genders, and skin tones), several constraints exist: the evaluation mainly reflects overall accuracy rather than disease-specific analysis, has varying disease coverage across anatomical locations, and focuses primarily on single imaging modalities. A more comprehensive evaluation framework \cite{trust} integrating these aspects will be necessary for further assessing the PanDerml's robustness. Further, recent studies \cite{f17k,ddi,fair2} have highlighted important challenges in dermatological AI systems, particularly in human-AI interactions. Although our evaluations show stable cross-skin tone performance without explicitly balanced training data (as demonstrated necessary by Daneshjou et al \cite{ddi}), comprehensive bias assessment requires metrics beyond overall accuracy. Groh et al \cite{fair2} further revealed that equitable standalone performance may not translate to unbiased human-AI collaboration, which is crucial for clinical deployment. To address these limitations, future work should develop standardized protocols for cross-demographic evaluations using more comprehensive fairness metrics and investigate biases in human-AI collaborative settings. International collaborations like ISIC \cite{isic} will be crucial for creating representative datasets and establishing robust fairness standards.}


\textcolor{black}{In conclusion, PanDerm demonstrates the potential of multimodal specialty-specific foundation models in addressing the diverse clinical needs across specialized and routine clinical practice in dermatology.} Through comprehensive pretraining on diverse dermatological images and validation across multiple clinical scenarios, the model showed robust performance across different use cases. Our development approach, combining systematic data curation, advanced self-supervised learning, and rigorous clinical validation, provides a framework for developing medical AI systems \textcolor{black}{that can adapt to varying levels of clinical expertise and healthcare settings.} These findings suggest promising directions for developing foundation models in other medical specialties where the integration of diverse imaging modalities and complex clinical workflows is crucial for patient care.

%% file: supplement/0-pretraining.tex
\heading{Pretraining dataset for developing PanDerm}

\noindent
We curated an extensive pretraining dataset comprising 2,149,706 unlabeled multimodal skin images to develop PanDerm. This diverse dataset encompasses 4 imaging modalities and 11 data sources, including total body photography (TBP), dermatopathology, clinical, and dermoscopic images. The composition of the dataset is as follows: 757,890 (35.3\%) TBP tiles, 537,047 (25.4\%) dermatopathology tiles, 460,328 (21.4\%) clinical images, and 384,441 (17.9\%) dermoscopic images. The inclusion of multiple imaging modalities aims to provide a comprehensive representation of skin lesions and conditions, enabling the model to learn robust features across different visual representations. This large-scale dataset serves as the foundation for pretraining PanDerm, allowing it to capture the intricate patterns and characteristics of various skin conditions across different imaging techniques.


\hheading{MYM cohort (TBP).} \textcolor{black}{The MYM cohort \cite{mym} represents an in-house dataset designed to study the natural history of melanocytic naevi. This cohort, which included 193 participants from a population-based cohort in Australia, was recruited from the electoral roll, with eligibility criteria including. The cohort maintained a relatively balanced gender distribution with a slight female predominance (42.5\% female, 57.5\% male). The mean age of participants was 52.2 years (SD ± 12.0 years), with ages ranging from 23 to 70 years. The study population was predominantly of European/British ancestry (97.9\%). Regarding sun response characteristics, 26.9\% of participants reported burning easily and tanning slightly, 31.4\% reported moderate burning and gradual tanning, and 11.0\% reported rarely burning and tanning well, representing a predominantly fair-skinned population with varying degrees of UV sensitivity. 3D TBP was conducted using a VECTRA WB360 (Canfield Scientific Inc., Parsippany, NJ, USA), which instantaneously captures 92 cross-polarised 2D images with standardized lighting and subsequently merges them to create a 3D avatar. The average number of lesion tiles per subject from TBP was approximately 500. Demographic factors were collected using standard questionnaires, with clinical characteristics collected by research assistants. The final dataset of automatically detected lesion image tiles $\geq$ 2 mm in diameter comprises 405,856 images.}

\hheading{HOP cohort (TBP).} \textcolor{black}{TThe HOP study \cite{hop} is an in-house sequential dataset targeting high-risk melanoma individuals. The study cohort comprised 314 participants (120 males, 194 females) with a mean age of 56.1 years (SD = 12.8). The majority of participants were of European/British or other European ancestry (94.6\%), with small proportions of other/mixed ancestries (5.4\%). Most participants were born in Australia (85.4\%), with 14.6\% born overseas. In terms of skin cancer history, 96.8\% of participants reported previous melanoma while 3.2\% had no melanoma history. For non-melanoma skin cancer, 62.1\% reported previous occurrence while 37.9\% had no such history. Innate skin color (non-sun exposed skin) was reported as fair for 85.7\%, medium for 14.0\%, and olive for 0.3\% of the cohort. Inclusion criteria for the HOP study were: at least one melanoma (including in situ) diagnosed before the age of 40 years, two or more melanomas (including in situ) diagnosed before the age of 65, a strong family history (two or more first-degree relatives affected) and/or known pathogenic mutations in a hereditary melanoma gene, and/or a diagnosis of dysplastic naevus syndrome. As with the MYM cohort, 3D TBP imaging was performed using the VECTRA WB360 system, following the same imaging protocol and data collection procedures. Demographic and clinical data were similarly collected through standardized questionnaires and by research assistants.}

\hheading{MYM \& HOP cohort (dermoscopic).} The MYM and HOP datasets also contain 38,110 dermoscopic images from suspicious lesions identified during the studies. These images provide a complementary view of the lesions of interest, offering detailed visualization of surface and subsurface structures that may be indicative of various skin conditions, particularly melanoma.

\hheading{MMT dataset.} The MMT dataset is an in-house collection amassed from over 150 clinics across Australia and New Zealand over a 15-year period. This extensive dataset primarily consists of paired polarized dermoscopic and clinical images. From this comprehensive collection, we curated a subset containing 316,399 dermoscopic images and 310,951 clinical images, providing a rich source of pretraining data for training purposes.

\hheading{ACEMID pathology pilot study.} \textcolor{black}{This dataset comprises 54 patients from two Australian centers: Queensland PAH (48.1\%) and NSW MIA (51.9\%), with a slight male predominance (57.4\%) and ages ranging from 19 to 75 years (mean 53.4 years). Most patients (81.5\%) were classified as "Very High" risk for melanoma, with others being "High" risk (14.8\%) or "Low/Average" risk (1.9\%). Based on the first pathologist's diagnoses, lesions were predominantly naevi (68.5\%, including common/dermal/congenital and dysplastic variants), followed by melanomas (24.1\%, mostly in situ), and other lesions (7.4\%). While most patients (66.7\%) had single lesions examined, others had multiple lesions reviewed (2-5 lesions per patient). Notable diagnostic variability between pathologists was observed, particularly in assessing dysplastic naevi and early melanomas.}

\hheading{NSSI.} \textcolor{black}{The NSSI dataset is an in-house sequential collection containing 29,832 dermoscopic images from 1,254 individuals residing in Brisbane, Australia. These images were collected as part of the Brisbane Naevus Morphology Study (2009-2014) using a digital dermatoscope attached to a Fotofinder ATBM imaging system, producing 768 × 576 pixel images at 96 dpi. The cohort comprised 48\% male and 52\% female participants, with a median age of 46 years (range: 11-88 years, $<$ 1\% under 18). Just under half the cohort (47\%) had a personal history of melanoma. Skin color was predominantly fair (51\%, Fitzpatrick Type 1), followed by medium (41\%, Type 2) and olive (8\%, Type 3). The majority of participants reported British/Irish ancestry (87\%), followed by Western/Northern European (21\%), with smaller proportions of Southern European (3\%), Asian (2\%), and other ancestries ($<$ 1\% each). The study design included up to 7 time points per participant at 6-month intervals over 3 years, with some variation due to enrollment timing and follow-up adherence. Individual lesions were assigned consistent identification numbers across visits to facilitate temporal analysis.}

\hheading{Edu1 \& Edu2.} \textcolor{black}{The Edu1 and Edu2 datasets comprise 81,947 and 67,430 clinical images respectively, curated from in-house educational resources. These comprehensive collections cover a wide range of dermatological conditions including inflammatory and autoimmune disorders (such as psoriasis, and atopic dermatitis), common infections (such as herpes simplex, molluscum contagiosum, and tinea corporis), pigmentary disorders (like melasma and vitiligo), nail conditions (such as psoriatic nail disease and onychomycosis), vascular lesions (like port-wine stains and pyogenic granulomas), and both benign and malignant tumors (such as melanoma, basal cell carcinoma, and squamous cell carcinoma). The datasets also include rare conditions and genetic disorders.}

\hheading{ISIC2024.}
ISIC2024 \cite{isic24} is an open-source TBP-based dataset designed for identifying skin cancers among skin lesions cropped from 3D total body photographs. For our pretraining purposes, we selected a subset of the dataset, stratified by institutions, containing 352,034 tile images. 

\hheading{TCGA-SKCM.}
The TCGA-SKCM dataset \cite{tcga-skcm} is derived from The Cancer Genome Atlas (TCGA) project, which characterized the mutational landscape of human skin cutaneous melanoma (SKCM). This dataset contains 475 slides, which have been processed into 377,764 patch images. 

\hheading{UAH89k.}
The UAH89k dataset \cite{kriegsmann2022deep} is a subset of a larger collection, comprising 269 histopathology whole slide images (WSIs) sourced from the archives of the Institute of Pathology, Heidelberg University, the MVZ for Histology, Cytology and Molecular Diagnostics Trier, and the Institute for Dermatopathology. This dataset provides additional histopathological data, further enriching the model's understanding of skin conditions at the microscopic level.

\heading{Detail of model architecture and pretraining}

\noindent
PanDerm is a self-supervised learning model designed for the dermatology field, built upon the success of existing self-supervised learning techniques in the natural image domain \cite{caev2}. At its core, the architecture comprises a ViT-Large visual encoder \cite{vit}, a mask regressor, and a CLIP-Large \cite{clip} teacher model. The ViT-Large encoder, with its 24 transformer blocks and 1024 dimensional embeddings, processes 224 $\times$ 224-pixel images, while the CLIP-Large teacher model handles slightly smaller 196 $\times$ 196-pixel inputs. The training process incorporates two primary objectives: masked latent alignment and visible latent alignment loss. Initially, the input image undergoes masking, with the mask ratio proportional to the encoder's complexity (50\% for ViT-Large). The encoder then processes visible patches to produce latent representations, while the regressor predicts the latent representations of masked patches using these visible latent and mask tokens. The model focuses on the encoder-regressor structure without a separate decoder component. The regressor assumes the responsibility of predicting the latent representations of masked patches, allowing for more efficient processing and learning. For target supervision, the unmasked image is fed through the CLIP model, generating supervision divided according to visible and masked patch locations. The visible latent alignment loss is directly applied to the latent representations of visible patches computed by the encoder. Concurrently, the masked latent alignment loss acts on the latent representations of masked patches predicted by the regressor. Both of these loss functions use CLIP latent representations as their supervision signals. The regressor in PanDerm operates similarly to a cross-attention mechanism. It uses learnable mask tokens as queries, while the keys and values are derived from the concatenation of visible patch representations and the output of previous layers. This design allows the regressor to effectively infer the content of masked regions based on the context provided by visible areas.
Optimization primarily focuses on aligning the visible and masked patch predictions with their corresponding CLIP latent supervisions. This approach enables PanDerm to extract rich, semantically meaningful representations from dermatological images without relying on explicit labels.

For pretraining, we continued to train the model (initially trained on ImageNet-1K) on our dataset of over 2 million unlabeled multimodal skin images, representing diverse dermatological conditions. We set the batch size on each GPU to 480, with an effective batch size of 1920. Following masked image modeling practices \cite{beitv2}, we used a 50\% mask ratio. To pretrain our model, we used AdamW as the optimizer with an initial learning rate of 1.5e-3. We apply simple data augmentation such as random resized cropping and horizontal flipping during pretraining. We trained our model for 500 epochs with a warmup of 20 epochs. The pretraining phase used 4 80GB NVIDIA H100 GPUs and took approximately 5 days and 7 hours. We chose the last epoch checkpoint as our final model weights. Please refer to \textbf{Extended Data Table~\ref{tab:pretrain}} for more detailed pretraining hyperparameter configurations.

\hheading{Target representations (Teacher model) of PanDerm.}
Prior work \cite{beitv2,caev2,milan} demonstrated that target representations produced by the teacher model for masked image modeling are essential for impacting model performance. We ablated different teacher models, including two widely used models that demonstrated promising performance (CLIP-base and CLIP-large), a biomedical domain-specific CLIP (BiomedCLIP\cite{biomedclip}), and a dermatology-specific CLIP (MONET\cite{monet}). We observed that CLIP-large pretrained on the natural domain can outperform biomedical-specific CLIP and dermatology-specific CLIP. This can be attributed to the limited data scale of skin images in medical domain CLIP models. Thus, CLIP-large remains the best teacher model for creating target representations for masked image modeling in dermatology. Based on these findings, we selected CLIP-large as the teacher model; the performance of our model when incorporating CLIP-large teachers was significantly improved and also outperformed CLIP-large itself. Please refer to \textbf{Extended Data Table~\ref{tab:target_rep}} for detailed results.

\hheading{Linear probing vs fine-tuning for PanDerm.}
One emergent capacity of foundation models in the natural image domain is that the model's features are ready for downstream tasks without needing to fine-tune the encoder model, such as in DINOv2 \cite{dinov2}. We explored whether PanDerm could achieve this capacity despite having different training objectives. We found that our model using simple linear probing can perform comparably with expensive full-parameter fine-tuning. This suggests that PanDerm's features are already well-suited for diverse downstream multimodal skin-related tasks without requiring further training. Detailed results are in \textbf{Extended Data Table~\ref{tab:lp_ft}}.

%% file: supplement/1-evaluation.tex
\heading{Downstream Evaluation Details}

\hheading{Competing self-supervised learning baselines.}
For self-supervised learning methods comparison, we primarily evaluated DINOv2\cite{dinov2}, MAE\cite{mae}, and MILAN\cite{milan}, all utilizing the same ViT-large backbone. We employed the recommended hyperparameter configurations for these models and continued pretraining from their natural image training weights on our pretraining dataset. Subsequently, we fine-tuned these models using identical hyperparameter setups to ensure a fair comparison.

\hheading{Fine-tuning and linear probing.}
In adapting PanDerm to downstream tasks, only the encoder model is utilized. For most tasks, PanDerm's feature quality suffices to achieve competitive performance using simple linear probing. This involves applying a linear classifier (i.e., logistic regression) to the top of extracted features from the PanDerm encoder to evaluate its performance on downstream tasks. For more challenging tasks requiring higher performance, we opted to fine-tune the PanDerm encoder. The fine-tuning tasks include the two reader studies, short-term change detection, skin lesion segmentation, skin cancer detection in ISIC2024, and TBP-based risk stratification. For all other tasks, we employed linear probing. For linear probing, following practices recommended by the self-supervised learning community, we fix the \(\ell_2\) regularization coefficient \(\lambda\) to \(MC/100\), where \(M\) is the embedding dimension and \(C\) is the number of classes, and employ the L-BFGS solver with a maximum of \(1,000\) iterations. For fine-tuning, we adhere to the BEiT V2 setting\cite{beitv2}, utilizing cross-entropy loss with a learning rate of \(5 \times 10^{-4}\). We train models for \(50\) epochs with a warmup of \(10\) epochs. The model exhibiting the best performance on the validation set is selected as the final model. For detailed hyperparameter configurations, please refer to \textbf{Extended Data Table~\ref{tab:finetune}}. In the following sections, we describe tasks with more specific methodological details.

\hheading{Sequential data preprocessing for lesion change detection.} Our proposed sequential data preprocessing method consists of dark corner removal, skin inpainting, hair removal, image registration, and lesion segmentation. For the first two steps, we follow the approach outlined in \cite{dc}. Given an image with or without dark corner artifacts (DCA), we first convert it to grayscale. We then extract the contour by applying OpenCV's \cite{opencv} binary threshold function to the grayscale image, empirically setting the threshold at 100, and using the findContours function with RETR\_TREE mode and CHAIN\_APPROX\_SIMPLE method. We identify the largest area contour in the image, which most closely matches the edge of the DCA, by calculating the area of all existing contours. Using OpenCV's minEnclosingCircle function, we capture a circular area that encompasses this largest contour. To mitigate the effect of gradient colors at the dark corner edges, we scale this circle down to 80\% of its original radius and convert it into a binary mask. Finally, we inpaint the original image using this mask, employing OpenCV's implementation of the Telea algorithm with an inpaint radius of 10. Following dark corner removal and inpainting, we implement a hair removal step to further improve image quality and facilitate more accurate registration. This process begins by converting the image to grayscale. We then apply a black hat morphological operation using a 17$\times$17 structuring element to isolate dark, thin structures (hairs) from the background. The resulting image is thresholded to create a binary mask of the detected hair structures. Finally, we use OpenCV's inpaint function with the Telea algorithm to fill in the hair regions, effectively removing them from the image. This hair removal step is crucial for improving the accuracy of subsequent image registration and analysis. For image registration, we implement an AKAZE \cite{akaze} feature-based approach. The process begins by detecting key points and computing descriptors using the AKAZE algorithm, which is particularly effective for non-linear scale spaces. We utilize OpenCV's AKAZE\_create function, setting the descriptor size to 0, threshold to \(9 \times 10^{-5}\), and number of octaves to 4. Keypoint matching is performed using a Brute Force matcher with Hamming distance and cross-checking enabled. To refine the matches and estimate the transformation between images, we employ the RANSAC (Random Sample Consensus) algorithm implemented via skimage.measure.ransac. This estimates an EuclideanTransform model, which accounts for rotation and translation between the images. The resulting transformation is then applied to warp one image onto the other using skimage.transform.warp with reflection padding and linear interpolation.

\hheading{Siamese network for change detection.} Similar to \cite{short-term1}, we employ a simple Siamese network architecture for change detection, where two identical visual encoders with shared weights from our foundation model process a pair of sequential lesion images captured over a short time frame. Each encoder extracts features from its respective image. These learned features are then concatenated and passed through two fully connected layers, followed by a softmax layer for final classification. For training this siamese network in our binary change detection task, we use a contrastive loss function. This loss is particularly well-suited for Siamese networks as it helps the model learn to distinguish between pairs of images that have changed and those that have not. The contrastive loss encourages the network to minimize the distance between feature representations of image pairs with no significant changes while maximizing the distance for pairs that show meaningful changes. This approach allows the network to learn a similarity metric between image pairs, rather than simply classifying individual images. By doing so, it becomes more sensitive to subtle changes between images and more robust in detecting clinically relevant lesion changes over time. The contrastive loss thus helps the model focus on learning features that are most relevant for distinguishing between changed and unchanged lesion pairs, improving its overall performance in change detection tasks.

\hheading{{Melanoma metastasis prediction and survival analysis.}} We employ a linear probing classifier on our foundation model to predict melanoma metastasis using dermoscopic images from the private CombinMel dataset. Our evaluation encompasses two scenarios: binary metastasis prediction and multi-class metastasis prediction.
In the binary classification, we aim to differentiate between the presence of any metastasis (including local/satellite/in-transit metastasis, lymph node recurrence, and distant metastasis) and its absence. The multi-class prediction presents a more complex challenge, categorizing cases into three groups: control (no metastasis), local/satellite/in-transit metastasis, and distant metastasis. To enhance the robustness and mitigate potential data selection bias, we perform five iterations of dataset splitting into training and testing sets, stratified by melanoma stage. The model is trained using this five-fold data. We linear probe PanDerm with the setting mentioned above. We then generate out-of-fold (OOF) predictions for all lesions and compare these to the ground truth for performance evaluation.

Subsequently, we conduct a multivariate Cox regression analysis, incorporating the metastasis prediction score and clinical variables (age, sex, Breslow thickness, ulceration, dermal mitosis, melanoma subtype, and lesion location) to predict the recurrence-free interval (RFI). This analysis focuses on earlier stages of melanoma (stages I-II). We visualize the relative contribution of individual variables to prognosis prediction using a forest plot. To analyze the correlation between variables and RFI, we employ the Kaplan-Meier method. Patients are stratified into low-risk and high-risk groups based on their binary metastasis prediction scores (median value). The log-rank test is utilized to assess the classifier's ability to predict survival. To evaluate the predictive accuracy at various time points, we generate time-dependent receiver operating characteristic (ROC) curves and calculate areas under the curve (AUCs) at 3, 5, and 7 years. This approach allows us to assess the model's performance over different follow-up periods. 


\hheading{Melanoma screening using TBP.} The melanoma screening algorithm is designed to identify high-risk lesions among whole-body images, aiding clinicians in efficiently detecting potential malignancies. Lesions flagged as high-risk undergo further triage and dermoscopic examination. The screening model integrates three modules: a risk prediction head, an ugly duckling (UD) detection head, and a machine learning (ML) module, utilizing both TBP image data (image tiles) and metadata for comprehensive predictions.
We first fine-tune our foundation model, equipped with the risk prediction head, using TBP image tiles to classify lesions as high-risk or low-risk. All lesion images are resized to 224 × 224 pixels and subjected to data augmentation, including color and geometric transformations. The risk prediction head, comprising a single linear layer, identifies lesions as high-risk if subjected to dermoscopy examination and low-risk otherwise. The UD detection head leverages the ``Ugly Duckling sign'', an effective diagnostic strategy that compares all lesions from the same patient to identify outliers. This approach capitalizes on lesion contextual information. We use the fine-tuned foundation model to extract deep learning features, which are then processed by the UD detection head. This module calculates the distance between each lesion's features and the average features of all lesions from the same patient, employing the interquartile range (IQR) method to select outlier lesions.
The ML module, an extra tree classifier, is trained using TBP metadata, which includes 32 measurements for each lesion from the 3D TPB machine. This module directly predicts malignancy based on pathology labels.
The final screening result combines predictions from all three modules. A lesion is flagged as suspicious for malignancy if any module yields a positive prediction. We evaluate the screening performance at both the lesion and patient levels to ensure comprehensive accuracy assessment.

\hheading{Weakly supervised slide classification.} 
Weakly supervised slide classification tasks are approached using the established two-stage multiple instance learning (MIL) framework. This process encompasses: 1) extraction of instance-level features from discrete tissue regions within the whole slide image (WSI), and 2) development of an adaptable, order-invariant aggregation method to consolidate patch-level data into a comprehensive slide-level representation. For slide preprocessing, we employ the CLAM toolbox\cite{lu2021data}, utilizing its built-in parameters for tissue segmentation. The resulting regions are partitioned into 256 × 256 non-overlapping sections at 20× magnification or its equivalent. During the feature extraction phase, these sections undergo resizing to 224 × 224 and normalization using ImageNet mean and standard deviation parameters. To ensure consistency, all pretrained encoders utilize an identical set of patch coordinates for feature extraction across each WSI.

To evaluate the efficacy of various pretrained encoders in weakly-supervised learning, we implement the Attention-Based Multiple Instance Learning (ABMIL) algorithm\cite{ilse2018attention} with consistent architectural configurations across all comparisons. Our implementation features a two-tier gated ABMIL structure, where the initial fully connected (FC) layer maps input embeddings to a 512-dimensional space, followed by intermediate layers with 384 hidden units. To enhance generalization, we incorporate dropout regularization, applying rates of 0.10 and 0.25 to the input embeddings and subsequent intermediate layers, respectively. The training protocol employs the AdamW optimizer\cite{loshchilovdecoupled} with a cosine learning rate schedule, initializing the learning rate at 1e-4 and setting weight decay to 1e-5. We utilize cross-entropy as our loss metric. The training process is limited to 20 epochs, implementing an early stopping mechanism based on validation loss performance. To ensure robust evaluation, we employ a five-fold cross-validation strategy, stratifying our slide dataset by both case and label attributes.

\hheading{Skin lesion segmentation.}
For skin lesion segmentation, we employ a conventional segmentation paradigm, utilizing a network encoder connected to a segmentation decoder and head. Our proposed PanDerm serves as the encoder in this setup. We benchmark PanDerm against three established models: ViT-large \cite{vit}, autoSMIM \cite{autosmim}, and BATFormer \cite{batformer}. Both ViT and PanDerm use a UperNet decoder, following the official ViT implementation. For autoSMIM and BATFormer, we adhere to their official repository settings. ViT-large and autoSMIM encoders are initialized with ImageNet pretrained weights. To ensure a fair comparison, all images are resized to 224$\times$224. We apply online data augmentation, including color jittering, random rotation, and random flipping, to mitigate overfitting. The training employs an AdamW optimizer with an initial learning rate of 5e-4 and a weight decay of 0.01, with the learning rate decaying according to a cosine schedule. The models are trained for 100 epochs, and we save the model that achieves the best evaluation metrics on the validation set.

\hheading{{Early melanoma detection (Reader study 1).}} We fine-tuned our foundation model on the private SDDI-Alfred dataset\cite{sddi_alfred} using a ten-fold cross-validation approach. We utilized cross-entropy loss with a learning rate of \(5 \times 10^{-4}\). We train models for 50 epochs with a warmup of 10 epochs. The model exhibiting the best AUROC on the validation set is selected as the final model. We then employed an out-of-fold (OOF) prediction approach to generate melanoma predictions for all sequential images. For each image sequence, we recorded the time point at which the model first made a correct diagnosis of melanoma; otherwise, the model was considered to have failed in detecting the melanoma. \textcolor{black}{While biopsy serves as our reference standard, we aimed to explore the algorithm's potential to detect early signs of melanoma progression. Our study focused on identifying suspicious changes in sequential images prior to clinical diagnosis, with the goal of enabling earlier intervention when melanomas are most treatable.} For the human evaluation, 12 clinicians—seven dermatologists with over five years of experience and five dermatology residents with less than five years of experience—were invited to assess the serial dermoscopic data. The images were presented to the reviewers using Qualtrics™ (Provo, UT, USA), with the reviewers blinded to the true diagnoses. For each case, information such as the patient's age, sex, lesion location, and date of imaging was provided. Initially, only the first dermoscopic image in the sequence was shown, and reviewers were asked to classify the lesion as either benign or malignant. As they progressed through the sequence, side-by-side image comparisons were made available to assess changes over time. Once a diagnosis was submitted, it could not be revised. To mitigate bias, we included 10 single time-point melanoma images, preventing reviewers from assuming that the first image in a series was benign. We then compared the diagnostic performance of the clinicians with our model, focusing on the time point at which a malignant diagnosis was first made by either the clinicians or the algorithm.

\hheading{Human-AI collaboration for skin cancer diagnosis (Reader study 2).}
\noindent
The reader study was conducted using DermaChallenge, a web-based platform developed and hosted by the Medical University of Vienna for online education on dermatoscopy, as described in previous studies \cite{humanai,rl}. To ensure proper authentication and data management, readers were 
required to register with a unique username, valid email address, and 
password. Active users on the platform, who previously actively agreed to be contacted, were recruited via a single email. Before commencing to the study phase, all users had to finish three introduction levels to be familiarized with the platforms' user interface and image types. The number of correct answers in the first iteration of these levels normalized against the mean score of the entire DermaChallenge platform user base, served as a score of experience. Users were grouped into ``Low'' (n=11), ``Medium'' (n=21), and ``High'' (n=9) experience based on quantiles with cuts at 0.25 and 0.75 probability (R \emph{stats::quantile()} function).
Within the study level, users were shown batches of 10 images, randomly selected from a pool of 1,511 images, i.e. the ISIC 2018 Task 3 test set, with a predefined diagnosis distribution (AKIEC: 1, BCC: 1, BKL: 1, DF: 1, VASC: 1, MEL: 2, NV: 3). For each image a user had to choose one diagnosis out of seven options, and subsequently again after assistance from our foundation model, presented as multi-class probabilities visualized as bars and numbers for each class. Readers had the flexibility to complete multiple survey rounds with different image batches at their discretion, incompletely answered batches were omitted. The study was conducted online from August 20 to September 12, 2024, during which we collected data from 41 raters. Our foundation model for decision support utilized a weighted random sampler strategy, following the approach from \cite{humanai} but excluding test-time augmentation. The model demonstrated robust performance, achieving an 80.4\% mean (macro-averaged) recall, with notably high recall rates for critical skin lesions: 87.2\% for melanoma and 86.0\% for Basal Cell Carcinoma (BCC).

\hheading{\textcolor{black}{Human-AI collaboration for 128 skin conditions differential diagnosis (Reader study 3).}}
\noindent
\textcolor{black}{The reader study was conducted using a web-based platform developed for online dermatological assessment. A total of 37 healthcare professionals participated in the study, categorized into two groups based on specialization: a \textbf{dermatology group} (n=20) comprising 9 dermatology specialists and 11 specialty trainees, and a \textbf{generalist group} (n=17) including 7 GPs, 7 general medicine practitioners, and 3 other healthcare professionals (nursing, clinical trial assistants) who manage skin conditions within their broader practice scope. This grouping strategy reflects the real-world clinical setting where non-dermatologist healthcare professionals routinely perform initial skin assessments. The diverse range of 128 skin conditions enabled evaluation of diagnostic performance between dermatologically trained professionals and those with general medical training.
Readers were presented with clinical images and asked to provide their assessment through a structured questionnaire. Each participant rated image quality on a 5-point scale (from "Not at all" to "Completely" assessable), provided a primary diagnosis through free-text entry, and optionally listed two differential diagnoses ranked by likelihood. Diagnostic confidence was recorded on a 4-point scale (1: Not at all confident, 2: Somewhat confident, 3: Confident, 4: Highly confident). Following their initial assessment, readers were shown PanDerm's top three predicted diagnoses and given the opportunity to maintain or modify their original diagnosis and differential diagnoses, followed by a reassessment of their confidence using the same 4-point scale. The study collected 1,342 responses between 07/01/2025 and 10/02/2025.
Prior to evaluation, four experienced dermatologists collaboratively developed a standard ontology tree to systematically categorize the 128 skin conditions and facilitate expert evaluation (\textbf{Extended Data Fig~\ref{supp_fig_ontology_sunburst_plot}}). The evaluation process involved multiple expert assessors who independently scored diagnostic accuracy using a 4-point scale: 4 (direct match with pre-defined term in the ontology), 3 (match within same diagnostic category in the ontology), 2 (inconsequential misdiagnosis), and 1 (significant mismatch, potentially dangerous misdiagnosis). To ensure robust assessment, each case was evaluated by three assessors, with cases showing significant scoring discordance (differences between 3/4 vs 1/2) reviewed in consensus meetings to establish final scores. For the top-3 accuracy evaluation, both human readers and AI assistance were evaluated based on whether the correct diagnosis appeared within their top three diagnostic choices.}

\hheading{Evaluation metrics.}
For multi-class tasks, we primarily use a weighted F1 score, which averages class-specific F1 scores (harmonic means of precision and recall) weighted by class size. It addresses class imbalance in multi-class scenarios. For binary classification, we primarily use AUROC (Area Under the Receiver Operating Characteristic curve), measuring the model's ability to distinguish between classes across all classification thresholds. An AUROC of 1.0 indicates perfect classification, while 0.5 suggests random guessing. This metric is particularly useful for imbalanced datasets and when we need to evaluate trade-offs between true positive and false positive rates. For the \textcolor{black}{three} reader studies, we report accuracy (\textcolor{black}{top-1 or top-3}). In skin lesion segmentation, we use the Dice Similarity Coefficient (DSC) and Jaccard index (JAC) to assess segmentation quality. For TBP-based melanoma screening, we primarily report the sensitivity (recall) in malignant lesions, focusing on the model's ability to correctly identify malignant cases.

\hheading{Statistical analysis.} For skin tumor patch classification, melanoma slides classification, reader studies, metastasis prediction, and skin lesion segmentation, we conduct k-fold cross-validation due to either a relatively small sample size or following conventional practice. We compute the mean and standard deviation of performance across the folds, then calculate the standard error by dividing the standard deviation by the square root of the number of folds. The 95\% confidence interval is derived using 1.96 times the standard error. To assess statistical significance, we conduct two-sided t-tests comparing PanDerm's performance against the baseline model for each task. For the remaining datasets, we utilize nonparametric bootstrapping with 1,000 replicates to estimate 95\% confidence intervals for each model's performance. To compare models, we implement pairwise permutation tests, conducting 1,000 permutations per pair and recalculating performance metrics after each permutation. We derive two-sided \textit{P} values to evaluate the null hypothesis that paired observations stem from identical distributions. Additionally, we perform t-tests to assess the statistical significance of inter-model performance variations. Our null hypothesis posits no discernible difference between PanDerm's performance and that of its competitors. A \textit{P} value $<$ 0.05 was regarded as statistically significant.

%% file: supplement/2-datasets.tex
\heading{Skin cancer and general skin condition classification datasets}

\hheading{HAM10000 \cite{ham10000} (7 classes)}:
This dataset contains 10,015 dermoscopic images across 7 classes: actinic keratoses, basal cell carcinoma, benign keratosis, dermatofibroma, melanocytic nevi, melanoma, and vascular lesions. It is stratified into 60\% training, 20\% validation, and 20\% test sets. For human-AI collaboration, we used the official dataset. All other experiments used the clean version from \cite{clean}, which prevents data leakage by ensuring lesions from the same patient are not split across sets.

\hheading{BCN20000 \cite{bcn} (9 classes)}:
This dataset comprises 12,413 dermoscopic images in 9 categories: nevus, melanoma, basal cell carcinoma, seborrheic keratosis, actinic keratosis, solar lentigo, squamous cell carcinoma, dermatofibroma, and vascular lesions, including lesions in hard-to-diagnose locations. It is similarly stratified (60/20/20 split). We used the clean version of BCN20000, which, like the HAM10000, addresses data leakage issues.

\hheading{MSKCC \cite{isic} (2 classes)}:
The dataset is curated from the MSKCC data from ISIC archive \cite{isic}, containing 8,984 dermoscopic images with melanoma and other classes.

\hheading{HIBA \cite{isic} (2 classes)}:
The dataset is curated from the HIBA data from ISIC archive \cite{isic}, containing 1,635 dermoscopic images with melanoma and other classes.

\hheading{PAD-UFES-20 \cite{pad} (6 classes)}: 
This dataset from Brazil contains 2,298 close-up clinical images with 6 classes, including Actinic Keratosis, Basal Cell Carcinoma of skin, Malignant Melanoma, Melanocytic Nevus of Skin, Squamous Cell Carcinoma, Seborrheic Keratosis.

\hheading{DDI \cite{ddi} (2 classes)}:
We grouped the classes of the DDI dataset into melanoma and others. The dataset contains 647 clinical images from the US.

\hheading{Derm7pt \cite{dermc} (2 classes)}: 
Derm\_D is a subset of Derm7pt, containing 839 dermoscopic images and Derm\_C contains 839 clinical images with melanoma and other classes.

\hheading{ISIC2024 \cite{isic24} (2 classes)}: 
ISIC2024 is a multi-center dataset with skin lesion crops from total body photography (TBP). We chose a hold-out data with 49,025 crop images with three institutions (FNQH Cairns, Alfred Hospital, Melanoma Institute Australia) as the evaluation dataset.

\hheading{PH2 \cite{ph2} (3 classes)}:
PH2 is a clinical image dataset from Portugal with 200 images and 3 classes. We reorganize it to a binary melanoma detection task.

\hheading{Med-Node \cite{mednode} (2 classes)}: 
The dataset contains 170 clinical images. We reorganize it to a binary melanoma detection task.

\hheading{DermNet \cite{Dermnet} (23 classes)}: 
DermNet contains 19,559 clinical images, the dataset consists of images of 23 types of skin diseases, \textcolor{black}{captures common clinical presentations including inflammatory conditions (eczema, psoriasis), infections (bacterial, viral, fungal), and neoplastic diseases.}

\hheading{Fitzpatrick17K \cite{f17k} (114 classes)}: \textcolor{black}{This dataset comprises 16,577 clinical images annotated with both dermatological diagnoses and Fitzpatrick skin types (I-VI). It encompasses 114 distinct conditions (minimum 53 images per condition) spanning major dermatological categories: inflammatory dermatoses (psoriasis, lichen planus, various eczematous conditions), cutaneous malignancies (melanoma, morpheiform and solid-cystic variants of BCC, SCC), papulosquamous disorders (pityriasis rosea, pityriasis rubra pilaris), autoimmune conditions (lupus erythematosus, bullous diseases), benign neoplasms (seborrheic keratosis, dermatofibroma), and various other clinically significant entities (acanthosis nigricans, granuloma annulare, necrobiosis lipoidica).}

\hheading{MMT-09 (9 classes)}: 
The dataset is an in-house clinical dataset with 9 skin condition classes, including benign keratinocytic, malignant keratinocytic, melanocytic, inflammatory conditions and benign tumors, vascular lesion, basal cell carcinoma, malignant keratinocytic, melanoma, squamous cell carcinoma. We chose 38,476 images as our evaluation dataset.

\hheading{MMT-74 (74 classes)}: 
\textcolor{black}{The MMT-74 dataset is a comprehensive in-house clinical collection comprising 38,476 dermatological images across 74 detailed skin condition classes, building upon and refining the broader 9-class structure of MMT-09. This structured dataset encompasses diverse dermatological conditions, including detailed classifications of basal cell carcinoma variants (nodular, pigmented, superficial, and recurrent), melanocytic lesions with specific pattern recognition (such as acral patterns and various nevus types), inflammatory disorders (dermatitis, psoriasis), benign proliferations (including seborrheic keratosis variants), and vascular lesions (angiomas, telangiectasias). The dataset was specifically designed to evaluate deep learning models' performance across a diverse and clinically relevant range of skin conditions, with categories spanning inflammatory, infective, benign proliferations, melanocytic, and eczema classifications.}

\hheading{SD-128 (128 classes)}: \textcolor{black}{This dataset encompasses 5,619 clinical images covering 128 dermatological conditions spanning the complete spectrum of clinical practice. The dataset provides substantial coverage of inflammatory dermatoses, ranging from common presentations (such as psoriasis and atopic dermatitis) to less common entities (like leukocytoclastic vasculitis). It includes diverse infectious diseases of bacterial, viral, and fungal etiologies, as well as a comprehensive range of proliferative lesions from benign nevi to malignant melanomas. The collection also extends to appendageal disorders, physical trauma-related changes, nail disorders, and hair loss conditions. This extensive compilation represents both frequently encountered conditions in everyday practice and challenging rare cases, providing a robust resource for clinical diagnostic support. This dataset contains 5,619 clinical images encompassing diverse dermatological conditions commonly encountered in clinical practice. The dataset provides substantial coverage of inflammatory conditions from common presentations (psoriasis, atopic dermatitis) to less common entities (leukocytoclastic vasculitis), various infectious diseases spanning bacterial, viral, and fungal etiologies, a range of proliferative lesions from benign nevi to malignant melanomas, as well as appendageal disorders and physical trauma-related changes. We utilized 10\% of the data stratified by disease labels for benchmark evaluation. Additionally, we selected 200 images stratified by disease classes for our reader study., which were further annotated with the Fitzpatrick skin tone scale (I-VI)  by three expert dermatologists to enable subgroup performance analysis.}

\hheading{Skin Tumor Patch Classification (PATCH16) (16 classes) \cite{kriegsmann2022deep}}: The skin tumor patch classification task consists of tissue patches of 378 histopathology WSIs from the archive of the Institute of Pathology, Heidelberg University, the MVZ for Histology, Cytology and Molecular Diagnostics Trier and the Institute for Dermatopathology Hannover for classification of 16 categories including 4 tumor types and 12 normal tissue structures. We obtained a total of 129,364 image patches of 100 $\times$ 100 $\mu$m (395 $\times$ 395) size. The dataset was stratified by label, with 55\% allocated for training, 15\% for validation, and 30\% for testing.

\hheading{Melanoma Slide Classification (WSI) (2 classes)\cite{clark2013cancer}}: The melanoma slide classification task from the National Cancer Institute's Clinical Proteomic Tumor Analysis Consortium Cutaneous Melanoma (CPTAC-CM) cohort consists of histopathology WSIs for cancer detection. After selecting labeled WSIs, we obtained 302 slides (71 normal, 231 tumor). For training and evaluation, we employed a five-fold cross-validation strategy with label-stratified splits to maintain class balance.

\hheading{Early melanoma detection based on SDDI-Alfred (2 classes)}: \textcolor{black}{The dataset consists of 179 serial dermoscopic imaging sequences from 122 patients, totaling 730 dermoscopic images. The patients were recruited from a private specialist dermatology clinic, with follow-up periods ranging from January 2007 to December 2019. The study population showed distinct characteristics between melanoma and benign groups: melanoma patients had a mean age of 56.6 years (SD=11.8) compared to 49.6 years (SD=11.4) in the benign group, with slightly different gender distributions (53.9\% male in melanoma vs 40.0\% male in benign cases). Both melanoma and benign lesions that underwent short- or long-term sequential digital dermoscopic imaging (SDDI) at least once prior to biopsy were included. The dataset is well-balanced, with 90 benign lesions and 89 malignant lesions. Of the 89 melanomas, 34 (38.2\%) were invasive, with a mean Breslow thickness of 0.5 mm, while 55 (61.8\%) were in situ. The melanoma subtypes included invasive SSM (36.0\%), in situ SSM (31.4\%), unspecified in situ (18.0\%), lentigo maligna (12.3\%), and invasive LMM (2.2\%). The benign lesions were predominantly dysplastic naevi (40.0\%), followed by compound naevi (27.8\%), junctional naevi (18.9\%), and intradermal naevi (8.9\%). Anatomically, lesions were most commonly located on the lower limb (29.2\% melanoma, 26.7\% benign) and back (23.5\% melanoma, 25.6\% benign). All lesions were monitored via digital dermoscopy, excised due to clinical concerns, and confirmed by pathological examination. The number of images per sequence varied from 1 to 12, with an average of approximately 4 images per sequence.}

\heading{Longitudinal and melanoma metastasis datasets}

\hheading{Short-term lesion change detection based on SDDI1 \cite{isic} (2 classes)}:
This dataset is sourced from the ``repeated dermoscopic images of melanocytic lesions" by University Hospital Basel, available in the ISIC archive. It comprises 116 sequential lesions, each with a sequence length of 5, from 66 patients. The dataset is categorized into two classes for lesion change detection.

\hheading{Short-term lesion change detection based on SDDI2 (2 classes)}:
SDDI2 is an in-house dataset from the Medical University of Vienna. It contains 229 sequential dermoscopic images with a sequence length of 2. The dataset includes both binary change labels and more fine-grained malignant change labels. This dataset is also used for short-term lesion change detection.

\hheading{Melanoma metastasis prediction and prognosis based on CombinMel dataset (2 or 3 classes)}: \textcolor{black}{The CombinMel dataset encompasses 680 dermoscopic images of invasive melanoma from 370 patients recruited across 10 hospital sites in multiple countries, including Australia and 5 European nations. For large melanomas, multiple images were captured to ensure comprehensive coverage of the entire lesion area. The study population showed a relatively balanced distribution in both age (51.4\% $>$60 years, 48.6\% $\leq$60 years) and gender (47.3\% female, 53.5\% male). Regarding disease staging, the majority of cases were classified as Stage I (70.5\%), followed by Stage III (16.5\%), followed by Stage II (12.2\%), and Stage IV (0.8\%). In terms of T classification, T1a was the most common (59.2\%), followed by T2a (18.6\%) and T4b (13.2\%). Sentinel lymph node biopsy (SLNB) was not performed in most cases (71.6\%), with 10.8\% positive and 17.6\% negative results among those tested. For nodal status, N1 disease was the most common (10.8\%), followed by N2 (3.8\%) and N3 (1.8\%). Regarding metastasis status, 248 (67.0\%), of cases showed no metastasis, while 66 (17.8\%) presented with metastasis at the time of diagnosis. Additionally, 56 (15.1\%) of cases developed metastasis during the follow-up period.}

\hheading{Skin lesion segmentation based on ISIC2018 and HAM10000}: The skin lesion segmentation task is evaluated using two publicly available datasets. The ISIC2018 dataset \cite{isic2018} comprises 3,694 dermoscopic images with 2594 images for training, 100 for validation, and 1000 for testing. We follow this official dataset split for our experiments. The HAM10000 dataset \cite{ham10000} includes 10,015 dermoscopic images, each with corresponding binary segmentation labels. A randomized selection approach is adopted, with 64\% of the images used for training, 16\% for validation, and the remaining 20\% for testing.

\heading{3D total body photography datasets} 

This dataset comprises 3D total body photography (TBP) images captured using the VECTRA WB360 system (Canfield Scientific Inc., Parsippany, NJ, USA). The system employs 92 cameras to simultaneously capture cross-polarised 2D images with standardized lighting within seconds, which are then merged to create a high-fidelity 3D avatar of each patient's entire skin surface. From these 3D avatars, individual lesion tiles were exported for further analysis. \textcolor{black}{Unlike standalone clinical photographs, TBP represents a higher-order imaging modality where 2D tiles are systematically derived from comprehensive 3D reconstructions, maintaining intrinsic spatial relationships within the 3D framework. The standardized acquisition process with calibrated lighting and positioning enables systematic capture of the entire body surface with overlapping views. This provides consistent anatomical landmarks and contextual information critical for comprehensive patient-level assessment, including evaluation of skin phenotype patterns, lesion measurements, and application of the "ugly duckling" sign. The captured images undergo specific calibration and stitching processes, resulting in standardized 2D tiles that maintain consistent quality and perspective across all body regions.}

\hheading{Photodamage risk assessment datasets (3 classes)}:
This in-house dataset \cite{solar} contains image tiles (693$\times$693 pixels) created from 92 raw 2D photos, each representing approximately 10cm² of cutaneous surface. Tiles with $< 33\%$ skin surface were excluded using pixel color analysis. Manual review removed out-of-focus images, tiles with multiple body sites, or identifying features. The final dataset comprises 6,195 image tiles from MYM \cite{mym} and HOP \cite{hop} studies, labeled as low, moderate, or severe photodamage risk labeled primarily by dermatology students.

\hheading{Nevus counting datasets (2 classes)}:
This dataset, derived from the in-house MYM \cite{mym} study, contains 32,582 lesion tiles annotated as nevus or non-nevus. Three expert physicians independently labeled lesions on-screen, with consensus determined by $\geq $ 2 clinicians' agreement. A senior dermatologist manually identified naevi in-clinic using a dermatoscope, serving as the gold standard for the test set. To ensure consistency, lesions under underwear, on the scalp, or on foot soles were excluded, and only lesions $\geq $ 2 mm were considered. A minimum one-month interval was maintained between on-screen and in-clinic labeling sessions.

\hheading{Lesion risk prediction and TBP screening datasets (2 classes)}:
This dataset comprises 2,038 total body photography (TBP) scans from 480 patients, collected from the MYM and HOP studies. The raw TBP scans include nevi images and a variety of non-relevant images such as normal skin, scars, and freckles. To focus only on nevi, we applied filtering parameters based on built-in Vectra data settings: majorAxisMM $\geq$ 2, deltaLBnorm $\geq$ 4.5, out\_of\_bounds\_fraction $\leq$ 0.25, dnn\_lesion\_confidence $\geq$ 50 and  nevi\_confidence $>$ 80. This process resulted in 196,933 lesion image tiles. We stratified the data by the patient for training, validation, and testing: 360 patients for training (146,752 images), 40 patients for validation (19,483 images), and 80 patients for testing (30,698 images, including 28 malignant lesions). Of the total dataset, 216 images represent malignant lesions, with 40 confirmed melanoma cases.

\hheading{Measurements in TBP}: Alongside the image tiles, Vectra provides a range of measurements for each lesion, mainly including size, color, and shape. Our TBP screening model incorporates 32 such measurements: ``A'', ``Aext'', ``B'', ``Bext'', ``C'', ``Cext'', ``H'', ``Hext'', ``L'', ``Lext'', ``areaMM2'', ``area\_perim\_ratio'', ``color\_std\_mean'', ``deltaA'', ``deltaB'', ``deltaL'', ``deltaLB'', ``deltaLBnorm'', ``dnn\_lesion\_confidence'', ``eccentricity'', ``location\_simple'', ``majorAxisMM'', ``minorAxisMM'', ``nevi\_confidence'', ``norm\_border'', ``norm\_color'', ``perimeterMM'', ``radial\_color\_std\_max'', ``stdL'', ``stdLExt'', ``symm\_2axis'', and ``symm\_2axis\_angle''.

%% file: supplement/3-additional.tex
\heading{Computing hardware and software}\\
For self-supervised pretraining, we used 4 $\times$ 80GB NVIDIA H100 GPUs configured for multi-GPU single-node training using DistributedDataParallel (DDP) as implemented by Python (v.3.9.13), PyTorch (v.2.2.1, CUDA 11.8) and Torchvision (v.0.17.1). The CAE-v2 code is used as the codebase to develop our foundation model, which can be found in its official repository\footnote{\href{https://github.com/Atten4Vis/CAE}{github.com/Atten4Vis/CAE}}. For downstream task evaluation, all experiments were conducted on 4$ \times$ 49 GB NVIDIA 6000 Ada GPUs. We used Python (v.3.9.19), PyTorch (v.2.2.2, CUDA 11.8), and Torchvision (v.0.17.2) for finetuning tasks, and Python (v.3.10.14), PyTorch (v.2.2.2, CUDA 11.8) and Torchvision (v.0.17.2) for linear probing tasks. We used Scikit-learn (v1.2.1) for logistic regression in the linear probing setting. Implementation of other comparative pretrained models was modified based on the official configuration in their respective repositories: MAE\footnote{\href{https://github.com/facebookresearch/mae}{github.com/facebookresearch/mae}}, SL\_ImageNet\footnote{\href{https://huggingface.co/timm/vit_large_patch16_224.orig_in21k}{huggingface.co/timm/vit\_large\_patch16\_224.orig\_in21k}}, DINOv2\footnote{\href{https://github.com/facebookresearch/dinov2}{github.com/facebookresearch/dinov2}}, SwAVDerm\footnote{\href{https://github.com/shenyue-98/SwAVDerm}{github.com/shenyue-98/SwAVDerm}}, autoSMIM\footnote{\href{https://github.com/Wzhjerry/autoSMIM}{github.com/Wzhjerry/autoSMIM}}, BATFormer\footnote{\href{https://github.com/xianlin7/BATFormer}{github.com/xianlin7/BATFormer}}, MedSAM\footnote{\href{https://github.com/bowang-lab/MedSAM}{github.com/bowang-lab/MedSAM}}, ResNet50\footnote{\href{https://pytorch.org/vision/main/models/generated/torchvision.models.resnet50.html}{pytorch.org/vision/main/models/generated/torchvision.models.resnet50.html}}, MILAN\footnote{\href{https://github.com/zejiangh/MILAN}{github.com/zejiangh/MILAN}}, CLIP\footnote{\href{https://github.com/openai/CLIP}{github.com/openai/CLIP}}, BiomedCLIP\footnote{\href{https://huggingface.co/microsoft/BiomedCLIP-PubMedBERT_256-vit_base_patch16_224}{huggingface.co/microsoft/BiomedCLIP-PubMedBERT\_256-vit\_base\_patch16\_224}}, and MONET\footnote{\href{https://github.com/suinleelab/MONET/tree/main}{github.com/suinleelab/MONET/tree/main}}.

\heading{Ethics statement}\\
MYM study was approved by the Metro South Health Human Research Ethics Committee on 21 April 2016 (approval number: HREC/16/QPAH/125). Ethics approval has also been obtained from the University of Queensland Human Research Ethics Committee (approval number: 2016000554), Queensland University of Technology Human Research Ethics Committee (approval number: 1600000515) and QIMR Berghofer (approval number: P2271).
The HOP study has received Human Research Ethics Committee (HREC) approval from Metro South Health HREC (HREC/17/QPAH/816) and The University of Queensland HREC (2018000074).
The ComBineMel dataset is part of the Computer biomarkers evaluation of invasive melanoma (ComBine Mel) study. The study was approved by the Alfred Hospital Ethics Committee on 08 August 2023 (approval number: HREC/98200/Alfred-2023). The study follows the National Statement on Ethical Conduct in Human Research (2007) protocols. 
SDDI2 dataset is approved by the Ethics Review Board of the Medical University of Vienna.
MMT data study is part of a research agreement study with Monash eResearch Centre and was approved through the Monash University Human Research Ethics Committee (MUHREC).
The NSSI dataset is part of the Brisbane Naevus Morphology Study, circa 2009-2014. The study followed the Declaration of Helsinki protocols and was approved by the Princess Alexandra Hospital human research ethics committee. The ACEMID\_path study has received approval from the Alfred Hospital Ethics Committee (approval number: 746/23) to share data accrued for registered trial ACTRN12619001706167 (ACEMID) under the Metro South Human Research Committee protocol HREC/2019/QMS/57206 and the University of Queensland Human Research Ethics Committee protocol 2019003077.
The SDDI\_Alfred study has received approval from the Alfred Hospital Ethics Committee (approval number: 198/19) for the use of sequential dermoscopic imaging data. Only de-identified retrospective data was used for research, without the active involvement of patients.

\heading{Data availability}\\
Most datasets used in this study are publicly available. These datasets used for skin lesion diagnosis and segmentation tasks can be accessed through various repositories. The ISIC archive\footnote{\href{https://www.isic-archive.com/}{isic-archive.com}} hosts several datasets, including MSKCC and HIBA. Other widely used benchmark datasets are available through their respective portals: BCN20000\footnote{\href{https://figshare.com/articles/journal_contribution/BCN20000_Dermoscopic_Lesions_in_the_Wild/24140028/1}{figshare.com/articles/journal\_contribution/BCN20000\_Dermoscopic\_Lesions\_in\_the\_Wild/24140028/1}}, PAD-UFES-20\footnote{\href{https://www.kaggle.com/datasets/mahdavi1202/skin-cancer}{kaggle.com/datasets/mahdavi1202/skin-cancer}}, DDI\footnote{\href{https://ddi-dataset.github.io/index.html}{ddi-dataset.github.io/index.html}}, Derm7pt\footnote{\href{https://derm.cs.sfu.ca/Welcome.html}{derm.cs.sfu.ca/Welcome.html}}, ISIC2024\footnote{\href{https://www.kaggle.com/competitions/isic-2024-challenge}{kaggle.com/competitions/isic-2024-challenge}}, Med-Node\footnote{\href{https://www.kaggle.com/datasets/prabhavsanga/med-node}{kaggle.com/datasets/prabhavsanga/med-node}}, DermNet\footnote{\href{https://www.kaggle.com/datasets/shubhamgoel27/dermnet}{kaggle.com/datasets/shubhamgoel27/dermnet}}, WSI\footnote{\href{https://portal.gdc.cancer.gov/projects/TCGA-SKCM}{portal.gdc.cancer.gov/projects/TCGA-SKCM}}, PATCH16\footnote{\href{https://heidata.uni-heidelberg.de/dataset.xhtml?persistentId=doi:10.11588/data/7QCR8S}{heidata.uni-heidelberg.de/dataset.xhtml?persistentId=doi:10.11588/data/7QCR8S}}, ISIC2018\_task1 and HAM1000\footnote{\href{https://challenge.isic-archive.com/data/}{challenge.isic-archive.com/data/}}, SDDI1\footnote{\href{https://api.isic-archive.com/collections/328/}{api.isic-archive.com/collections/328/}}, PH2\footnote{\href{https://www.fc.up.pt/addi/ph2\%20database.html}{fc.up.pt/addi/ph2\%20database.html}}, and SD-128 \footnote{\href{https://huggingface.co/datasets/resyhgerwshshgdfghsdfgh/SD-198}{huggingface.co/datasets/resyhgerwshshgdfghsdfgh/SD-198}}.
Access to in-house datasets is restricted due to patient privacy considerations. These include MMT for dermoscopic and clinical image pretraining and downstream multi-skin condition classification, NSSI for sequential dermoscopic image pretraining, ACEMID\_path for dermatopathology pretraining, Edu1 and Edu2 for clinical image pretraining, SDDI2 for lesion change detection, SDDI\_Alfred for reader study 1 (early-melanoma detection), and the TBP data from MYM and HOP studies for all TBP-based pretraining and evaluation. Researchers interested in accessing these datasets should direct their requests to the corresponding author. Requests will be evaluated according to institutional and departmental policies to ensure compliance with intellectual property rights and patient privacy obligations. The availability of these data may be subject to additional restrictions or requirements.

\heading{Code availability}\\
We have made the encoder code and weights available for downstream task applications. They can be accessed at https://github.com/SiyuanYan1/PanDerm. We have documented all experiments in detail in our Methods section to enable independent replication. To facilitate the broader use of our model, we have provided tutorial Jupyter notebooks and downstream evaluation code suitable for a wide scientific audience. These resources have been made available to ensure transparency and to promote further research in this field. 

\heading{Author contributions}\\
S.Y., H.K., and Z.G. conceived the study and designed experiments. S.Y., C.V.A., M.H., L.Y., L.J., H.K., V.M., M.J., H.P.S., and Z.G. contributed to data acquisition, preprocessing, and organization. S.Y. performed model development and pretraining. S.Y., Z.Y., Z.W., L.Y., H.K., and P.T. contributed to downstream task evaluation. S.Y., Z.Y., C.V.A., E.M., P.G, J.B., and V.M. contributed to the metastasis prediction and prognosis tasks. S.Y. and Z.Y. performed experimental analyses regarding TBP-based screening. Z.W. and S.Y. performed experimental analysis regarding lesion segmentation. S.Y., L.Y., and P.F. contributed to dermatopathology image analysis tasks. S.Y., H.K., P.T., V.M., and H.P.S. contributed to the two human-AI collaboration reader studies. S.Y. also contributed to all remaining tasks of experimental analysis. H.K. and P.T. developed the web-based reader platforms and conducted the human-AI reader study. V.M., J.N., and Z.Y. contributed to the early melanoma detection reader study data. H.K., P.T., C.V.A., C.P., V.M., M.J., and H.P.S. contributed clinical inputs to this research. G.T., V.T., A.B.N., D.P., P.B., and S.S. provided computing resources and data management. All authors contributed to the drafting and revising of the manuscript.

\heading{Acknowledgements}\\
We thank the following readers who participated in the reader study 3: Ali Abid, Ana Brkic, Marija Buljan, Giuliana Carlos Minano, Charmaine Chamberlin, Mary ElSharouni, Jon Fulgencio, Klara Gacina, Emilia Garcia, Victor Giang, Prue Gramp, Claudia Huang, Emma Karlsen, Damian Kostner, Jack Lai, Francesco Leo, Sophie Lim, Elisabetta Magnaterra, Isabel Matheus, Rebeca Mielgo Salvador, Christoph Mueller, Nicholas Muller, Jennifer Nguyen, Maria Onteniente, Nisal Punchihewa, Polina Reztsova, Belen Rosales Trujillo, Cliff Rosendahl, Helmut Schaider, Christine Shen, Juliet Smith, Corey Stone, Tom Sun, Deborah Toledo Flores, Cathleen Ventura, Anna Wolber, and Mabel Yan.

%% file: figures/figure.tex
\setcounter{figure}{0}
\renewcommand{\figurename}{Extended Data Figure}
\vspace{2mm}
\begin{figure*}
\centering

\includegraphics[width=\textwidth]{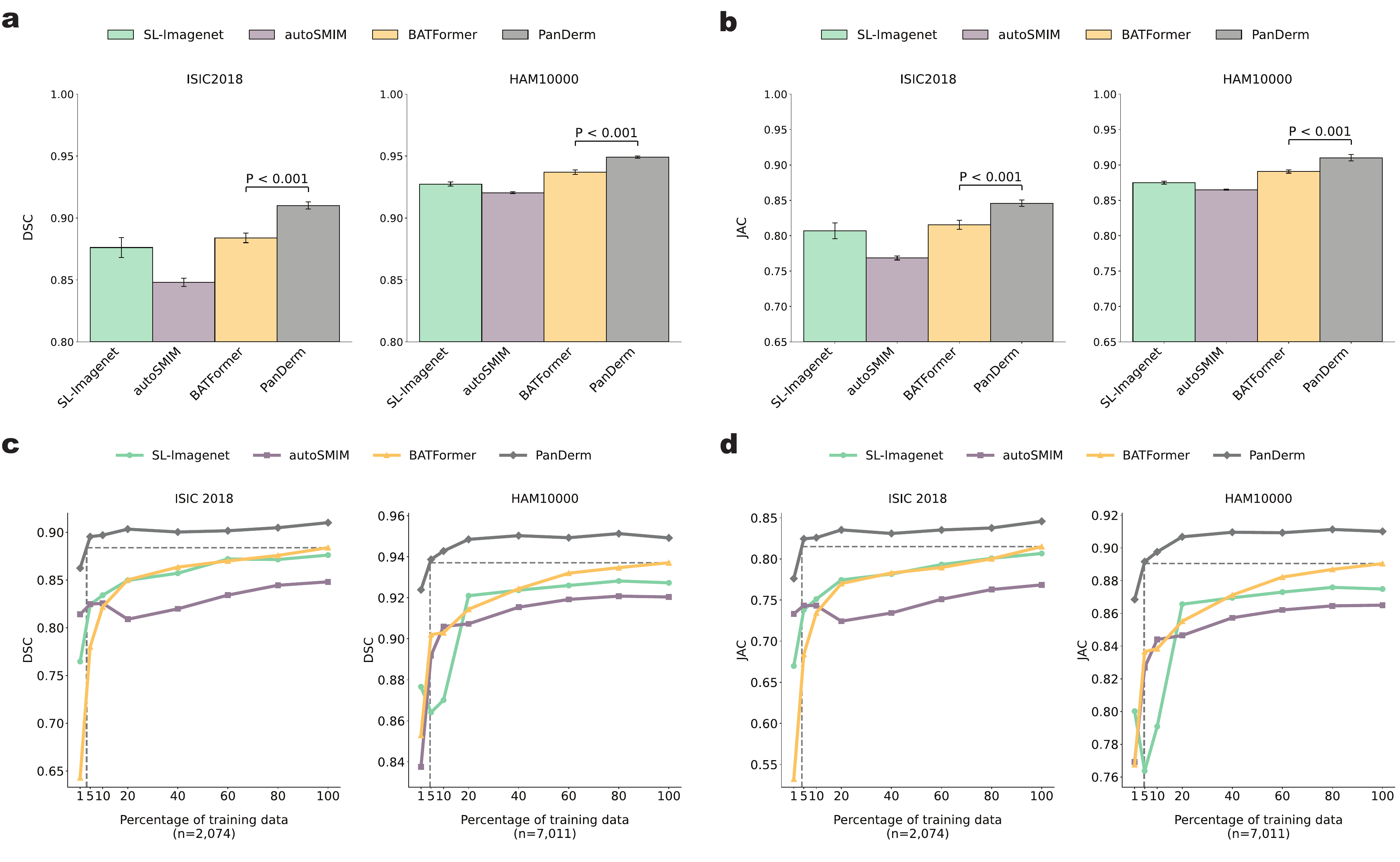}
\caption{\textbf{Quantitative skin lesion segmentation results.} \textbf{a, b.} Segmentation performance measured by dice score (DSC) and Jaccard index (JAC) for PanDerm and baseline models on ISIC2018 and HAM10000 datasets. \textbf{c, d.} Label efficiency generalization performance for PanDerm and baselines, showing mean DSC and JAC on ISIC2018 and HAM10000 datasets. Error bars in \textbf{a, b} indicate 95\% confidence intervals; bar centers represent mean values. Points in \textbf{c, d} denote mean values. All estimates are derived from five replicas with different seeds. Statistical significance was assessed using two-sided t-tests.}
\label{supp_fig1}
\end{figure*}

\begin{figure*}
\centering

\includegraphics[width=\textwidth]{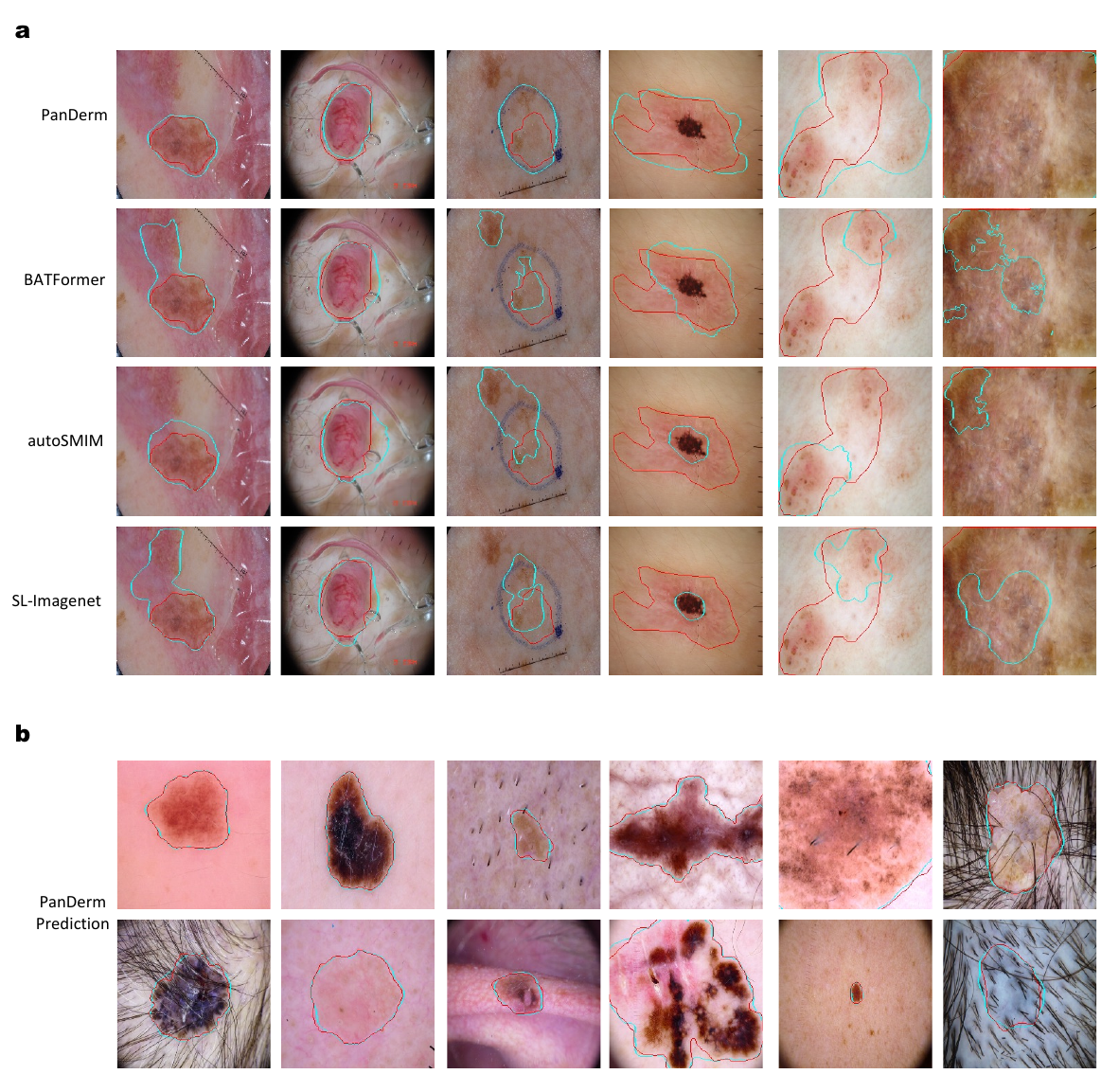}
\caption{\textbf{Qualitative skin lesion segmentation results.} \textbf{a.} Comparison of PanDerm against baseline models on challenging examples from HAM10000. Red contours indicate ground truth masks, while cyan contours show model predictions. \textbf{b.} PanDerm segmentation results on a random selection of images from HAM10000.}
\label{supp_fig2}
\end{figure*}

\begin{figure*}
\centering
\includegraphics[width=\textwidth]{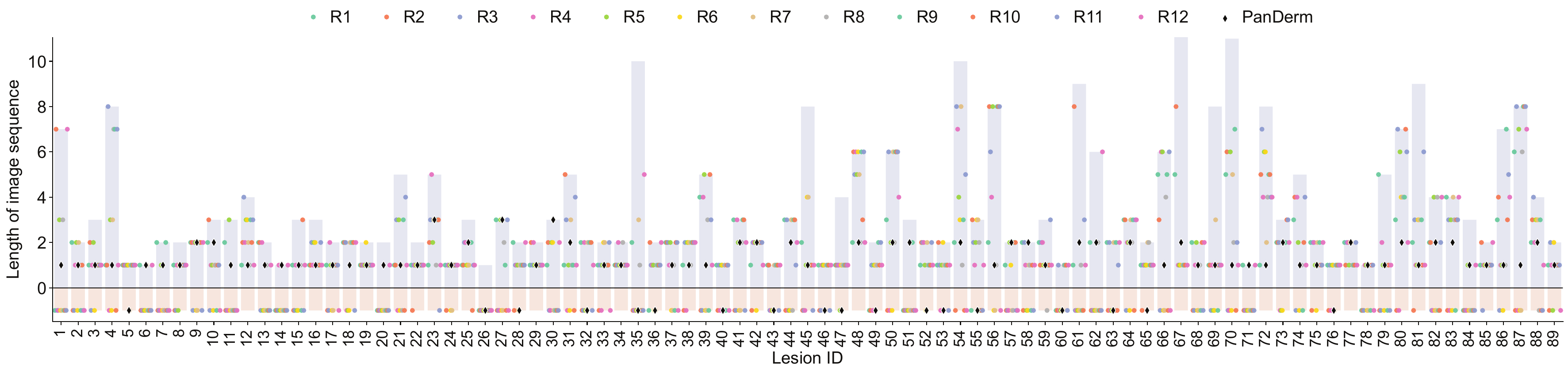}
\caption{\textbf{Early melanoma detection results (reader study 1)}: Comparing PanDerm to 12 clinicians (7 experienced dermatologists, 5 dermatology residents). X-axis: 89 melanoma lesion IDs; Y-axis: lesion image sequence length. Points on the histogram represent the initial time points of correct melanoma diagnoses. Points below y=0 correspond to melanoma lesions undetected throughout the sequence}
\label{supp_fig_r1}
\end{figure*}

\begin{figure*}
\centering
\vspace{-5mm}
\includegraphics[width=\textwidth]{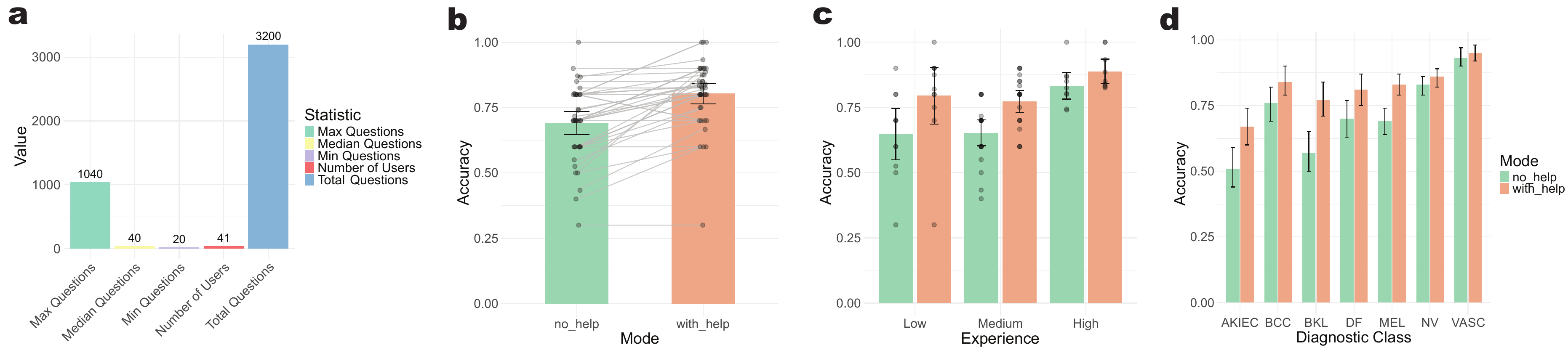}

\caption{\textbf{Performance of PanDerm in Human-AI Collaborative 7 Skin Cancer Diagnosis using Dermoscopic Images.}
\textbf{a.} Overview of the reader study: 41 users participated, answering a total of 3,200 questions. The maximum, median, and minimum number of questions answered per user were 1,040, 40, and 20, respectively.
\textbf{b.} Comparison of diagnostic accuracy without support and with PanDerm support (\textit{P} $<$ 0.001; two-sided paired t-test; n = 41 readers).
\textbf{c.} Comparison of diagnostic accuracy without support and with PanDerm support, grouped by experience level: Low (n = 11 readers), Medium (n = 21 readers), and High (n = 9 readers).
\textbf{d.} Comparison of diagnostic accuracy without support and with PanDerm support, grouped by diagnostic classes. Abbreviations: MEL, melanoma; BCC, basal cell carcinoma; AKIEC, actinic keratosis/intraepidermal carcinoma; BKL, benign keratinocytic lesion; NV, melanocytic nevus; DF, dermatofibroma; VASC, vascular lesion.}

\label{supp_fig_r2}
\end{figure*}

\begin{figure*}
\centering
\includegraphics[width=\textwidth]{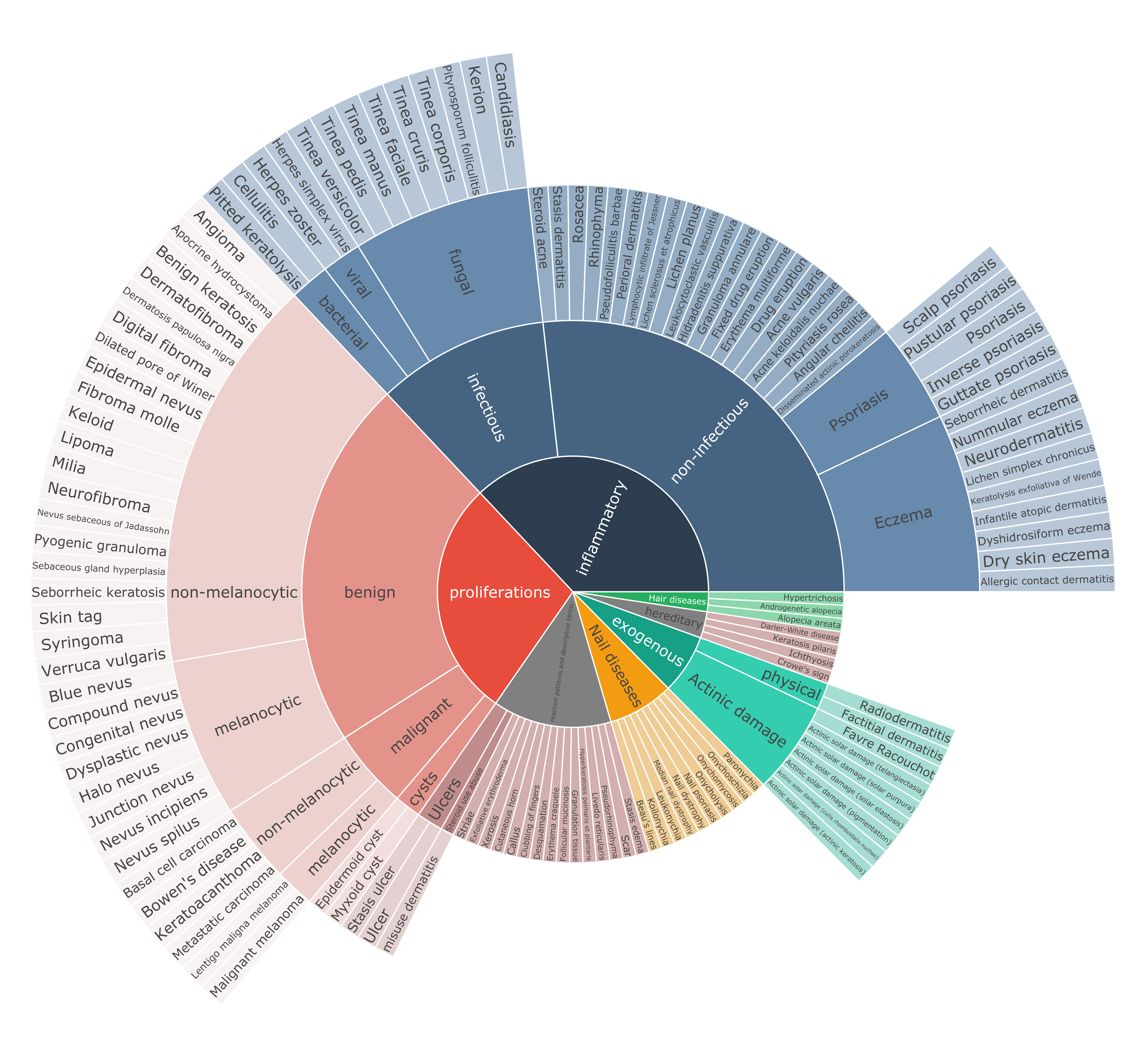}
\caption{\textcolor{black}{\textbf{Sunburst plot of standard ontology on SD-128 dataset.}}}
\label{supp_fig_ontology_sunburst_plot}
\end{figure*}

\begin{figure*}
\centering
\includegraphics[width=0.55\textwidth]{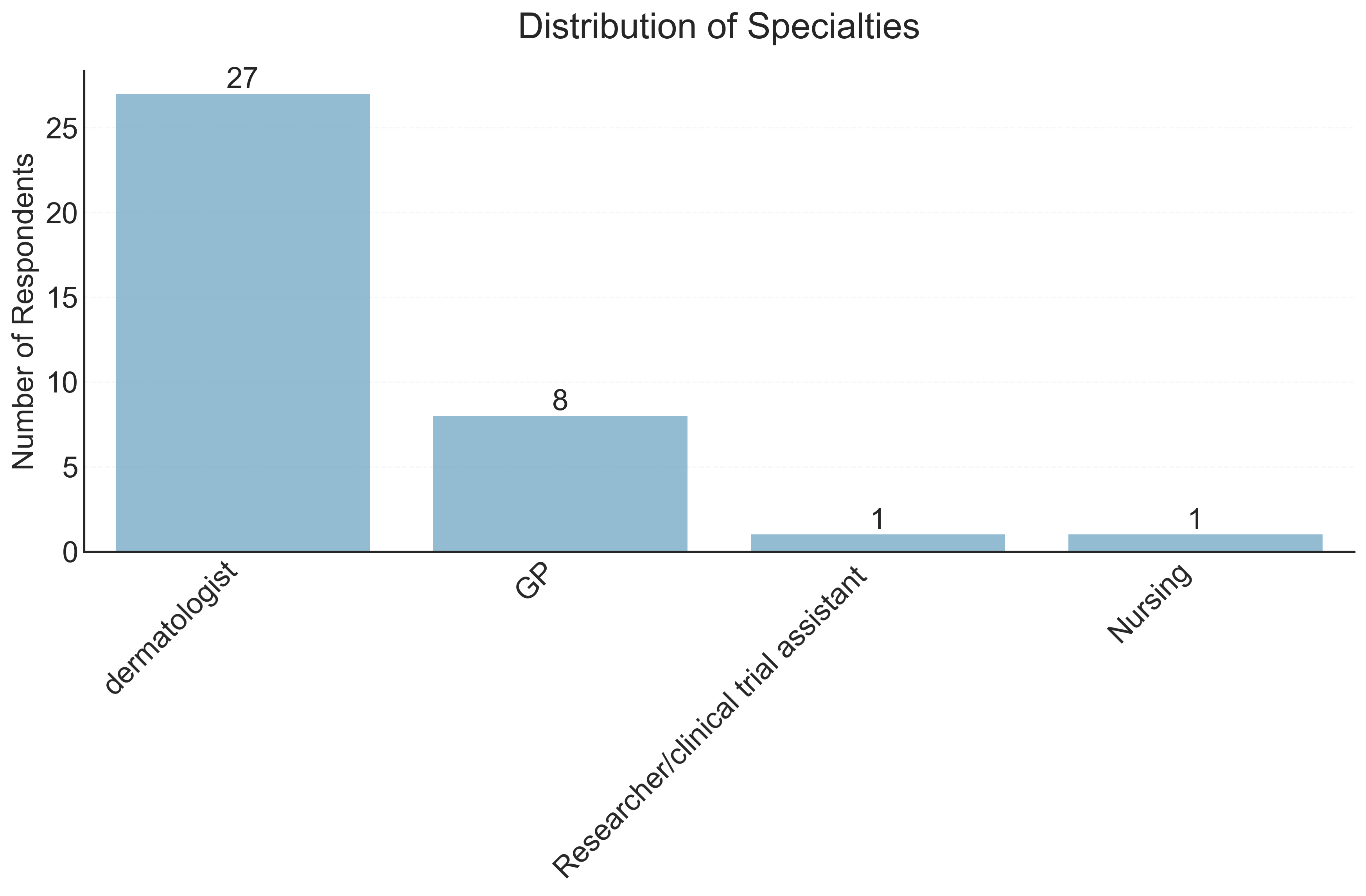}
\caption{\textcolor{black}{\textbf{Participant specialties distribution for reader study 3.}}}
\label{rr1}
\end{figure*}

\begin{figure*}
\centering
\includegraphics[width=0.55\textwidth]{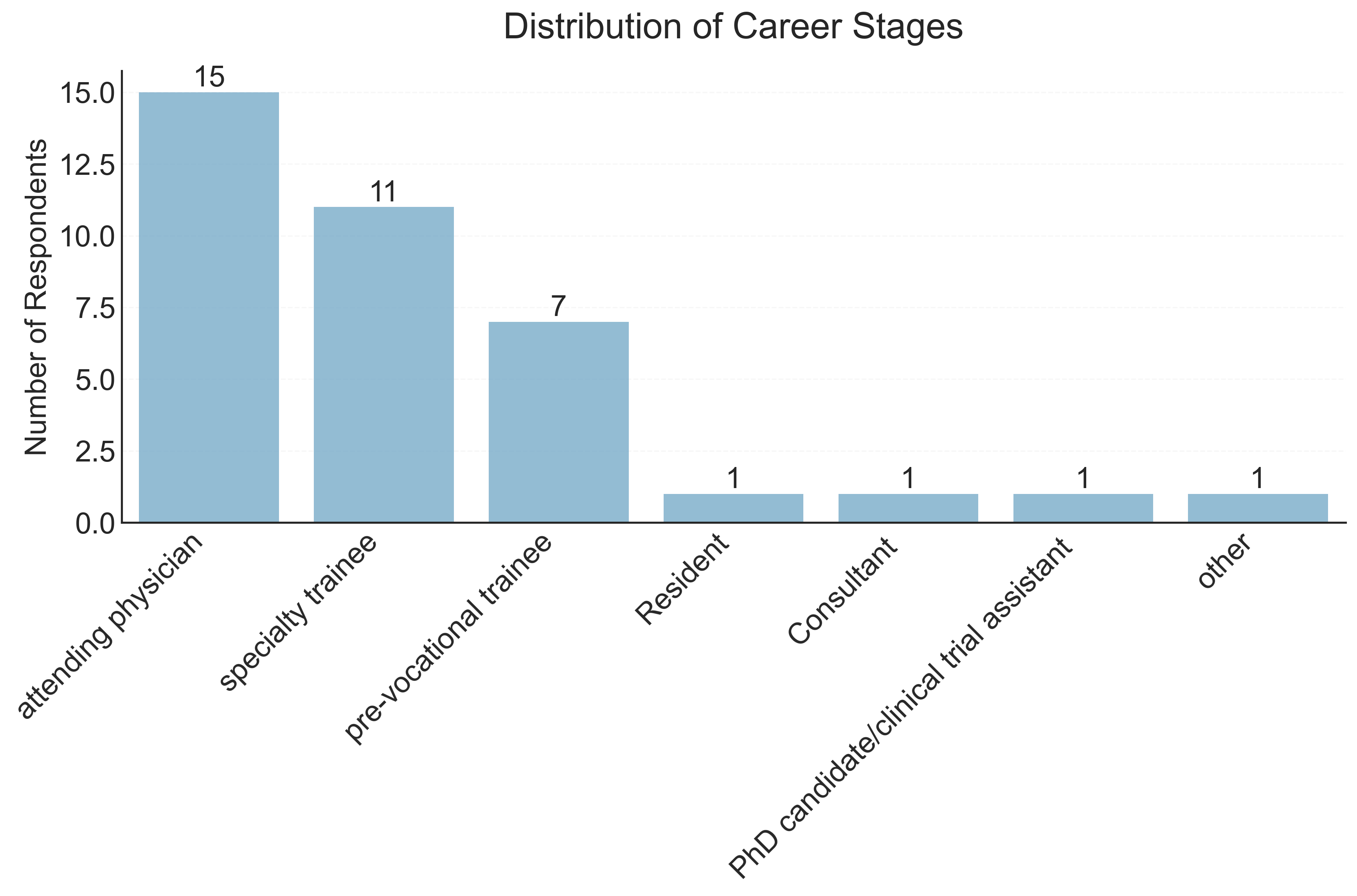}
\caption{\textcolor{black}{\textbf{Participant career distribution for reader study 3.}}}
\label{rr2}
\end{figure*}

\begin{figure*}
\centering
\includegraphics[width=0.55\textwidth]{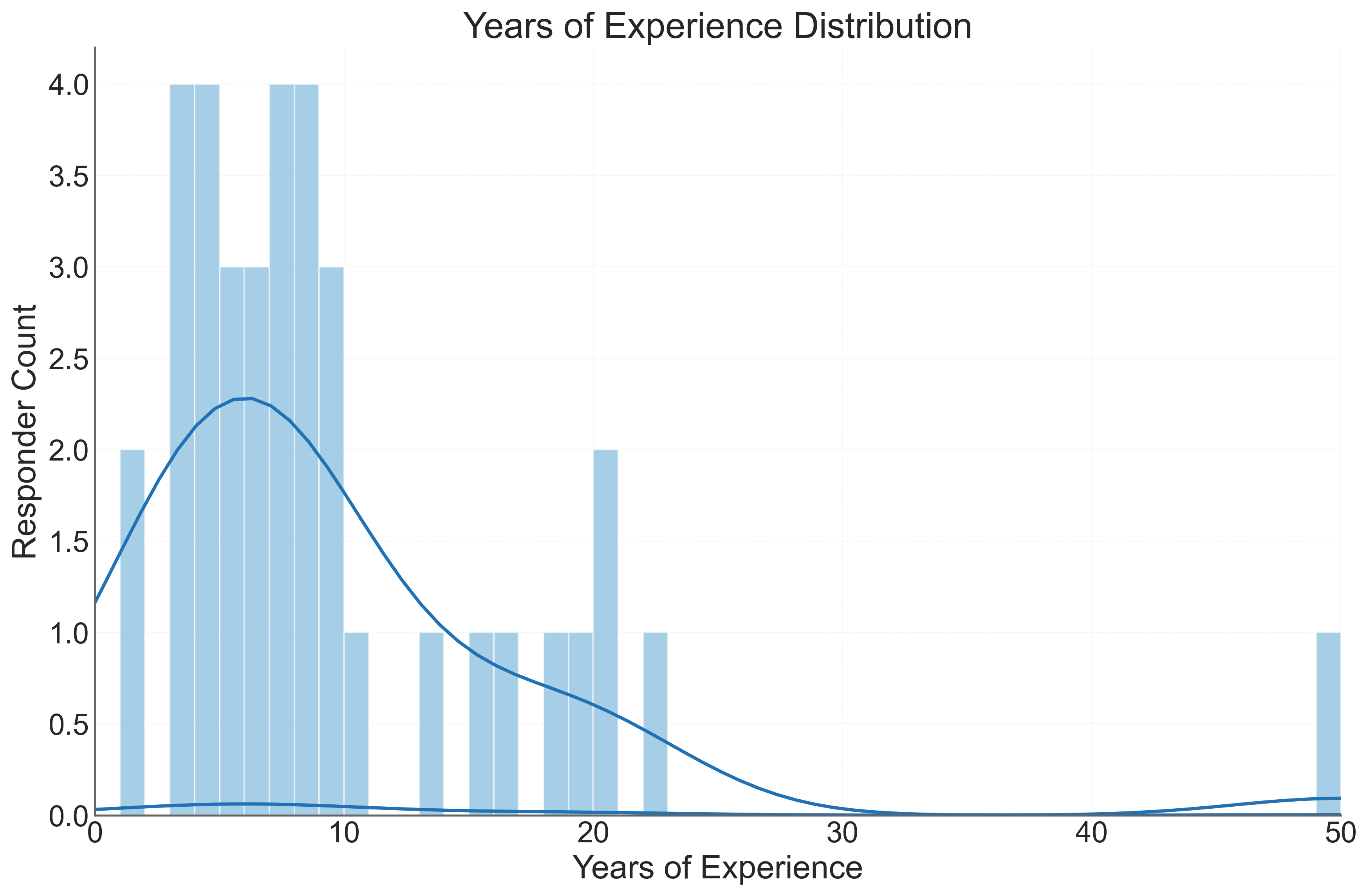}
\caption{\textcolor{black}{\textbf{Participant experience levels by years for reader study 3.}}}
\label{rr3}
\end{figure*}

\begin{figure*}
\centering

\includegraphics[width=0.95\textwidth]{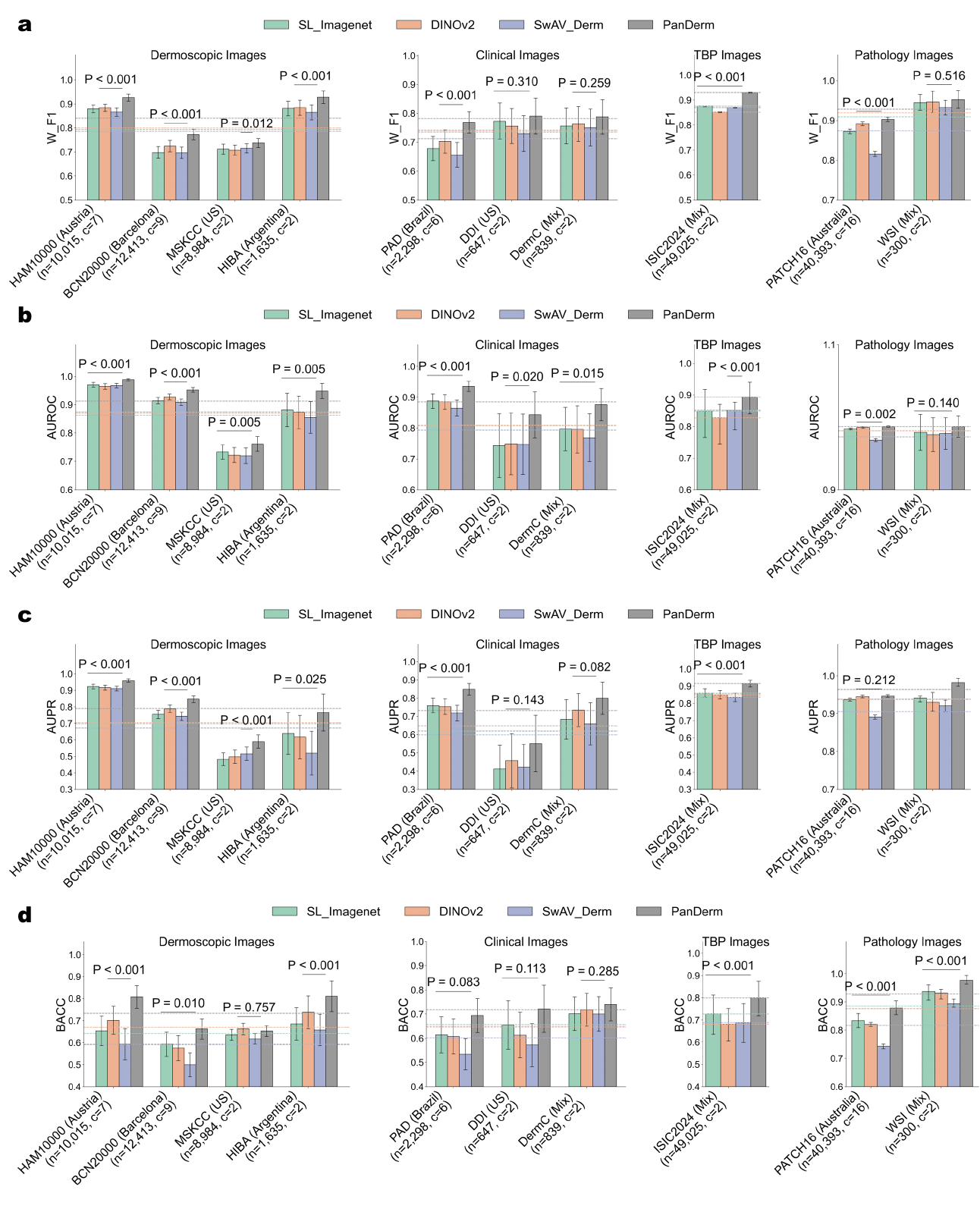}
\caption{\textbf{Performance of PanDerm versus other pretrained models on 10 pigmented skin lesion datasets across multiple centers and modalities.} \textbf{a.} Performances are measured by weighted F1 (W\_F1). \textbf{b.} Performances are measured by AUROC. \textbf{c.} Performances are measured by AUPR. \textbf{d.} Performances are measured by BACC. n: data size, c: class number. Dashed lines show the average performance of each model across different datasets.}
\label{supp_fig3}
\end{figure*}

\begin{figure*}
\centering

\includegraphics[width=\textwidth]{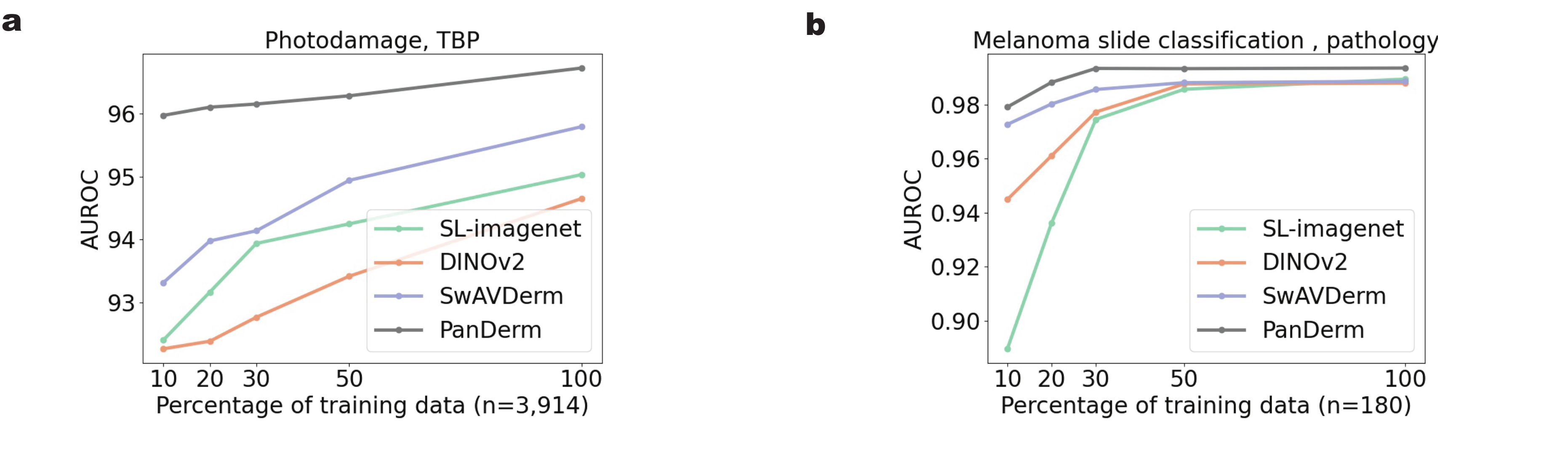}
\caption{\textbf{Label efficiency generalization results on additional tasks}. \textbf{a.} Label efficiency analysis for photodamage risk assessment using Total Body Photography (TBP) images. Results demonstrate model performance with limited labeled data available. PanDerm outperformed the second-best models using only 10\% of labeled images. \textbf{b.} Label efficiency analysis for melanoma classification using whole slide dermatopathology images. Results illustrate model performance with limited labeled data. PanDerm surpassed the second-best models using less than 30\% of labeled images.}
\label{supp_fig4}
\end{figure*}

\begin{figure*}
\centering

\includegraphics[width=0.5\textwidth]{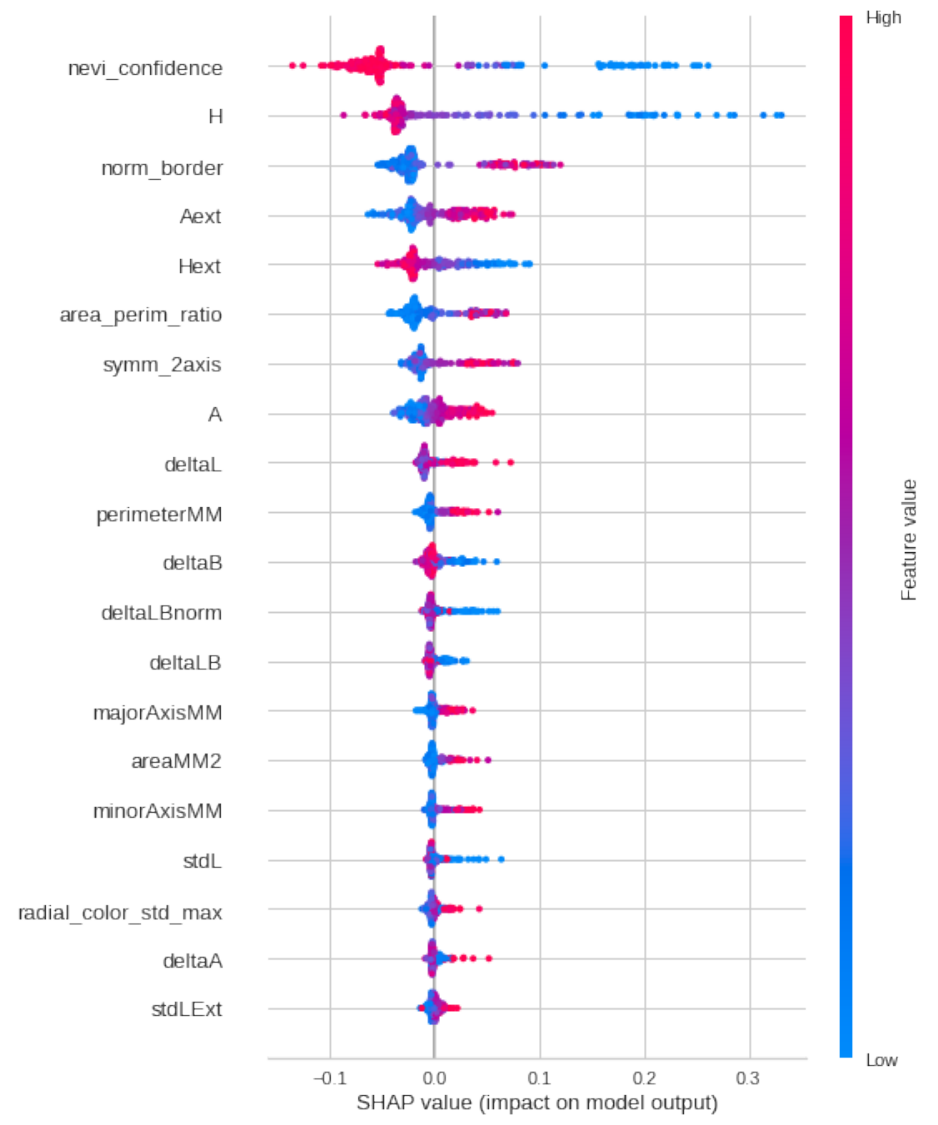}
\caption{\textbf{SHAP (SHapley Additive exPlanations) value plot}. It shows the impact of various measurement variables captured by the 3D TBP machine on the model output. The plot displays the relative importance and directional influence of each feature, with colors indicating high (red) to low (blue) feature values, and the x-axis representing the SHAP value or impact on the model's prediction. Features are ordered by their overall importance, with 'nevi\_confidence' having the highest impact and 'stdLExt' the lowest.}
\label{supp_fig5}
\end{figure*}

\begin{figure*}
\centering

\includegraphics[width=\textwidth]{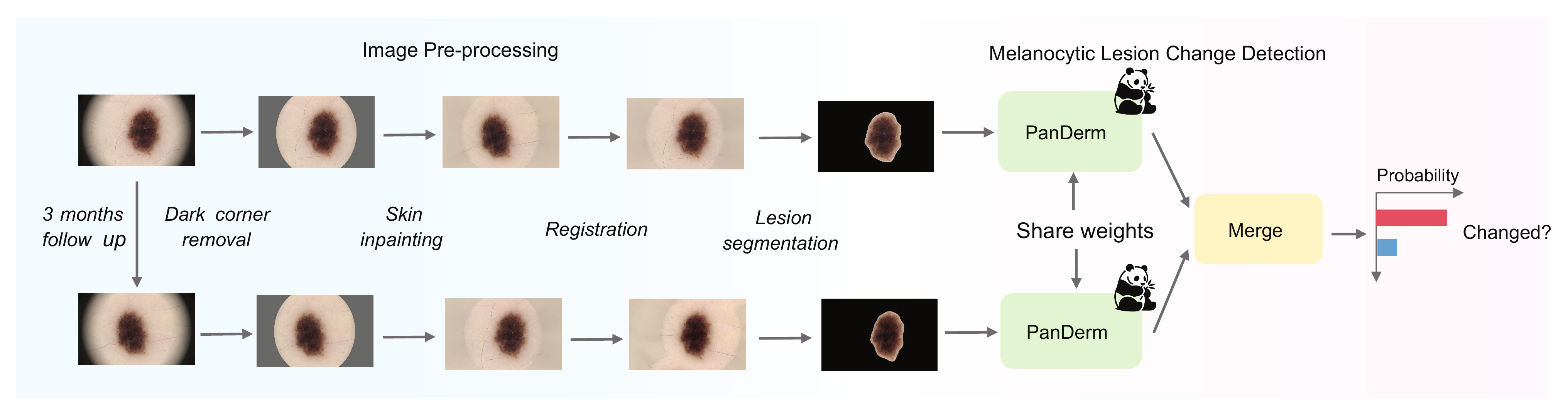}
\caption{\textbf{Longitudinal dermoscopic image-based lesion change detection using PanDerm}. For comparing subtle changes in paired lesions during short-term follow-up (e.g., 3 months), images undergo dark corner detection and removal, skin inpainting, registration, and lesion segmentation. This allows models to focus on subtle differences between lesions at different time points. }
\label{supp_change_method}
\end{figure*}

%% file: tables/table_detail_performance.tex
\begin{table}[h]
  \centering
  \renewcommand{\arraystretch}{1.2}
  \begin{tabular}{l|r}  
    \toprule
    \textbf{Characteristic} & \textbf{n (\%) or Mean ± SD} \\
    \midrule
    \textbf{Total Participants} & \\
    Total & 193 \\
    \midrule
    \textbf{Age} & \\
    Age (years) & 52.2 ± 12.0 \\
    \midrule
    \textbf{Sex} & \\
    Female & 82 (42.5\%) \\
    Male & 111 (58.5\%) \\
    \midrule
    \textbf{Ancestry/Ethnicity} & \\
    European/British & 164 (85.0\%) \\
    Mixed/Other & 29 (15.0\%) \\
    \midrule
    \textbf{Skin Phototype} & \\
    Burns easily, tans slightly & 52 (26.9\%) \\
    Burns moderately, tans gradually & 123 (63.7\%) \\
    Rarely burns, tans well & 18 (9.3\%) \\
    \midrule
    \textbf{Innate Skin Color} & \\
    Fair/Type I & 144 (74.6\%) \\
    Medium/Type II & 48 (24.9\%) \\
    Olive/Type III & 1 (0.5\%) \\

    \bottomrule
  \end{tabular}
  \caption{ \textcolor{black}{Demographic characteristics of MYM cohort.}}
  \label{tab2:demographics}
\end{table}
\begin{table}[h]
  \centering
  \renewcommand{\arraystretch}{1.2}
  \begin{tabular}{l|r}  
    \toprule
    \textbf{Characteristic} & \textbf{N (\%) or Mean ± SD} \\
    \midrule
    \textbf{Sample Size} & \\
    Total & 314 \\
    \midrule
    \textbf{Age} & \\
    Age (years) & 56.1 ± 12.8 \\
    \midrule
    \textbf{Sex} & \\
    Male & 120 (38.2\%) \\
    Female & 194 (61.8\%) \\
    \midrule
    \textbf{Ancestry} & \\
    European/British & 297 (94.6\%) \\
    Other/Mixed & 17 (5.4\%) \\
    \midrule
    \textbf{Birth Place} & \\
    Australia & 268 (85.4\%) \\
    Overseas & 46 (14.6\%) \\
    \midrule
    \textbf{Melanoma History} & \\
    Yes & 304 (96.8\%) \\
    No & 10 (3.2\%) \\
    \midrule
    \textbf{Non-melanoma Skin Cancer} & \\
    Yes & 195 (62.1\%) \\
    No & 119 (37.9\%) \\
    \midrule
    \textbf{Innate Skin Color} & \\
    Fair/Type I & 269 (85.7\%) \\
    Medium/Type II & 44 (14.0\%) \\
    Olive/Type III & 1 (0.3\%) \\
    \bottomrule
  \end{tabular}
  \caption{\textcolor{black}{Demographic and clinical characteristics of HOP cohort.}}
  \label{tab2:demographics}
\end{table}

\begin{table}[h]
 \centering
 \renewcommand{\arraystretch}{1.2}
 \begin{tabular}{l|r}
   \toprule
   \textbf{Characteristic} & \textbf{N (\%)} \\
   \midrule
   \textbf{Age} & \\
   $>$ 60 years & 190 (51.4) \\
   $\leq$ 60 years & 180 (48.6) \\
   \midrule
   \textbf{Sex} & \\
   Female & 172 (47.3) \\
   Male & 198 (53.5) \\
   \midrule
   \textbf{Stage at Diagnosis} & \\
   Stage I & 261 (70.5) \\
   Stage II & 45 (12.2) \\
   Stage III & 61 (16.5) \\
   Stage IV & 3 (0.8) \\
   \midrule
   \textbf{T Classification} & \\
   T1a & 193 (59.2) \\
   T1b & 28 (8.6) \\
   T2a & 56 (18.6) \\
   T2b & 13 (4.0) \\
   T3a & 21 (6.4) \\
   T3b & 11 (3.4) \\
   T4a & 7 (2.1) \\
   T4b & 42 (13.2) \\
   Unknown & 1 (0.3) \\
   \midrule
   \textbf{SLNB Status} & \\
   Not Performed & 265 (71.6) \\
   Positive & 40 (10.8) \\
   Negative & 65 (17.6) \\
   \midrule
   \textbf{N Classification} & \\
   N1 & 40 (10.8) \\
   N2 & 14 (3.8) \\
   N3 & 7 (1.8) \\
   Unknown & 2 (0.5) \\
   \bottomrule
 \end{tabular}
 \caption{\textcolor{black}{Demographic characteristics of the CombinMel dataset.}}
 \label{tab3:clinical_characteristics}
\end{table}

\begin{table}[h]
 \centering
 \small  
 \renewcommand{\arraystretch}{1.1}  
 \begin{tabular}{l|r|r}
   \toprule
   \textbf{Characteristic} & \textbf{Melanoma N (\%)} & \textbf{Benign N (\%)} \\
   \midrule
   \textbf{Gender} & & \\
   Male & 48 (53.9) & 36 (40.0) \\
   Female & 41 (46.1) & 54 (60.0) \\
   \midrule
   Previous melanoma & 67 (75.2) & 57 (63.3) \\
   Previous NMSC & 44 (49.4) & 20 (22.2) \\
   Family history of melanoma & 49 (55.0) & 38 (42.2) \\
   \midrule
   \textbf{Naevi count} & & \\
   \textless20 & 0 (0.0) & 0 (0.0) \\
   20-50 & 3 (3.4) & 1 (1.1) \\
   50-100 & 27 (30.3) & 15 (16.7) \\
   100-200 & 14 (15.7) & 33 (36.7) \\
   200-500 & 42 (47.2) & 39 (43.3) \\
   \textgreater500 & 3 (3.4) & 2 (2.2) \\
   \midrule
   \textbf{Atypical Naevi count} & & \\
   \textless5 & 19 (21.3) & 22 (24.4) \\
   5-15 & 28 (31.5) & 29 (32.2) \\
   \textgreater15 & 42 (47.2) & 39 (43.0) \\
   \midrule
   \textbf{Fitzpatrick Skin Phototype} & & \\
   I & 12 (13.4) & 11 (12.2) \\
   II & 42 (47.1) & 47 (52.2) \\
   III & 28 (31.4) & 32 (35.6) \\
   IV & 0 (0.0) & 0 (0.0) \\
   V & 7 (7.8) & 0 (0.0) \\
   \midrule
   \textbf{Location} & & \\
   Head and neck & 7 (7.8) & 7 (7.8) \\
   Chest & 9 (10.1) & 9 (10.0) \\
   Abdomen & 7 (7.8) & 14 (15.6) \\
   Back & 21 (23.5) & 23 (25.6) \\
   Upper limb & 19 (21.3) & 13 (14.4) \\
   Lower limb & 26 (29.2) & 24 (26.7) \\
   \midrule
   \textbf{Histopathology (Melanoma)} & & \\
   In situ unspecified & 16 (18.0) & -- \\
   In situ SSM & 28 (31.4) & -- \\
   Lentigo maligna & 11 (12.3) & -- \\
   Invasive SSM & 32 (36.0) & -- \\
   Invasive LMM & 2 (2.2) & -- \\
   \midrule
   \textbf{Histopathology (Benign)} & & \\
   Dysplastic naevus & -- & 36 (40.0) \\
   Compound naevus & -- & 25 (27.8) \\
   Junctional naevus & -- & 17 (18.9) \\
   Intradermal naevus & -- & 8 (8.9) \\
   Other & -- & 4 (4.4) \\
   \bottomrule
 \end{tabular}
 \caption{\textcolor{black}{Demographic characteristics of the SDDI\_Alfred dataset.}}
 \label{tab:sddi_alfred_demographics}
\end{table}

\begin{table}[h]
 \centering
 \small
 \renewcommand{\arraystretch}{1.1}
 \begin{tabular}{l|r}
   \toprule
   \textbf{Characteristic} & \textbf{N (\%)} \\
   \midrule
   \textbf{Total Participants} & 1254 \\
   \midrule
   \textbf{Age (years)} & \\
   Mean & 47.22 \\
   Median & 46 \\
   Range & 11-88 \\
   \midrule
   \textbf{Sex} & \\
   Male & 602 (48.0) \\
   Female & 652 (52.0) \\
   \midrule
   \textbf{Melanoma History} & \\
   Personal history of melanoma & 589 (47.0) \\
   No personal history of melanoma & 665 (53.0) \\
   \midrule
   \textbf{Facultative Skin Colour (n=1245)} & \\
   Fair/Type 1 & 635 (51.0) \\
   Medium/Type 2 & 510 (41.0) \\
   Olive/Type 3 & 100 (8.0) \\
   \midrule
   \textbf{Self-reported Ancestry*} & \\
   British/Irish & 963 \\
   West/Northern European & 228 \\
   No data or other (not specified) & 104 \\
   Southern European & 36 \\
   Eastern European & 23 \\
   Asian & 18 \\
   Middle Eastern & 8 \\
   Other (specified) & 8 \\
   Indigenous Australian & 4 \\
   Pacific Islander & 4 \\
   \bottomrule
 \end{tabular}
 \caption{\textcolor{black}{Demographic characteristics of the NSSI dataset.}}
 \label{tab:nssi_demographics}
 
 \small{* Multiple ancestries could be reported by each participant}
\end{table}

\begin{table}[h]
 \centering
 \small
 \renewcommand{\arraystretch}{1.1}
 \begin{tabular}{l|r}
   \toprule
   \textbf{Characteristic} & \textbf{N (\%)} \\
   \midrule
   \textbf{Total Participants} & 54 \\
   \midrule
   \textbf{Study Sites} & \\
   QLD 1 - PAH & 26 (48.1) \\
   NSW 3 - MIA & 28 (51.9) \\
   \midrule
   \textbf{Age (years)} & \\
   Range & 19-75 \\
   Mean & 53.4 \\
   \midrule
   \textbf{Gender} & \\
   Male & 31 (57.4) \\
   Female & 23 (42.6) \\
   \midrule
   \textbf{Melanoma Risk Group} & \\
   Very High & 44 (81.5) \\
   High & 8 (14.8) \\
   Low/Average & 1 (1.9) \\
   Unknown & 1 (1.9) \\
   \midrule
   \textbf{Primary Diagnosis (Pathologist 1)} & \\
   Total Naevi & 37 (68.5) \\
   \quad Common/Dermal/Congenital & 14 (25.9) \\
   \quad Dysplastic Compound & 22 (40.7) \\
   \quad Dysplastic Junctional & 8 (14.8) \\
   Total Melanomas & 13 (24.1) \\
   \quad In Situ & 11 (20.4) \\
   \quad Thin Invasive & 2 (3.7) \\
   Other Lesions & 4 (7.4) \\
   \midrule
   \textbf{Number of Lesions per Patient} & \\
   1 & 36 (66.7) \\
   2 & 12 (22.2) \\
   3 & 2 (3.7) \\
   4 & 2 (3.7) \\
   5 & 2 (3.7) \\
   \bottomrule
 \end{tabular}
 \caption{\textcolor{black}{Demographic characteristics of the ACEMID\_pilot\_study dataset.}}
 \label{tab:acemid_pilot_demographics}
\end{table}
\begin{table}[h]
 \centering
 \begin{tabular}{p{3.2cm}|p{11.8cm}}
   \toprule
   \textbf{Major Category} & \textbf{Classes} \\
   \midrule
   \textbf{Skin Cancer and High-Risk Lesions} \textbf{(13)} & 
   Basal Cell Carcinoma - Nodular, Basal Cell Carcinoma - Superficial, Basal Cell Carcinoma - Pigmented, 
   Basal Cell Carcinoma - Recurrent, Melanoma - Nodular, Lentigo Maligna, Squamous Cell Carcinoma - Common, 
   Squamous Cell Carcinoma - Bowen's, Keratoacanthoma, Actinic Keratosis - Common, Actinic Keratosis - Hypertrophic,
   Actinic Keratosis - Pigmented, Actinic Cheilitis \\
   \midrule
   \textbf{Melanocytic Lesions} \textbf{(22)} & 
   Compound Nevus, Junctional Nevus, Dermal Nevus, Acral Parallel Pattern Nevus, Acral Lattice Pattern Nevus,
   Acral Untyped Pattern Nevus, Blue Nevus, Spitz Nevus, Reed Nevus, Congenital Nevus, Halo Nevus, Agminated Nevus,
   Atypical Nevus, Solar Lentigo, Lentigo Simplex, Ink-spot Lentigo, Ephelides, Melanosis, En Cocarde Nevus,
   Lentiginous Nevus, Subungual Melanocytic Nevus, Nevus with Regression \\
   \midrule
   \textbf{Common Benign Growths} \textbf{(13)} & 
   Seborrheic Keratosis - Common, Seborrheic Keratosis - Inflamed, Seborrheic Keratosis - Pigmented,
   Dermatofibroma, Epidermal Cyst, Sebaceous Hyperplasia, Skin Tag, Fibrous Papule of Face, Cutaneous Horn,
   Comedone, Accessory Nipple, Myxoid Cyst, Chondrodermatitis \\
   \midrule
   \textbf{Inflammatory Conditions} \textbf{(8)} & 
   Psoriasis, Eczema, Dermatitis, Folliculitis, Granuloma Annulare, Lichenoid Dermatitis, Porokeratosis,
   Hypertrophic Lichen Planus \\
   \midrule
   \textbf{Vascular and Infectious Lesions} \textbf{(10)} & 
   Angiokeratoma, Angioma, Telangiectasia, Hematoma - Common, Hematoma - Subcorneal, Hematoma - Subungual,
   Molluscum Contagiosum, Wart, Vascular Malformation, Pyogenic Granuloma \\
   \midrule
   \textbf{Others} \textbf{(8)} & 
   Post-inflammatory Changes, Nail Dystrophy, Hypertrophic Scar, Atrophic Scar, Excoriation, Traumatic Changes,
   Striae, Chemical Burn \\
   \bottomrule
 \end{tabular}
 \caption{\textcolor{black}{The taxonomy of 74 dermatological conditions covered in MMT-74 dataset.}}
 \label{tab:mmt74_taxonomy}
\end{table}

\begin{table}
\footnotesize
\centering
\begin{tabular}{ll|llll}
\toprule
Dataset & Model & W\_F1 & AUROC & BACC & AUPR \\
\midrule\midrule
\multirow{4}{*}{\rotatebox[origin=c]{90}{PAD}}  & SL\_Imagenet & 0.678 (0.636-0.720)$^{***}$ & 0.887 (0.864-0.911)$^{***}$ & 0.614 (0.540-0.688) & 0.759 (0.718-0.799)$^{***}$ \\
& DINOv2 & 0.702 (0.662-0.743)$^{**}$ & 0.885 (0.861-0.908)$^{***}$ & 0.607 (0.535-0.679)$^{*}$ & 0.753 (0.710-0.796)$^{***}$ \\
& SwaVDerm & 0.656 (0.614-0.698)$^{***}$ & 0.865 (0.838-0.891)$^{***}$ & 0.534 (0.469-0.599)$^{**}$ & 0.718 (0.675-0.761)$^{***}$ \\
&  PanDerm & \textbf{0.768 (0.732-0.805)} & \textbf{0.935 (0.919-0.951)} & \textbf{0.694 (0.624-0.764)} & \textbf{0.849 (0.817-0.880)} \\
\midrule
\multirow{4}{*}{\rotatebox[origin=c]{90}{HAM10000}}  & SL\_Imagenet & 0.879 (0.863-0.895)$^{***}$ & 0.970 (0.962-0.978)$^{***}$ & 0.653 (0.586-0.720)$^{***}$ & 0.922 (0.909-0.936)$^{***}$ \\
& DINOv2 & 0.883 (0.868-0.899)$^{***}$ & 0.964 (0.954-0.974)$^{***}$ & 0.701 (0.637-0.765)$^{***}$ & 0.916 (0.901-0.931)$^{***}$ \\
& SwaVDerm & 0.865 (0.848-0.883)$^{***}$ & 0.967 (0.959-0.975)$^{***}$ & 0.592 (0.521-0.663)$^{***}$ & 0.910 (0.896-0.924)$^{***}$ \\
& PanDerm & \textbf{0.926 (0.912-0.940)} & \textbf{0.988 (0.984-0.992)} & \textbf{0.807 (0.756-0.859)} & \textbf{0.959 (0.949-0.970)} \\
\midrule
\multirow{4}{*}{\rotatebox[origin=c]{90}{DermC}}  & SL\_Imagenet & 0.756 (0.694-0.818) & 0.797 (0.726-0.867)$^{*}$ & 0.700 (0.631-0.770) & 0.683 (0.575-0.791)$^{*}$ \\
& DINOv2 & 0.763 (0.702-0.824) & 0.796 (0.719-0.872)$^{**}$ & 0.717 (0.649-0.786) & 0.735 (0.644-0.826) \\
& SwaVDerm & 0.750 (0.687-0.814) & 0.768 (0.690-0.846)$^{**}$ & 0.700 (0.628-0.771) & 0.658 (0.543-0.774)$^{**}$ \\
& PanDerm & \textbf{0.788 (0.728-0.847)} & \textbf{0.876 (0.824-0.928)} & \textbf{0.740 (0.672-0.808)} & \textbf{0.798 (0.710-0.886)} \\
\midrule
\multirow{4}{*}{\rotatebox[origin=c]{90}{BCN20000}}  & SL\_Imagenet & 0.698 (0.673-0.722)$^{***}$ & 0.914 (0.903-0.925)$^{***}$ & 0.592 (0.537-0.647)$^{*}$ & 0.754 (0.728-0.779)$^{***}$ \\
& DINOv2 & 0.724 (0.701-0.747)$^{***}$ & 0.927 (0.917-0.938)$^{***}$ & 0.575 (0.518-0.632)$^{**}$ & 0.787 (0.765-0.810)$^{***}$ \\
& SwaVDerm & 0.696 (0.672-0.720)$^{***}$ & 0.908 (0.897-0.919)$^{***}$ & 0.499 (0.444-0.554)$^{***}$ & 0.742 (0.717-0.768)$^{***}$ \\
&  PanDerm & \textbf{0.772 (0.750-0.795)} & \textbf{0.952 (0.944-0.960)} & \textbf{0.662 (0.616-0.708)} & \textbf{0.846 (0.825-0.867)} \\
\midrule
\multirow{4}{*}{\rotatebox[origin=c]{90}{DDI}}  & SL\_Imagenet & 0.773 (0.710-0.836) & 0.743 (0.639-0.847)$^{**}$ & 0.655 (0.554-0.755) & 0.412 (0.283-0.541)$^{*}$ \\
& DINOv2 & 0.756 (0.695-0.816) & 0.749 (0.649-0.849)$^{*}$ & 0.612 (0.518-0.706)$^{*}$ & 0.456 (0.308-0.605) \\
& SwaVDerm & 0.730 (0.668-0.792) & 0.747 (0.650-0.845)$^{**}$ & 0.571 (0.483-0.660)$^{**}$ & 0.421 (0.294-0.548)$^{*}$ \\
& PanDerm & \textbf{0.790 (0.728-0.852)} & \textbf{0.843 (0.768-0.918)} & \textbf{0.722 (0.624-0.819)} & \textbf{0.551 (0.397-0.705)} \\
\midrule
\multirow{4}{*}{\rotatebox[origin=c]{90}{HIBA}}  & SL\_Imagenet & 0.881 (0.850-0.911)$^{**}$ & 0.881 (0.823-0.940)$^{**}$ & 0.685 (0.612-0.759)$^{**}$ & 0.638 (0.511-0.765)$^{*}$ \\
& DINOv2 & 0.884 (0.852-0.915)$^{***}$ & 0.873 (0.816-0.930)$^{**}$ & 0.737 (0.663-0.812)$^{***}$ & 0.616 (0.486-0.747)$^{*}$ \\
& SwaVDerm & 0.865 (0.834-0.895)$^{***}$ & 0.854 (0.799-0.910)$^{***}$ & 0.657 (0.585-0.729)$^{***}$ & 0.519 (0.387-0.651)$^{***}$ \\
 & PanDerm & \textbf{0.928 (0.901-0.954)} & \textbf{0.948 (0.922-0.975)} & \textbf{0.810 (0.740-0.880)} & \textbf{0.765 (0.652-0.878)} \\
\midrule
\multirow{4}{*}{\rotatebox[origin=c]{90}{MSKCC}}  & SL\_Imagenet & 0.712 (0.691-0.732)$^{**}$ & 0.733 (0.708-0.758)$^{**}$ & 0.635 (0.611-0.660) & 0.482 (0.444-0.521)$^{***}$ \\
& DINOv2 & 0.707 (0.687-0.728)$^{**}$ & 0.722 (0.695-0.748)$^{**}$ & \textbf{0.662 (0.636-0.689)} & 0.497 (0.457-0.537)$^{***}$ \\
& SwaVDerm & 0.715 (0.696-0.733)$^{*}$ & 0.720 (0.692-0.748)$^{***}$ & 0.617 (0.593-0.641)$^{**}$ & 0.515 (0.474-0.556)$^{***}$ \\
 & PanDerm & \textbf{0.737 (0.718-0.756)} & \textbf{0.761 (0.735-0.787)} & 0.653 (0.628-0.677) & \textbf{0.589 (0.548-0.630)} \\
\midrule
\multirow{4}{*}{\rotatebox[origin=c]{90}{PATCH16}}  & SL\_Imagenet & 0.873 (0.867-0.878)$^{***}$ & 0.992 (0.991-0.992)$^{***}$ & 0.834 (0.808-0.859)$^{***}$ & 0.936 (0.932-0.941)$^{***}$ \\
& DINOv2 & 0.892 (0.886-0.897)$^{***}$ & 0.993 (0.992-0.994)$^{**}$ & 0.820 (0.813-0.828)$^{***}$ & 0.945 (0.941-0.949) \\
& SwaVDerm & 0.816 (0.809-0.822)$^{***}$ & 0.984 (0.983-0.985)$^{***}$ & 0.742 (0.734-0.751)$^{***}$ & 0.891 (0.885-0.896)$^{***}$ \\
& PanDerm & \textbf{0.903 (0.898-0.908)} & \textbf{0.994 (0.993-0.994)} & \textbf{0.879 (0.854-0.903)} & \textbf{0.946 (0.943-0.950)} \\
\midrule
\multirow{4}{*}{\rotatebox[origin=c]{90}{ISIC2024}}  & SL\_Imagenet & 0.877 (0.873-0.877)$^{***}$ & 0.849 (0.765-0.917)$^{***}$ & 0.727 (0.635-0.811)$^{**}$ & 0.860 (0.835-0.885)$^{***}$ \\
& DINOv2 & 0.851 (0.849-0.853)$^{***}$ & 0.827 (0.745-0.870)$^{***}$ & 0.682 (0.606-0.752)$^{***}$ & 0.850 (0.825-0.875)$^{***}$ \\
& SwaVDerm & 0.869 (0.866-0.871)$^{***}$ & 0.852 (0.789-0.877)$^{***}$ & 0.689 (0.599-0.774)$^{***}$ & 0.835 (0.810-0.860)$^{***}$ \\
 & PanDerm & \textbf{0.929 (0.927-0.931)} & \textbf{0.893 (0.839-0.940)} & \textbf{0.799 (0.718-0.873)} & \textbf{0.915 (0.895-0.935)} \\
\midrule
\multirow{4}{*}{\rotatebox[origin=c]{90}{WSI}}  & SL\_Imagenet & 0.945 (0.925-0.965) & 0.989 (0.977-1.002) & 0.937 (0.906-0.960)$^{***}$ & 0.941 (0.930-0.947)$^{***}$ \\
& DINOv2 & 0.947 (0.919-0.974) & 0.988 (0.976-1.000) & 0.932 (0.910-0.945)$^{***}$ & 0.930 (0.906-0.956)$^{***}$ \\
& SwaVDerm & 0.932 (0.914-0.951) & 0.989 (0.978-1.000)$^{*}$ & 0.893 (0.882-0.910)$^{***}$ & 0.920 (0.905-0.935)$^{***}$ \\
& PanDerm & \textbf{0.953 (0.930-0.975)} & \textbf{0.994 (0.986-1.001)} & \textbf{0.976 (0.963-0.994)} & \textbf{0.981 (0.972-0.993)} \\
\bottomrule
\end{tabular}
\caption{\textbf{Skin cancer diagnosis performance of different models across multinational datasets.} Models include SL\_Imagenet (supervised learning on ImageNet), DINOv2, SwaVDerm, and PanDerm. Performance is reported using Weighted F1 score (W\_F1), Area Under the Receiver Operating Characteristic curve (AUROC), Balanced Accuracy (BACC), and Area Under the Precision-Recall curve (AUPR). Further details on the experimental setup, datasets, and metrics are provided in \textbf{Methods}. Best-performing model for each metric and dataset is bolded and highlighted. 95\% CI is included in parentheses. Significance levels for comparisons with the best model: $^{*}p<0.05$, $^{**}p<0.01$, $^{***}p<0.001$.}
\label{tab1}
\end{table}

\begin{table}
\footnotesize\centering
\begin{tabular}{ll|llll}
\toprule
Model & Dataset & W\_F1 & AUROC & BACC & AUPR \\
\midrule\midrule
SL\_Imagenet & MMT-09 & 0.661 (0.653, 0.668)$^{***}$ & 0.860 (0.846, 0.874)$^{***}$ & 0.404 (0.377, 0.430)$^{***}$ & 0.482 (0.460, 0.503)$^{***}$ \\
DINOv2 & MMT-09 & 0.672 (0.657, 0.687)$^{***}$ & 0.858 (0.839, 0.877)$^{***}$ & 0.433 (0.412, 0.454)$^{***}$ & 0.474 (0.459, 0.490)$^{***}$ \\
SwaVDerm & MMT-09 & 0.624 (0.615, 0.634)$^{***}$ & 0.814 (0.802, 0.825)$^{***}$ & 0.356 (0.347, 0.365)$^{***}$ & 0.411 (0.401, 0.422)$^{***}$ \\
\rowcolor{gray!10} PanDerm & MMT-09 & \textbf{0.704 (0.699, 0.709)} & \textbf{0.901 (0.888, 0.913)} & \textbf{0.462 (0.436, 0.488)} & \textbf{0.560 (0.539, 0.581)} \\
\midrule
SL\_Imagenet & MMT-74 & 0.417 (0.411, 0.423)$^{***}$ & 0.822 (0.807, 0.836)$^{***}$ & 0.119 (0.101, 0.138)$^{***}$ & 0.144 (0.125, 0.162)$^{***}$ \\
DINOv2 & MMT-74 & 0.414 (0.401, 0.428)$^{***}$ & 0.842 (0.830, 0.853)$^{***}$ & 0.115 (0.103, 0.127)$^{***}$ & 0.146 (0.127, 0.165)$^{***}$ \\
SwaVDerm & MMT-74 & 0.349 (0.340, 0.358)$^{***}$ & 0.774 (0.758, 0.790)$^{***}$ & 0.085 (0.073, 0.098)$^{***}$ & 0.105 (0.085, 0.124)$^{***}$ \\
\rowcolor{gray!10} PanDerm & MMT-74 & \textbf{0.488 (0.482, 0.494)} & \textbf{0.887 (0.872, 0.902)} & \textbf{0.174 (0.159, 0.189)} & \textbf{0.211 (0.186, 0.235)} \\
\midrule
SL\_Imagenet & DermNet & 0.497 (0.481, 0.512)$^{***}$ & 0.885 (0.878, 0.892)$^{***}$ & 0.462 (0.444, 0.480)$^{***}$ & 0.426 (0.407, 0.444)$^{***}$ \\
DINOv2 & DermNet & 0.536 (0.521, 0.551)$^{***}$ & 0.902 (0.896, 0.909)$^{***}$ & 0.505 (0.487, 0.523)$^{***}$ & 0.476 (0.456, 0.496)$^{***}$ \\
SwaVDerm & DermNet & 0.474 (0.458, 0.490)$^{***}$ & 0.884 (0.878, 0.891)$^{***}$ & 0.442 (0.424, 0.460)$^{***}$ & 0.428 (0.410, 0.446)$^{***}$ \\
\rowcolor{gray!10} PanDerm & DermNet & \textbf{0.619 (0.603, 0.634)} & \textbf{0.944 (0.939, 0.949)} & \textbf{0.586 (0.568, 0.603)} & \textbf{0.623 (0.603, 0.642)} \\
\bottomrule
\end{tabular}
\caption{\textbf{General multi-class skin condition classification performance of different models on MMT-09, MMT-74, and Dermnet datasets}. All models were evaluated on three datasets: MMT-09, MMT-74, and DermNet. Models include SL\_Imagenet (supervised learning on ImageNet), DINOv2, SwaVDerm, and PanDerm. Performance is reported using Weighted F1 score (W\_F1), Area Under the Receiver Operating Characteristic curve (AUROC), Balanced Accuracy (BACC), and Area Under the Precision-Recall curve (AUPR). The best-performing model for each metric and dataset is bolded. 95\% CI is included in parentheses. $^{***}p<0.001$ compared to PanDerm.}
\label{tab2}
\end{table}

\begin{table}
\footnotesize\centering
\setlength{\tabcolsep}{4pt}
\centering
\begin{tabular}{ll|llll}
\toprule
Dataset & Model & AUROC & Sensitivity & Specificity & BACC \\
\midrule\midrule
\multirow{4}{*}{\rotatebox[origin=c]{90}{SDDI1P}}
 & Default & 0.596 (0.567-0.624)$^{***}$ & 0.173 (0.141-0.205)$^{***}$ & \textbf{0.969 (0.939-0.988)} & 0.571 (0.542-0.611)$^{***}$ \\
 & w/ Warp & 0.673 (0.643-0.702)$^{***}$ & 0.533 (0.493-0.562)$^{***}$ & 0.765 (0.735-0.794)$^{***}$ & 0.649 (0.608-0.685)$^{***}$ \\
 & w/ Mask & 0.648 (0.629-0.662)$^{***}$ & 0.600 (0.571-0.626)$^{***}$ & 0.646 (0.617-0.675)$^{***}$ & 0.623 (0.594-0.652)$^{***}$ \\
\rowcolor{gray!10} &  w/ Whole pipeline & \textbf{0.706 (0.686-0.725)} & \textbf{0.653 (0.634-0.673)} & 0.741 (0.722-0.751)$^{***}$ & \textbf{0.697 (0.688-0.717)} \\
\midrule
\multirow{4}{*}{\rotatebox[origin=c]{90}{SDDI2P}}
 & Default & 0.683 (0.517-0.849)$^{***}$ & \textbf{0.940 (0.864-1.000)} & 0.239 (0.115-0.593)$^{***}$ & 0.590 (0.449-0.730)$^{***}$ \\
 & w/ Warp & 0.710 (0.579-0.841)$^{***}$ & 0.942 (0.862-1.000) & 0.273 (0.013-0.533)$^{***}$ & 0.607 (0.506-0.709)$^{***}$ \\
 & w/ Mask & 0.695 (0.564-0.822)$^{***}$ & 0.935 (0.853-0.995) & 0.255 (0.012-0.511)$^{***}$ & 0.595 (0.495-0.695)$^{***}$ \\
\rowcolor{gray!10} & w/ Whole pipeline & \textbf{0.767 (0.649-0.886)} & 0.854 (0.797-0.911)$^{***}$ & \textbf{0.577 (0.387-0.768)} & \textbf{0.716 (0.621-0.810)} \\
\bottomrule
\end{tabular}
\caption{\textbf{Ablation study on pre-processing methods for short-term lesion change detection based on SDDI1P and SDDI2P datasets.} Metrics: AUROC, Sensitivity, Specificity, BACC (Balanced Accuracy). Warp denoted image registration, Mask denoted lesion segmentation, and the Whole pipeline denoted our proposed pre-processing pipeline. The best model is bolded and highlighted. 95\% CI in parentheses. $^{*}p<0.05$, $^{**}p<0.01$, $^{***}p<0.001$.}
\label{tab3}
\end{table}

\begin{table}
\footnotesize\centering
\setlength{\tabcolsep}{4pt}
\centering
\begin{tabular}{ll|llll}
\toprule
Dataset & Model & AUROC & Sensitivity & Specificity & BACC \\
\midrule\midrule
\multirow{4}{*}{\rotatebox[origin=c]{90}{\raisebox{3em}{SDDI1}}}
 & SL\_Imagenet & 0.616 (0.599-0.634)$^{***}$ & 0.520 (0.501-0.543)$^{***}$ & 0.647 (0.628-0.669)$^{***}$ & 0.584 (0.567-0.613)$^{***}$ \\
 & DINOv2 & 0.660 (0.649-0.678)$^{***}$ & 0.573 (0.554-0.592)$^{***}$ & 0.601 (0.586-0.622)$^{***}$ & 0.587 (0.559-0.607)$^{***}$ \\
 & SwaVDerm & 0.632 (0.614-0.652)$^{***}$ & 0.191 (0.163-0.214)$^{***}$ & \textbf{0.985 (0.961-0.999)} & 0.588 (0.567-0.604)$^{***}$ \\
 \rowcolor{gray!10} & PanDerm & \textbf{0.706 (0.686-0.725)} & \textbf{0.653 (0.634-0.673)} & 0.741 (0.722-0.751)$^{***}$ & \textbf{0.697 (0.688-0.717)} \\
\midrule
\multirow{4}{*}{\rotatebox[origin=c]{90}{\raisebox{3em}{SDDI2}}}
 & SL\_Imagenet & 0.715 (0.594-0.837) & 0.870 (0.737-1.000) & 0.392 (0.234-0.550) & 0.631 (0.582-0.681) \\
 & DINOv2 & 0.730 (0.533-0.928) & 0.826 (0.673-0.979) & \textbf{0.584 (0.160-1.000)} & 0.705 (0.556-0.853) \\
 & SwaVDerm & 0.656 (0.547-0.764)$^{*}$ & \textbf{0.970 (0.920-1.000)} & 0.181 (0.051-0.412)$^{**}$ & 0.575 (0.482-0.669)$^{*}$ \\
 \rowcolor{gray!10} & PanDerm & \textbf{0.767 (0.649-0.886)} & 0.854 (0.797-0.911) & 0.577 (0.387-0.768) & \textbf{0.716 (0.621-0.810)} \\
\bottomrule
\end{tabular}
\caption{\textbf{Short-term lesion change detection performance of different models on SDDI1 and SDDI2 datasets.} Models: SL\_Imagenet, DINOv2, SwaVDerm, and PanDerm. Metrics: AUROC, Sensitivity, Specificity, BACC (Balanced Accuracy). The best model is bolded and highlighted. 95\% CI in parentheses. $^{*}p<0.05$, $^{**}p<0.01$, $^{***}p<0.001$.}
\label{tab4}
\end{table}

\begin{table}
\footnotesize\centering
\setlength{\tabcolsep}{4pt}
\centering
\begin{tabular}{ll|ll}
\toprule
Dataset & Model & AUROC & BACC \\
\midrule\midrule
\multirow{4}{*}{SDDI2M}
 & SL\_Imagenet & 0.690 (0.580-0.800)$^{**}$ & 0.588 (0.435-0.740)$^{***}$ \\
 & DINOv2 & 0.665 (0.565-0.755)$^{**}$ & 0.614 (0.460-0.770)$^{***}$ \\
 & SwaVDerm & 0.650 (0.558-0.742)$^{**}$ & 0.540 (0.429-0.762)$^{***}$ \\
 \rowcolor{gray!10} & PanDerm & \textbf{0.840 (0.769-0.911)} & \textbf{0.660 (0.472-0.848)} \\
\bottomrule
\end{tabular}
\caption{\textbf{Malignant lesion change detection performance of different models on SDDI2 dataset.} classes: malignant lesion change vs others. Models: SL\_Imagenet, DINOv2, SwaVDerm, and PanDerm. Metrics: AUROC and BACC (Balanced Accuracy). The best model is bolded and highlighted. 95\% CI in parentheses. $^{*}p<0.05$, $^{**}p<0.01$, $^{***}p<0.001$.}
\label{tab5}
\end{table}

\begin{table}
\footnotesize\centering
\setlength{\tabcolsep}{4pt}
\begin{tabular}{ll|llll}
\toprule
Dataset & Model & W\_F1 & AUROC & BACC & AUPR \\
\midrule\midrule
\multirow{4}{*}{\makecell{Combinemel\\(2 classes)}}
 & SL\_Imagenet & 0.839 (0.781-0.897) & 0.938 (0.898-0.978) & 0.833 (0.767-0.899) & 0.915 (0.867-0.964) \\
 & DINOv2 & 0.841 (0.780-0.901)$^{*}$ & 0.921 (0.872-0.971)$^{**}$ & 0.842 (0.778-0.907) & 0.886 (0.823-0.949)$^{***}$ \\
 & SwaVDerm & 0.847 (0.790-0.904) & 0.944 (0.905-0.983) & 0.858 (0.796-0.920) & 0.924 (0.876-0.962) \\
 \rowcolor{gray!10} & PanDerm & \textbf{0.889 (0.837-0.941)} & \textbf{0.964 (0.937-0.991)} & \textbf{0.882 (0.822-0.942)} & \textbf{0.944 (0.907-0.982)} \\
\midrule
\multirow{4}{*}{\makecell{Combinemel\\(3 classes)}}
 & SL\_Imagenet & 0.661 (0.598-0.725)$^{*}$ & 0.834 (0.789-0.879)$^{**}$ & 0.526 (0.429-0.624) & 0.734 (0.694-0.773)$^{*}$ \\
 & DINOv2 & 0.632 (0.571-0.692)$^{*}$ & 0.833 (0.785-0.880)$^{**}$ & 0.474 (0.376-0.571)$^{**}$ & 0.728 (0.689-0.767)$^{**}$ \\
 & SwaVDerm & 0.693 (0.633-0.744)$^{*}$ & 0.875 (0.832-0.917)$^{*}$ & 0.601 (0.505-0.697) & 0.775 (0.728-0.822) \\
 \rowcolor{gray!10} & PanDerm & \textbf{0.721 (0.662-0.780)} & \textbf{0.896 (0.860-0.932)} & \textbf{0.624 (0.530-0.719)} & \textbf{0.792 (0.746-0.838)} \\
\bottomrule
\end{tabular}
\caption{\textbf{Metastasis prediction performance of different models on Combinemel dataset (2 classes and 3 classes)}. 2 classes: metastasis vs control (no metastasis). 3 classes: local metastasis vs distant metastasis vs control. Models include SL\_Imagenet (supervised learning on ImageNet), DINOv2, SwaVDerm, and PanDerm. Metrics: Weighted F1 score (W\_F1), Area Under the Receiver Operating Characteristic curve (AUROC), Balanced Accuracy (BACC), and Area Under the Precision-Recall curve (AUPR). The best-performing model for each metric and dataset is bolded and highlighted. 95\% CI in parentheses. Significance levels: $^{*}p<0.05$, $^{**}p<0.01$, $^{***}p<0.001$.}
\label{tab6}
\end{table}

\begin{table}
\footnotesize\centering
\setlength{\tabcolsep}{4pt}
\begin{tabular}{ll|llll}
\toprule
Dataset & Model & W\_F1 & AUROC & BACC & AUPR \\
\midrule\midrule
\multirow{4}{*}{\begin{tabular}[c]{@{}c@{}}Solardamage \end{tabular}}
 & SL\_Imagenet & 0.890 (0.873-0.907) & 0.950 (0.939-0.962) & 0.829 (0.799-0.860) & 0.899 (0.874-0.924) \\
 & DINOv2 & 0.868 (0.849-0.888) & 0.947 (0.935-0.958) & 0.810 (0.779-0.840) & 0.875 (0.847-0.903) \\
 & SwaVDerm & 0.888 (0.870-0.905) & 0.958 (0.948-0.968) & 0.826 (0.795-0.857) & 0.908 (0.886-0.931) \\
 \rowcolor{gray!10} & PanDerm & \textbf{0.896 (0.879-0.913)} & \textbf{0.961 (0.951-0.971)} & \textbf{0.845 (0.816-0.874)} & \textbf{0.917 (0.893-0.940)} \\
\bottomrule
\end{tabular}
\caption{\textbf{Solar damage risk assessment performance of different models on HOP\&MYM\_solar dataset}. 3 classes: low vs medium vs high risk. Models include SL\_Imagenet (supervised learning on ImageNet), DINOv2, SwaVDerm, and PanDerm. Metrics: Weighted F1 score (W\_F1), Area Under the Receiver Operating Characteristic curve (AUROC), Balanced Accuracy (BACC), and Area Under the Precision-Recall curve (AUPR). The best-performing model for each metric is bolded and highlighted. 95\% CI in parentheses.}
\label{tab7}
\end{table}

\begin{table}
\footnotesize\centering
\setlength{\tabcolsep}{4pt}
\begin{tabular}{ll|llll}
\toprule
Dataset & Model & W\_F1 & AUROC & BACC & AUPR \\
\midrule\midrule
\multirow{4}{*}{MYM}  
 & SL\_Imagenet & 0.952 (0.950-0.954)$^{***}$ & 0.979 (0.973-0.985)$^{***}$ & 0.810 (0.803-0.817)$^{***}$ & 0.836 (0.821-0.850)$^{***}$ \\
 & DINOv2 & \textbf{0.956 (0.953-0.959)} & 0.980 (0.977-0.983)$^{***}$ & \textbf{0.838 (0.825-0.851)}$^{***}$ & 0.834 (0.808-0.861)$^{***}$ \\
 & SwaVDerm & 0.944 (0.938-0.950)$^{***}$ & 0.973 (0.969-0.978)$^{***}$ & 0.774 (0.765-0.782)$^{***}$ & 0.819 (0.799-0.839)$^{***}$ \\
 \rowcolor{gray!10} & PanDerm & 0.956 (0.953-0.958) & \textbf{0.983 (0.979-0.987)} & 0.820 (0.799-0.842) & \textbf{0.844 (0.820-0.868)} \\
\bottomrule
\end{tabular}
\caption{\textbf{Nevus counting performance of different models on a subset of MYM dataset}. Models include SL\_Imagenet (supervised learning on ImageNet), DINOv2, SwaVDerm, and PanDerm. Metrics: Weighted F1 score (W\_F1), Area Under the Receiver Operating Characteristic curve (AUROC), Balanced Accuracy (BACC), and Area Under the Precision-Recall curve (AUPR). The best-performing model for each metric is bolded and highlighted. 95\% CI in parentheses. $^{***}p<0.001$ compared to PanDerm.}
\label{tab8}
\end{table}

\begin{table}
\footnotesize\centering
\setlength{\tabcolsep}{4pt}
\begin{tabular}{l|lll}
\toprule
Model & AUC & W-F1 & BACC \\
\midrule
SL-Imagenet & 68.09 (67.40-68.80)$^{***}$ & 71.78 (71.27-72.33)$^{***}$ & 62.09 (61.56-62.59)$^{***}$ \\
DINOv2 & 67.83 (66.69-68.03)$^{***}$ & 69.83 (68.92-70.34)$^{***}$ & 59.40 (58.75-59.92)$^{***}$ \\
SwavDerm & 66.74 (66.22-67.10)$^{***}$ & 68.15 (67.76-68.59)$^{***}$ & 61.01 (60.53-61.49)$^{***}$ \\
\rowcolor{gray!10} PanDerm & \textbf{70.47 (69.76-71.15)} & \textbf{73.63 (62.74-73.92)} & \textbf{65.72 (65.13-66.28)} \\
\bottomrule
\end{tabular}
\caption{\textbf{Risk stratification performance of different models on HOP\&MYM datasets}. The table shows the performance metrics for different models. Metrics include Area Under the Curve (AUC), Weighted F1 score (W-F1), and Balanced Accuracy (BACC). The best-performing model for each metric is bolded and highlighted. 95\% CI in parentheses. All p-values $< 0.001$ ($^{***}$).}
\label{tab9}
\end{table}

\begin{table}
\footnotesize\centering
\begin{tabular}{ll|lll|lll}
\toprule
Model & Prediction & \multicolumn{3}{c|}{Benign} & \multicolumn{3}{c}{Malignant} \\
 & head & Precision & Recall & F1-score & Precision & Recall & F1-score \\
\midrule
SL\_Imagenet & UD & 0.999 & 0.943 & 0.971 & 0.007 & 0.464 & 0.015 \\
 & CLS & 1.000 & 0.955 & 0.977 & 0.013 & 0.643 & 0.025 \\
 & CMB & 1.000 & 0.921 & 0.959 & 0.008 & 0.679 & 0.015 \\
 & CMB\_ML & 1.000 & 0.883 & 0.938 & 0.006 & 0.821 & 0.013 \\
\midrule
DINOv2 & UD & 1.000 & 0.967 & 0.983 & 0.014 & 0.500 & 0.027 \\
 & CLS & 1.000 & 0.954 & 0.977 & 0.013 & 0.679 & 0.026 \\
 & CMB & 1.000 & 0.939 & 0.968 & 0.010 & 0.679 & 0.020 \\
 & CMB\_ML & 1.000 & 0.900 & 0.947 & 0.007 & 0.821 & 0.015 \\
\midrule
SwavDerm & UD & 0.999 & 0.958 & 0.978 & 0.009 & 0.393 & 0.017 \\
 & CLS & 1.000 & 0.961 & 0.980 & 0.012 & 0.500 & 0.023 \\
 & CMB & 1.000 & 0.936 & 0.967 & 0.008 & 0.571 & 0.016 \\
 & CMB\_ML & 1.000 & 0.896 & 0.945 & 0.007 & 0.857 & 0.015 \\
\midrule
PanDerm & UD & 1.000 & 0.943 & 0.971 & 0.009 & 0.571 & 0.018 \\
 & CLS & 1.000 & 0.971 & 0.985 & 0.016 & 0.500 & 0.031 \\
 & CMB & 1.000 & 0.928 & 0.962 & 0.009 & 0.714 & 0.018 \\
 & CMB\_ML & 1.000 & 0.887 & 0.940 & 0.007 & \textbf{0.893} & 0.014 \\
\bottomrule
\end{tabular}
\caption{\textbf{TBP-based Malignant lesion screening performance of different models and prediction head types on HOP\&MYM dataset}. Maligant recall is the most crucial metrics.  Results are shown for both benign and malignant classifications, including precision, recall, and F1-score.}
\label{tab10}
\end{table}

\begin{table}
\footnotesize\centering
\setlength{\tabcolsep}{4pt}
\begin{tabular}{ll|ll}
\toprule
Dataset & Model & DSC & JAC \\
\midrule\midrule
\multirow{4}{*}{ISIC2018}
 & SL-Imagenet & 0.876 (0.870-0.887)$^{***}$ & 0.807 (0.799-0.822)$^{***}$ \\
 & autoSMIM & 0.848 (0.845-0.851)$^{***}$ & 0.769 (0.766-0.771)$^{***}$ \\
 & BATFormer & 0.884 (0.880-0.889)$^{***}$ & 0.815 (0.809-0.823)$^{***}$ \\
 \rowcolor{gray!10} & PanDerm & \textbf{0.910 (0.907-0.913)} & \textbf{0.846 (0.842-0.850)} \\
\bottomrule
\end{tabular}
\caption{\textbf{Lesion segmentation performance of different models on ISIC2018 dataset}. Models include SL-Imagenet, autoSMIM, BATFormer, and PanDerm. Metrics: Dice Similarity Coefficient (DSC) and Jaccard Index (JAC). The best-performing model for each metric is bolded and highlighted. 95\% CI in parentheses. Significance levels: $^{*}p<0.05$, $^{**}p<0.01$, $^{***}p<0.001$.}
\label{tab11}
\end{table}

\begin{table}
\footnotesize\centering
\setlength{\tabcolsep}{4pt}
\begin{tabular}{ll|ll}
\toprule
Dataset & Model & DSC & JAC \\
\midrule\midrule
\multirow{4}{*}{HAM10000}
 & SL-Imagenet & 0.927 (0.926-0.929)$^{***}$ & 0.875 (0.873-0.878)$^{***}$ \\
 & autoSMIM & 0.920 (0.920-0.921)$^{***}$ & 0.865 (0.864-0.866)$^{***}$ \\
 & BATFormer & 0.937 (0.935-0.939)$^{***}$ & 0.891 (0.889-0.893)$^{***}$ \\
 \rowcolor{gray!10} & PanDerm & \textbf{0.949 (0.949-0.950)} & \textbf{0.910 (0.908-0.917)} \\
\bottomrule
\end{tabular}
\caption{\textbf{Lesion segmentation performance of different models on HAM10000 dataset}. Models include SL-Imagenet, autoSMIM, BATFormer, and PanDerm. Metrics: Dice Similarity Coefficient (DSC) and Jaccard Index (JAC). The best-performing model for each metric is bolded and highlighted. 95\% CI in parentheses. Significance levels: $^{*}p<0.05$, $^{**}p<0.01$, $^{***}p<0.001$.}
\label{tab12}
\end{table}

\begin{table}
\footnotesize\centering
\setlength{\tabcolsep}{4pt}
\begin{tabular}{ll|ll}
\toprule
Dataset & Model & DSC & JAC \\
\midrule\midrule
\multirow{2}{*}{\raisebox{2em}{ISIC2018}}
 & MedSAM & 0.904 (0.900-0.911) & 0.841 (0.836-0.848) \\
 \rowcolor{gray!10} & PanDerm & \textbf{0.910 (0.907-0.913)}$^{*}$ & \textbf{0.846 (0.842-0.850)}$^{*}$ \\
\midrule
\multirow{2}{*}{\raisebox{2em}{HAM10000}}
 & MedSAM & 0.949 (0.948-0.951) & 0.905 (0.904-0.907) \\
 \rowcolor{gray!10} & PanDerm & \textbf{0.949 (0.949-0.950)} & \textbf{0.910 (0.908-0.917)} \\
\bottomrule
\end{tabular}
\caption{\textbf{Lesion segmentation performance comparison of PanDerm and MedSAM on ISIC2018 and HAM10000 datasets}. Metrics: Dice Similarity Coefficient (DSC) and Jaccard Index (JAC). The best-performing model for each metric is bolded and highlighted. 95\% CI in parentheses. Significance levels: $^{*}p<0.05$, $^{**}p<0.01$, $^{***}p<0.001$. For ISIC2018: DSC p-value = 0.017, JAC p-value = 0.025. For HAM10000: DSC p-value = 0.793, JAC p-value = 0.112.}
\label{tab13}
\end{table}

\begin{table}
\footnotesize\centering
\setlength{\tabcolsep}{4pt}
\begin{tabular}{llllll}
\toprule
GPU & Model & Dataset & Training Time & Inference Time \\
\midrule\midrule
\multirow{4}{*}{A6000} 
 & \multirow{2}{*}{PanDerm} & ISIC2018 & 2h 11min & 52s \\
 & & HAM10000 & 6h 29min & 59s \\
\cmidrule{2-5}
 & \multirow{2}{*}{MedSAM} & ISIC2018 & 8h 32min & 2m 8s \\
 & & HAM10000 & 28h 57min & 4m 4s \\
\midrule
\multirow{4}{*}{RTX3090} 
 & \multirow{2}{*}{PanDerm} & ISIC2018 & 2h 33min & 46s \\
 & & HAM10000 & 7h 48min & 1m 6s \\
\cmidrule{2-5}
 & \multirow{2}{*}{MedSAM} & ISIC2018 & 11h 58min & 2m 16s \\
 & & HAM10000 & 38h 23min & 4m 22s \\
\bottomrule
\end{tabular}
\caption{\textbf{Training and inference times comparison between PanDerm and MedSAM on lesion segmentation}. The table shows the training and inference times for PanDerm and MedSAM on ISIC2018 and HAM10000 datasets, using A6000 and RTX3090 GPUs.}
\label{tab14}
\end{table}

\begin{table}
\footnotesize\centering
\setlength{\tabcolsep}{4pt}
\begin{tabular}{llllll}
\toprule
Percent & Model & W\_F1 & AUROC & BACC & AUPR \\
\midrule\midrule
\multirow{4}{*}{5\%} 
 & SL\_ImageNet & 0.792 (0.774-0.809)$^{***}$ & 0.919 (0.905-0.934)$^{***}$ & 0.337 (0.294-0.379)$^{***}$ & 0.853 (0.838-0.867)$^{***}$ \\
 & DINOv2 & 0.803 (0.786-0.821)$^{***}$ & 0.928 (0.913-0.943)$^{***}$ & 0.380 (0.322-0.437)$^{***}$ & 0.852 (0.837-0.867)$^{***}$ \\
 & SwAVDerm & 0.793 (0.776-0.810)$^{***}$ & 0.917 (0.901-0.932)$^{***}$ & 0.322 (0.288-0.356)$^{***}$ & 0.841 (0.826-0.855)$^{***}$ \\
 & PanDerm & \textbf{0.851 (0.834-0.868)} & \textbf{0.960 (0.950-0.970)} & \textbf{0.524 (0.459-0.589)} & \textbf{0.902 (0.888-0.915)} \\
\midrule
\multirow{4}{*}{10\%} 
 & SL\_ImageNet & 0.816 (0.797-0.834)$^{***}$ & 0.937 (0.925-0.949)$^{***}$ & 0.414 (0.365-0.464)$^{***}$ & 0.871 (0.856-0.886)$^{***}$ \\
 & DINOv2 & 0.824 (0.807-0.841)$^{***}$ & 0.939 (0.926-0.953)$^{***}$ & 0.448 (0.389-0.506)$^{***}$ & 0.871 (0.856-0.886)$^{***}$ \\
 & SwAVDerm & 0.811 (0.794-0.827)$^{***}$ & 0.923 (0.907-0.938)$^{***}$ & 0.365 (0.329-0.402)$^{***}$ & 0.854 (0.840-0.869)$^{***}$ \\
 & PanDerm & \textbf{0.872 (0.855-0.888)} & \textbf{0.969 (0.961-0.978)} & \textbf{0.618 (0.550-0.686)} & \textbf{0.923 (0.910-0.936)} \\
\midrule
\multirow{4}{*}{20\%} 
 & SL\_ImageNet & 0.838 (0.820-0.855)$^{***}$ & 0.940 (0.928-0.952)$^{***}$ & 0.506 (0.439-0.573)$^{***}$ & 0.880 (0.865-0.896)$^{***}$ \\
 & DINOv2 & 0.839 (0.822-0.856)$^{***}$ & 0.947 (0.935-0.959)$^{***}$ & 0.534 (0.467-0.600)$^{**}$ & 0.884 (0.869-0.899)$^{***}$ \\
 & SwAVDerm & 0.825 (0.809-0.842)$^{***}$ & 0.931 (0.916-0.946)$^{***}$ & 0.408 (0.357-0.460)$^{***}$ & 0.871 (0.855-0.886)$^{***}$ \\
 & PanDerm & \textbf{0.889 (0.873-0.905)} & \textbf{0.975 (0.968-0.982)} & \textbf{0.665 (0.594-0.736)} & \textbf{0.935 (0.924-0.947)} \\
\midrule
\multirow{4}{*}{30\%} 
 & SL\_ImageNet & 0.837 (0.820-0.855)$^{***}$ & 0.945 (0.933-0.956)$^{***}$ & 0.505 (0.438-0.571)$^{***}$ & 0.883 (0.867-0.899)$^{***}$ \\
 & DINOv2 & 0.848 (0.831-0.864)$^{***}$ & 0.951 (0.939-0.962)$^{***}$ & 0.559 (0.488-0.630)$^{**}$ & 0.894 (0.880-0.909)$^{***}$ \\
 & SwAVDerm & 0.839 (0.822-0.857)$^{***}$ & 0.939 (0.925-0.953)$^{***}$ & 0.493 (0.428-0.558)$^{***}$ & 0.879 (0.864-0.895)$^{***}$ \\
 & PanDerm & \textbf{0.895 (0.880-0.911)} & \textbf{0.979 (0.973-0.985)} & \textbf{0.721 (0.653-0.788)} & \textbf{0.943 (0.931-0.954)} \\
\midrule
\multirow{4}{*}{50\%} 
 & SL\_ImageNet & 0.855 (0.838-0.872)$^{***}$ & 0.957 (0.947-0.967)$^{***}$ & 0.565 (0.492-0.638)$^{***}$ & 0.904 (0.889-0.919)$^{***}$ \\
 & DINOv2 & 0.855 (0.838-0.872)$^{***}$ & 0.953 (0.940-0.965)$^{***}$ & 0.597 (0.529-0.664)$^{**}$ & 0.902 (0.888-0.916)$^{***}$ \\
 & SwAVDerm & 0.854 (0.836-0.871)$^{***}$ & 0.953 (0.941-0.964)$^{***}$ & 0.557 (0.489-0.626)$^{***}$ & 0.893 (0.878-0.908)$^{***}$ \\
 & PanDerm & \textbf{0.909 (0.894-0.924)} & \textbf{0.981 (0.976-0.987)} & \textbf{0.749 (0.685-0.814)} & \textbf{0.950 (0.939-0.961)} \\
\midrule
\multirow{4}{*}{100\%} 
 & SL\_ImageNet & 0.872 (0.856-0.888)$^{***}$ & 0.967 (0.958-0.976)$^{***}$ & 0.652 (0.584-0.720)$^{***}$ & 0.919 (0.905-0.934)$^{***}$ \\
 & DINOv2 & 0.876 (0.860-0.892)$^{***}$ & 0.962 (0.951-0.972)$^{***}$ & 0.686 (0.621-0.751)$^{**}$ & 0.913 (0.898-0.928)$^{***}$ \\
 & SwAVDerm & 0.864 (0.847-0.881)$^{***}$ & 0.963 (0.954-0.972)$^{***}$ & 0.592 (0.520-0.664)$^{***}$ & 0.904 (0.889-0.919)$^{***}$ \\
 & PanDerm & \textbf{0.922 (0.908-0.936)} & \textbf{0.988 (0.984-0.992)} & \textbf{0.797 (0.744-0.850)} & \textbf{0.959 (0.949-0.969)} \\
\bottomrule
\end{tabular}
\caption{\textbf{Label efficiency generalization performance for dermoscopic image-based skin cancer diagnosis based on  HAM\_clean dataset.} Metrics: W\_F1 (Weighted F1), AUROC, BACC (Balanced Accuracy), AUPR (Area Under Precision-Recall Curve). The best model for each setting is bolded. 95\% CI in parentheses. $^{*}p<0.05$, $^{**}p<0.01$, $^{***}p<0.001$ compared to PanDerm.}
\label{tab15}
\end{table}

\begin{table}
\footnotesize\centering
\setlength{\tabcolsep}{4pt}
\begin{tabular}{llllll}
\toprule
Percent & Model & W\_F1 & AUROC & BACC & AUPR \\
\midrule\midrule
\multirow{4}{*}{5\%} 
 & SL\_ImageNet & 0.543 (0.518-0.568)$^{***}$ & 0.815 (0.799-0.831)$^{***}$ & 0.305 (0.272-0.338) & 0.567 (0.539-0.594)$^{***}$ \\
 & DINOv2 & 0.565 (0.541-0.590) & 0.840 (0.824-0.855)$^{**}$ & 0.305 (0.278-0.332) & 0.616 (0.589-0.642) \\
 & SwAVDerm & 0.541 (0.517-0.565)$^{***}$ & 0.809 (0.793-0.825)$^{***}$ & 0.272 (0.248-0.297)$^{**}$ & 0.565 (0.538-0.593)$^{***}$ \\
 & PanDerm & \textbf{0.586 (0.563-0.609)} & \textbf{0.859 (0.846-0.872)} & \textbf{0.304 (0.278-0.330)} & \textbf{0.628 (0.601-0.654)} \\
\midrule
\multirow{4}{*}{10\%} 
 & SL\_ImageNet & 0.561 (0.536-0.586)$^{***}$ & 0.837 (0.822-0.851)$^{***}$ & 0.332 (0.295-0.369)$^{***}$ & 0.599 (0.569-0.628)$^{***}$ \\
 & DINOv2 & 0.602 (0.577-0.627)$^{**}$ & 0.855 (0.840-0.870)$^{***}$ & 0.373 (0.329-0.417)$^{*}$ & 0.642 (0.614-0.671)$^{***}$ \\
 & SwAVDerm & 0.571 (0.546-0.596)$^{***}$ & 0.833 (0.818-0.848)$^{***}$ & 0.321 (0.288-0.355)$^{***}$ & 0.590 (0.562-0.618)$^{***}$ \\
 & PanDerm & \textbf{0.650 (0.626-0.675)} & \textbf{0.890 (0.877-0.903)} & \textbf{0.417 (0.374-0.459)} & \textbf{0.704 (0.678-0.731)} \\
\midrule
\multirow{4}{*}{20\%} 
 & SL\_ImageNet & 0.612 (0.587-0.638)$^{***}$ & 0.862 (0.849-0.876)$^{***}$ & 0.392 (0.346-0.438)$^{*}$ & 0.647 (0.619-0.675)$^{***}$ \\
 & DINOv2 & 0.614 (0.589-0.638)$^{***}$ & 0.872 (0.859-0.886)$^{***}$ & 0.374 (0.330-0.418)$^{**}$ & 0.667 (0.638-0.695)$^{***}$ \\
 & SwAVDerm & 0.591 (0.566-0.615)$^{***}$ & 0.851 (0.836-0.865)$^{***}$ & 0.357 (0.318-0.396)$^{**}$ & 0.629 (0.601-0.656)$^{***}$ \\
 & PanDerm & \textbf{0.681 (0.658-0.704)} & \textbf{0.910 (0.899-0.921)} & \textbf{0.434 (0.395-0.473)} & \textbf{0.735 (0.708-0.761)} \\
\midrule
\multirow{4}{*}{30\%} 
 & SL\_ImageNet & 0.613 (0.587-0.639)$^{***}$ & 0.866 (0.853-0.880)$^{***}$ & 0.426 (0.373-0.478)$^{*}$ & 0.663 (0.636-0.690)$^{***}$ \\
 & DINOv2 & 0.648 (0.624-0.672)$^{***}$ & 0.880 (0.867-0.894)$^{***}$ & 0.435 (0.381-0.489)$^{*}$ & 0.687 (0.660-0.713)$^{***}$ \\
 & SwAVDerm & 0.610 (0.586-0.635)$^{***}$ & 0.861 (0.847-0.875)$^{***}$ & 0.343 (0.315-0.372)$^{***}$ & 0.639 (0.611-0.668)$^{***}$ \\
 & PanDerm & \textbf{0.703 (0.680-0.727)} & \textbf{0.923 (0.913-0.933)} & \textbf{0.509 (0.455-0.563)} & \textbf{0.766 (0.742-0.790)} \\
\midrule
\multirow{4}{*}{50\%} 
 & SL\_ImageNet & 0.649 (0.624-0.674)$^{***}$ & 0.888 (0.875-0.901)$^{***}$ & 0.495 (0.435-0.556)$^{*}$ & 0.698 (0.671-0.726)$^{***}$ \\
 & DINOv2 & 0.660 (0.636-0.684)$^{***}$ & 0.894 (0.883-0.906)$^{***}$ & 0.478 (0.424-0.532)$^{*}$ & 0.720 (0.695-0.746)$^{***}$ \\
 & SwAVDerm & 0.628 (0.602-0.653)$^{***}$ & 0.880 (0.867-0.893)$^{***}$ & 0.396 (0.352-0.439)$^{***}$ & 0.685 (0.657-0.713)$^{***}$ \\
 & PanDerm & \textbf{0.733 (0.709-0.757)} & \textbf{0.939 (0.930-0.948)} & \textbf{0.578 (0.521-0.636)} & \textbf{0.801 (0.778-0.825)} \\
\midrule
\multirow{4}{*}{100\%} 
 & SL\_ImageNet & 0.698 (0.673-0.723)$^{***}$ & 0.910 (0.898-0.922)$^{***}$ & 0.590 (0.539-0.641) & 0.747 (0.722-0.773)$^{***}$ \\
 & DINOv2 & 0.705 (0.681-0.729)$^{***}$ & 0.923 (0.913-0.933)$^{***}$ & 0.565 (0.506-0.624)$^{**}$ & 0.774 (0.748-0.800)$^{***}$ \\
 & SwAVDerm & 0.683 (0.659-0.708)$^{***}$ & 0.907 (0.895-0.918)$^{***}$ & 0.479 (0.428-0.529)$^{***}$ & 0.733 (0.707-0.758)$^{***}$ \\
 & PanDerm & \textbf{0.767 (0.742-0.791)} & \textbf{0.951 (0.944-0.959)} & \textbf{0.647 (0.602-0.692)} & \textbf{0.843 (0.822-0.864)} \\
\bottomrule
\end{tabular}
\caption{\textbf{Label efficiency generalization performance for dermoscopic image-based skin cancer diagnosis on BCN20000 dataset.} Metrics: W\_F1 (Weighted F1), AUROC, BACC (Balanced Accuracy), AUPR (Area Under Precision-Recall Curve). The best model for each setting is bolded. 95\% CI in parentheses. $^{*}p<0.05$, $^{**}p<0.01$, $^{***}p<0.001$ compared to PanDerm.}
\label{tab16}
\end{table}

\begin{table}
\footnotesize\centering
\setlength{\tabcolsep}{3pt}
\begin{tabular}{llllll}
\toprule
Percent & Model & W\_F1 & AUROC & BACC & AUPR \\
\midrule\midrule
\multirow{4}{*}{5\%} 
 & SL\_ImageNet & 0.826 (0.802-0.851) & 0.681 (0.593-0.769)$^{*}$ & 0.548 (0.502-0.594) & 0.329 (0.214-0.444) \\
 & DINOv2 & 0.825 (0.798-0.853) & 0.706 (0.620-0.791) & 0.553 (0.502-0.605) & 0.331 (0.219-0.444)$^{*}$ \\
 & SwAVDerm & 0.839 (0.812-0.866) & 0.660 (0.567-0.754)$^{*}$ & 0.572 (0.518-0.626) & 0.355 (0.231-0.480) \\
 & PanDerm & \textbf{0.842 (0.811-0.874)} & \textbf{0.749 (0.663-0.834)} & \textbf{0.594 (0.534-0.654)} & \textbf{0.431 (0.299-0.562)} \\
\midrule
\multirow{4}{*}{10\%} 
 & SL\_ImageNet & \textbf{0.858 (0.829-0.888)} & 0.784 (0.708-0.861) & 0.622 (0.556-0.688) & 0.470 (0.345-0.595) \\
 & DINOv2 & 0.853 (0.822-0.884) & \textbf{0.798 (0.721-0.875)} & 0.626 (0.557-0.694) & 0.461 (0.322-0.600) \\
 & SwAVDerm & 0.858 (0.826-0.889) & 0.732 (0.647-0.816) & \textbf{0.653 (0.581-0.725)} & 0.390 (0.263-0.517) \\
 & PanDerm & 0.857 (0.826-0.888) & 0.795 (0.712-0.878) & 0.652 (0.583-0.721) & \textbf{0.480 (0.347-0.613)} \\
\midrule
\multirow{4}{*}{20\%} 
 & SL\_ImageNet & 0.847 (0.818-0.876) & 0.802 (0.739-0.865)$^{***}$ & 0.612 (0.545-0.679)$^{*}$ & 0.458 (0.332-0.584) \\
 & DINOv2 & \textbf{0.908 (0.879-0.937)} & 0.799 (0.726-0.833) & \textbf{0.779 (0.708-0.851)} & \textbf{0.629 (0.500-0.758)} \\
 & SwAVDerm & 0.841 (0.811-0.872)$^{*}$ & 0.798 (0.724-0.872)$^{**}$ & 0.607 (0.541-0.673)$^{*}$ & 0.461 (0.333-0.589) \\
 & PanDerm & 0.879 (0.848-0.910) & \textbf{0.885 (0.839-0.930)} & 0.718 (0.644-0.791) & 0.573 (0.443-0.703) \\
\midrule
\multirow{4}{*}{30\%} 
 & SL\_ImageNet & 0.851 (0.821-0.881) & 0.851 (0.794-0.908)$^{*}$ & 0.631 (0.561-0.701)$^{*}$ & 0.506 (0.384-0.628) \\
 & DINOv2 & 0.869 (0.837-0.902) & 0.854 (0.794-0.914)$^{*}$ & 0.709 (0.638-0.781) & 0.580 (0.444-0.715) \\
 & SwAVDerm & \textbf{0.884 (0.854-0.915)} & 0.814 (0.745-0.883)$^{**}$ & 0.679 (0.609-0.748) & 0.508 (0.372-0.643) \\
 & PanDerm & 0.874 (0.842-0.906) & \textbf{0.899 (0.858-0.939)} & \textbf{0.731 (0.657-0.805)} & \textbf{0.601 (0.465-0.737)} \\
\midrule
\multirow{4}{*}{50\%} 
 & SL\_ImageNet & 0.872 (0.842-0.903) & 0.839 (0.778-0.899)$^{**}$ & 0.679 (0.607-0.752) & 0.524 (0.399-0.650)$^{*}$ \\
 & DINOv2 & 0.882 (0.850-0.913) & 0.831 (0.766-0.896)$^{**}$ & 0.718 (0.644-0.793) & 0.540 (0.403-0.677)$^{*}$ \\
 & SwAVDerm & 0.873 (0.841-0.904) & 0.820 (0.757-0.883)$^{**}$ & 0.687 (0.614-0.760) & 0.515 (0.382-0.648)$^{**}$ \\
 & PanDerm & \textbf{0.890 (0.860-0.920)} & \textbf{0.915 (0.877-0.953)} & \textbf{0.750 (0.677-0.823)} & \textbf{0.688 (0.570-0.805)} \\
\midrule
\multirow{4}{*}{100\%} 
 & SL\_ImageNet & 0.873 (0.841-0.904)$^{*}$ & 0.878 (0.823-0.933)$^{**}$ & 0.681 (0.610-0.751)$^{**}$ & 0.638 (0.524-0.752)$^{*}$ \\
 & DINOv2 & 0.879 (0.848-0.911)$^{*}$ & 0.877 (0.820-0.934)$^{***}$ & 0.734 (0.659-0.808) & 0.631 (0.507-0.754)$^{*}$ \\
 & SwAVDerm & 0.860 (0.828-0.891)$^{***}$ & 0.848 (0.791-0.905)$^{***}$ & 0.660 (0.590-0.731)$^{**}$ & 0.538 (0.414-0.661)$^{***}$ \\
 & PanDerm & \textbf{0.912 (0.884-0.940)} & \textbf{0.949 (0.922-0.976)} & \textbf{0.774 (0.700-0.847)} & \textbf{0.771 (0.660-0.881)} \\
\bottomrule
\end{tabular}
\caption{\textbf{Label efficiency generalization performance for dermoscopic image-based melanoma detection on HIBA dataset.} Metrics: W\_F1 (Weighted F1), AUROC, BACC (Balanced Accuracy), AUPR (Area Under Precision-Recall Curve). The best model for each setting is bolded. 95\% CI in parentheses. $^{*}p<0.05$, $^{**}p<0.01$, $^{***}p<0.001$ compared to PanDerm.}
\label{tab17}
\end{table}

\begin{table}
\footnotesize\centering
\setlength{\tabcolsep}{4pt}
\begin{tabular}{llllll}
\toprule
Percent & Model & W\_F1 & AUROC & BACC & AUPR \\
\midrule\midrule
\multirow{4}{*}{5\%} 
 & SL\_ImageNet & 0.693 (0.622-0.763) & 0.689 (0.602-0.777) & 0.630 (0.563-0.696) & 0.591 (0.483-0.700) \\
 & DINOv2 & 0.697 (0.632-0.762) & 0.713 (0.625-0.801) & 0.653 (0.580-0.726) & 0.654 (0.546-0.762) \\
 & SwAVDerm & 0.589 (0.533-0.645)$^{***}$ & 0.609 (0.519-0.700) & 0.530 (0.475-0.585)$^{***}$ & 0.478 (0.376-0.580)$^{*}$ \\
 & PanDerm & \textbf{0.707 (0.642-0.772)} & \textbf{0.674 (0.587-0.762)} & \textbf{0.655 (0.585-0.725)} & \textbf{0.595 (0.495-0.695)} \\
\midrule
\multirow{4}{*}{10\%} 
 & SL\_ImageNet & 0.677 (0.610-0.743) & 0.693 (0.607-0.779)$^{**}$ & 0.617 (0.551-0.683) & 0.552 (0.444-0.660)$^{*}$ \\
 & DINOv2 & 0.722 (0.655-0.788) & 0.771 (0.697-0.845) & 0.683 (0.611-0.755) & 0.673 (0.568-0.777) \\
 & SwAVDerm & 0.609 (0.544-0.674)$^{***}$ & 0.619 (0.526-0.712)$^{***}$ & 0.550 (0.477-0.622)$^{***}$ & 0.464 (0.367-0.560)$^{***}$ \\
 & PanDerm & \textbf{0.738 (0.674-0.802)} & \textbf{0.785 (0.714-0.856)} & \textbf{0.682 (0.612-0.753)} & \textbf{0.666 (0.556-0.777)} \\
\midrule
\multirow{4}{*}{20\%} 
 & SL\_ImageNet & 0.672 (0.610-0.734)$^{*}$ & 0.738 (0.655-0.821) & 0.608 (0.544-0.672)$^{*}$ & 0.601 (0.487-0.716) \\
 & DINOv2 & \textbf{0.747 (0.685-0.809)} & \textbf{0.791 (0.715-0.867)} & \textbf{0.694 (0.622-0.765)} & 0.662 (0.553-0.770) \\
 & SwAVDerm & 0.634 (0.572-0.697)$^{**}$ & 0.641 (0.556-0.725)$^{**}$ & 0.574 (0.508-0.640)$^{**}$ & 0.489 (0.392-0.586)$^{***}$ \\
 & PanDerm & 0.731 (0.670-0.793) & 0.765 (0.691-0.838) & 0.672 (0.603-0.740) & \textbf{0.668 (0.565-0.771)} \\
\midrule
\multirow{4}{*}{30\%} 
 & SL\_ImageNet & 0.714 (0.653-0.775) & 0.772 (0.696-0.849) & 0.654 (0.591-0.717) & 0.662 (0.544-0.781) \\
 & DINOv2 & 0.728 (0.663-0.794) & 0.813 (0.746-0.879) & \textbf{0.681 (0.609-0.753)} & 0.721 (0.629-0.813) \\
 & SwAVDerm & 0.636 (0.571-0.700)$^{**}$ & 0.644 (0.555-0.733)$^{***}$ & 0.577 (0.507-0.647)$^{**}$ & 0.496 (0.390-0.602)$^{***}$ \\
 & PanDerm & \textbf{0.729 (0.666-0.792)} & \textbf{0.816 (0.751-0.882)} & 0.676 (0.604-0.748) & \textbf{0.722 (0.628-0.817)} \\
\midrule
\multirow{4}{*}{50\%} 
 & SL\_ImageNet & 0.720 (0.655-0.785) & 0.772 (0.703-0.842) & 0.661 (0.593-0.730) & 0.677 (0.574-0.781) \\
 & DINOv2 & 0.726 (0.661-0.792) & \textbf{0.810 (0.743-0.877)} & 0.675 (0.602-0.747) & \textbf{0.722 (0.630-0.813)} \\
 & SwAVDerm & 0.686 (0.617-0.755) & 0.693 (0.610-0.775)$^{**}$ & 0.635 (0.562-0.708) & 0.541 (0.436-0.647)$^{**}$ \\
 & PanDerm & \textbf{0.741 (0.676-0.806)} & 0.804 (0.739-0.870) & \textbf{0.687 (0.616-0.758)} & 0.716 (0.625-0.808) \\
\midrule
\multirow{4}{*}{100\%} 
 & SL\_ImageNet & 0.754 (0.687-0.822) & 0.810 (0.744-0.875)$^{*}$ & 0.703 (0.632-0.774) & 0.704 (0.595-0.813)$^{*}$ \\
 & DINOv2 & 0.764 (0.699-0.829) & 0.797 (0.723-0.870)$^{**}$ & 0.723 (0.653-0.793) & 0.733 (0.646-0.820) \\
 & SwAVDerm & 0.747 (0.686-0.809) & 0.777 (0.702-0.853)$^{**}$ & 0.696 (0.625-0.767) & 0.660 (0.544-0.776)$^{**}$ \\
 & PanDerm & \textbf{0.802 (0.743-0.860)} & \textbf{0.878 (0.824-0.931)} & \textbf{0.767 (0.700-0.833)} & \textbf{0.799 (0.714-0.884)} \\
\bottomrule
\end{tabular}
\caption{\textbf{Label efficiency generalization performance for clinical image-based melanoma detection on DermC dataset.} Metrics: W\_F1 (Weighted F1), AUROC, BACC (Balanced Accuracy), AUPR (Area Under Precision-Recall Curve). The best model for each setting is bolded. 95\% CI in parentheses. $^{*}p<0.05$, $^{**}p<0.01$, $^{***}p<0.001$ compared to PanDerm.}
\label{tab18}
\end{table}

\begin{table}
\footnotesize\centering
\setlength{\tabcolsep}{4pt}
\begin{tabular}{llllll}
\toprule
Percent & Model & W\_F1 & AUROC & BACC & AUPR \\
\midrule\midrule
\multirow{4}{*}{5\%} 
 & SL\_ImageNet & 0.566 (0.526-0.605) & 0.803 (0.775-0.832) & 0.395 (0.338-0.452) & 0.598 (0.555-0.641)$^{***}$ \\
 & DINOv2 & 0.560 (0.521-0.599)$^{*}$ & 0.804 (0.776-0.833) & 0.393 (0.354-0.432)$^{***}$ & 0.607 (0.563-0.651)$^{***}$ \\
 & SwAVDerm & 0.468 (0.425-0.510)$^{***}$ & 0.695 (0.661-0.729)$^{***}$ & 0.328 (0.279-0.378)$^{***}$ & 0.485 (0.442-0.528)$^{***}$ \\
 & PanDerm & \textbf{0.607 (0.569-0.646)} & \textbf{0.835 (0.808-0.861)} & \textbf{0.504 (0.432-0.576)} & \textbf{0.675 (0.636-0.715)} \\
\midrule
\multirow{4}{*}{10\%} 
 & SL\_ImageNet & 0.580 (0.540-0.621)$^{***}$ & 0.824 (0.795-0.852)$^{***}$ & 0.431 (0.375-0.487)$^{**}$ & 0.630 (0.584-0.675)$^{***}$ \\
 & DINOv2 & 0.598 (0.557-0.638)$^{***}$ & 0.835 (0.810-0.861)$^{***}$ & 0.432 (0.388-0.477)$^{***}$ & 0.652 (0.608-0.695)$^{***}$ \\
 & SwAVDerm & 0.515 (0.474-0.556)$^{***}$ & 0.753 (0.721-0.785)$^{***}$ & 0.375 (0.317-0.433)$^{***}$ & 0.550 (0.503-0.597)$^{***}$ \\
 & PanDerm & \textbf{0.673 (0.632-0.713)} & \textbf{0.877 (0.854-0.900)} & \textbf{0.560 (0.490-0.630)} & \textbf{0.733 (0.691-0.774)} \\
\midrule
\multirow{4}{*}{20\%} 
 & SL\_ImageNet & 0.601 (0.561-0.640)$^{***}$ & 0.831 (0.803-0.858)$^{***}$ & 0.486 (0.415-0.556)$^{*}$ & 0.661 (0.616-0.705)$^{***}$ \\
 & DINOv2 & 0.621 (0.581-0.661)$^{**}$ & 0.843 (0.817-0.869)$^{***}$ & 0.472 (0.408-0.536)$^{***}$ & 0.678 (0.634-0.723)$^{***}$ \\
 & SwAVDerm & 0.535 (0.494-0.576)$^{***}$ & 0.773 (0.742-0.805)$^{***}$ & 0.398 (0.341-0.456)$^{***}$ & 0.579 (0.535-0.623)$^{***}$ \\
 & PanDerm & \textbf{0.685 (0.645-0.726)} & \textbf{0.882 (0.860-0.904)} & \textbf{0.598 (0.526-0.670)} & \textbf{0.740 (0.699-0.781)} \\
\midrule
\multirow{4}{*}{30\%} 
 & SL\_ImageNet & 0.639 (0.597-0.681)$^{***}$ & 0.854 (0.827-0.880)$^{***}$ & 0.493 (0.426-0.560)$^{***}$ & 0.702 (0.657-0.747)$^{***}$ \\
 & DINOv2 & 0.636 (0.596-0.675)$^{***}$ & 0.855 (0.830-0.880)$^{***}$ & 0.477 (0.412-0.543)$^{***}$ & 0.699 (0.658-0.740)$^{***}$ \\
 & SwAVDerm & 0.577 (0.532-0.621)$^{***}$ & 0.806 (0.778-0.835)$^{***}$ & 0.461 (0.390-0.532)$^{***}$ & 0.624 (0.578-0.671)$^{***}$ \\
 & PanDerm & \textbf{0.729 (0.689-0.768)} & \textbf{0.910 (0.891-0.929)} & \textbf{0.645 (0.573-0.716)} & \textbf{0.791 (0.753-0.829)} \\
\midrule
\multirow{4}{*}{50\%} 
 & SL\_ImageNet & 0.634 (0.592-0.675)$^{***}$ & 0.861 (0.835-0.887)$^{***}$ & 0.544 (0.472-0.615)$^{***}$ & 0.707 (0.663-0.752)$^{***}$ \\
 & DINOv2 & 0.656 (0.614-0.698)$^{***}$ & 0.867 (0.840-0.893)$^{***}$ & 0.565 (0.494-0.636)$^{***}$ & 0.716 (0.675-0.756)$^{***}$ \\
 & SwAVDerm & 0.595 (0.554-0.637)$^{***}$ & 0.826 (0.798-0.854)$^{***}$ & 0.490 (0.422-0.558)$^{***}$ & 0.651 (0.607-0.695)$^{***}$ \\
 & PanDerm & \textbf{0.768 (0.730-0.806)} & \textbf{0.920 (0.900-0.939)} & \textbf{0.729 (0.665-0.793)} & \textbf{0.816 (0.777-0.854)} \\
\midrule
\multirow{4}{*}{100\%} 
 & SL\_ImageNet & 0.658 (0.617-0.699)$^{***}$ & 0.878 (0.854-0.902)$^{***}$ & 0.576 (0.504-0.649)$^{***}$ & 0.747 (0.707-0.788)$^{***}$ \\
 & DINOv2 & 0.695 (0.653-0.737)$^{**}$ & 0.880 (0.855-0.905)$^{***}$ & 0.575 (0.505-0.646)$^{***}$ & 0.747 (0.706-0.787)$^{***}$ \\
 & SwAVDerm & 0.664 (0.624-0.704)$^{***}$ & 0.858 (0.833-0.882)$^{***}$ & 0.539 (0.473-0.606)$^{***}$ & 0.709 (0.667-0.750)$^{***}$ \\
 & PanDerm & \textbf{0.758 (0.720-0.796)} & \textbf{0.931 (0.914-0.947)} & \textbf{0.710 (0.644-0.776)} & \textbf{0.844 (0.812-0.876)} \\
\bottomrule
\end{tabular}
\caption{\textbf{Label efficiency generalization performance for dermoscopic image-based skin condition classification on PAD dataset.} Metrics: W\_F1 (Weighted F1), AUROC, BACC (Balanced Accuracy), AUPR (Area Under Precision-Recall Curve). The best model for each setting is bolded. 95\% CI in parentheses. $^{*}p<0.05$, $^{**}p<0.01$, $^{***}p<0.001$ compared to PanDerm.}
\label{tab19}
\end{table}

\begin{table}
\footnotesize\centering
\setlength{\tabcolsep}{4pt}
\begin{tabular}{llllll}
\toprule
Percent & Model & W\_F1 & AUROC & BACC & AUPR \\
\midrule\midrule
\multirow{4}{*}{5\%} 
 & SL\_ImageNet & 0.805 (0.799-0.812)$^{***}$ & 0.984 (0.983-0.985)$^{***}$ & 0.738 (0.711-0.765)$^{***}$ & 0.886 (0.880-0.891)$^{***}$ \\
 & DINOv2 & 0.835 (0.829-0.841)$^{***}$ & 0.986 (0.985-0.987)$^{***}$ & 0.756 (0.736-0.776)$^{***}$ & 0.905 (0.900-0.909)$^{***}$ \\
 & SwAVDerm & 0.738 (0.731-0.745)$^{***}$ & 0.968 (0.966-0.969)$^{***}$ & 0.673 (0.647-0.698)$^{***}$ & 0.806 (0.799-0.813)$^{***}$ \\
 & PanDerm & \textbf{0.867 (0.861-0.872)} & \textbf{0.991 (0.991-0.992)} & \textbf{0.844 (0.836-0.853)} & \textbf{0.931 (0.927-0.935)} \\
\midrule
\multirow{4}{*}{10\%} 
 & SL\_ImageNet & 0.831 (0.824-0.837)$^{***}$ & 0.988 (0.987-0.988)$^{***}$ & 0.789 (0.768-0.809)$^{***}$ & 0.908 (0.903-0.913)$^{***}$ \\
 & DINOv2 & 0.854 (0.848-0.860)$^{***}$ & 0.989 (0.988-0.989)$^{***}$ & 0.790 (0.764-0.816)$^{***}$ & 0.916 (0.911-0.921)$^{***}$ \\
 & SwAVDerm & 0.773 (0.766-0.780)$^{***}$ & 0.975 (0.974-0.976)$^{***}$ & 0.704 (0.679-0.729)$^{***}$ & 0.842 (0.835-0.849)$^{***}$ \\
 & PanDerm & \textbf{0.877 (0.872-0.882)} & \textbf{0.992 (0.992-0.993)} & \textbf{0.859 (0.850-0.867)} & \textbf{0.937 (0.933-0.941)} \\
\midrule
\multirow{4}{*}{20\%} 
 & SL\_ImageNet & 0.851 (0.845-0.857)$^{***}$ & 0.990 (0.989-0.991)$^{***}$ & 0.799 (0.772-0.826)$^{***}$ & 0.924 (0.919-0.928)$^{***}$ \\
 & DINOv2 & 0.869 (0.863-0.874)$^{***}$ & 0.991 (0.990-0.991)$^{***}$ & 0.812 (0.787-0.838)$^{***}$ & 0.930 (0.926-0.935)$^{***}$ \\
 & SwAVDerm & 0.780 (0.774-0.787)$^{***}$ & 0.977 (0.976-0.979)$^{***}$ & 0.708 (0.687-0.729)$^{***}$ & 0.854 (0.848-0.861)$^{***}$ \\
 & PanDerm & \textbf{0.884 (0.879-0.889)} & \textbf{0.993 (0.993-0.994)} & \textbf{0.874 (0.866-0.882)} & \textbf{0.945 (0.942-0.949)} \\
\midrule
\multirow{4}{*}{30\%} 
 & SL\_ImageNet & 0.862 (0.857-0.868)$^{***}$ & 0.991 (0.990-0.991)$^{***}$ & 0.803 (0.778-0.827)$^{***}$ & 0.929 (0.925-0.934)$^{***}$ \\
 & DINOv2 & 0.879 (0.873-0.884)$^{***}$ & 0.992 (0.991-0.992)$^{***}$ & 0.827 (0.801-0.852)$^{*}$ & 0.937 (0.933-0.941)$^{***}$ \\
 & SwAVDerm & 0.793 (0.786-0.799)$^{***}$ & 0.980 (0.978-0.981)$^{***}$ & 0.710 (0.702-0.719)$^{***}$ & 0.863 (0.857-0.870)$^{***}$ \\
 & PanDerm & \textbf{0.896 (0.891-0.901)} & \textbf{0.994 (0.993-0.994)} & \textbf{0.866 (0.841-0.891)} & \textbf{0.947 (0.943-0.951)} \\
\midrule
\multirow{4}{*}{50\%} 
 & SL\_ImageNet & 0.865 (0.859-0.871)$^{***}$ & 0.991 (0.990-0.992)$^{***}$ & 0.809 (0.784-0.833)$^{***}$ & 0.931 (0.927-0.935)$^{***}$ \\
 & DINOv2 & 0.885 (0.879-0.890)$^{***}$ & 0.992 (0.991-0.993)$^{***}$ & 0.821 (0.801-0.841)$^{***}$ & 0.938 (0.934-0.942)$^{***}$ \\
 & SwAVDerm & 0.798 (0.792-0.805)$^{***}$ & 0.981 (0.980-0.982)$^{***}$ & 0.718 (0.709-0.727)$^{***}$ & 0.871 (0.865-0.877)$^{***}$ \\
 & PanDerm & \textbf{0.900 (0.895-0.904)} & \textbf{0.994 (0.993-0.995)} & \textbf{0.872 (0.848-0.897)} & \textbf{0.948 (0.944-0.951)} \\
\midrule
\multirow{4}{*}{100\%} 
 & SL\_ImageNet & 0.869 (0.863-0.875)$^{***}$ & 0.992 (0.991-0.992)$^{***}$ & 0.829 (0.803-0.856)$^{***}$ & 0.934 (0.930-0.938)$^{***}$ \\
 & DINOv2 & 0.890 (0.885-0.895)$^{***}$ & 0.993 (0.992-0.993)$^{***}$ & 0.820 (0.813-0.828)$^{***}$ & 0.942 (0.938-0.947)$^{**}$ \\
 & SwAVDerm & -- & -- & -- & -- \\
 & PanDerm & \textbf{0.901 (0.896-0.905)} & \textbf{0.994 (0.993-0.994)} & \textbf{0.878 (0.854-0.902)} & \textbf{0.947 (0.943-0.950)} \\
\bottomrule
\end{tabular}
\caption{\textbf{Label efficiency generalization performance for fine-grained skin tumor classification on PATCH16 dataset.} Metrics: W\_F1 (Weighted F1), AUROC, BACC (Balanced Accuracy), AUPR (Area Under Precision-Recall Curve). The best model for each setting is bolded. 95\% CI in parentheses. $^{*}p<0.05$, $^{**}p<0.01$, $^{***}p<0.001$ compared to PanDerm. SwAVDerm results for the 100\% setting are missing.}
\label{tab20}
\end{table}

\begin{table}
\footnotesize\centering
\setlength{\tabcolsep}{4pt}
\begin{tabular}{llll}
\toprule
Dataset & Skin Tone & W\_F1 & Sensitivity \\
\midrule\midrule
\multirow{3}{*}{F17K} 
 & FST I-II (n=1195) & 0.825 (0.825-0.825) & 0.835 (0.835-0.835) \\
 & FST III-IV (n=786) & 0.840 (0.840-0.840) & 0.851 (0.851-0.851) \\
 & FST V-VI (n=238) & 0.864 (0.864-0.864) & 0.878 (0.878-0.878) \\
\midrule
\multirow{3}{*}{DDI} 
 & FST I-II (n=40) & 0.780 (0.780-0.780) & 0.750 (0.750-0.750) \\
 & FST III-IV (n=59) & 0.818 (0.818-0.818) & 0.814 (0.814-0.814) \\
 & FST V-VI (n=38) & 0.854 (0.854-0.854) & 0.842 (0.842-0.842) \\
\bottomrule
\end{tabular}
\caption{\textbf{PanDerm performance across different skin tones on Fitzpatrick17k and DDI datasets}. The table shows the performance metrics for PanDerm on F17K and DDI datasets, stratified by Fitzpatrick Skin Type (FST) groups. Metrics include Weighted F1 score (W\_F1) and Sensitivity. 95\% CI in parentheses (identical to point estimate due to single measurement).}
\label{tab21}
\end{table}

\begin{table}
\footnotesize\centering
\setlength{\tabcolsep}{3pt}
\begin{tabular}{lllll}
\toprule
Category & Subgroup & n & W\_F1 & Sensitivity \\
\midrule\midrule
Overall & All & 1232 & 0.957 & 0.959 \\
\midrule
\multirow{2}{*}{Sex} 
 & Female & 563 & 0.963 & 0.965 \\
 & Male & 659 & 0.951 & 0.953 \\
\midrule
\multirow{15}{*}{Location} 
 & Face & 70 & 0.862 & 0.871 \\
 & Lower extremity & 240 & 0.968 & 0.971 \\
 & Abdomen & 126 & 0.984 & 0.984 \\
 & Upper extremity & 120 & 0.967 & 0.967 \\
 & Back & 256 & 0.919 & 0.922 \\
 & Trunk & 259 & 0.993 & 0.992 \\
 & Scalp & 8 & 1.000 & 1.000 \\
 & Hand & 18 & 1.000 & 1.000 \\
 & Unknown & 40 & 1.000 & 1.000 \\
 & Chest & 33 & 0.879 & 0.879 \\
 & Neck & 18 & 1.000 & 1.000 \\
 & Foot & 30 & 0.950 & 0.967 \\
 & Genital & 8 & 1.000 & 1.000 \\
\midrule
\multirow{3}{*}{Age} 
 & Old & 517 & 0.923 & 0.927 \\
 & Medium & 669 & 0.984 & 0.984 \\
 & Young & 34 & 0.913 & 0.941 \\
\bottomrule
\end{tabular}
\caption{\textbf{Model robustness analysis across subgroups on HAM10000 dataset}. The table shows performance metrics across different subgroups based on sex, location, and age. Metrics include sample size (n), Weighted F1 score (W\_F1), and Sensitivity. All metric values are point estimates.}
\label{tab22}
\end{table}
\begin{table}
\footnotesize\centering
\setlength{\tabcolsep}{4pt}
\begin{tabular}{llll}
\toprule
Study & Mean Recall & Recall\_Mel & Recall\_BCC \\
\midrule
Tschandl et al. 2019 \cite{humanai} & 0.777 & 0.614 & 0.796 \\
PanDerm & \textbf{0.804} & \textbf{0.877} & \textbf{0.860} \\
\bottomrule
\end{tabular}
\caption{\textcolor{black}{\textbf{Performance comparison with prior Human-AI collaboration work in HAM10000}. Comparison of model performance between PanDerm and \cite{humanai} on HAM10000 dataset. PanDerm shows improvements in overall recall (+2.72\%) and particularly significant gains for critical cancers like melanoma (+26.32\%) and BCC (+6.42\%).}}
\label{tab_humanAIcomparison}
\end{table}
\begin{table}
\footnotesize\centering
\setlength{\tabcolsep}{4pt}
\begin{tabular}{ll}
\toprule
Condition & Accuracy \\
\midrule
Without AI assistance & 0.69 (0.65-0.73) \\
With AI assistance & \textbf{0.80 (0.76-0.84)$^{***}$} \\
\bottomrule
\end{tabular}
\caption{\textbf{Human AI collaboration performance comparison on skin cancer classification using HAM10000 dataset}. Comparison of accuracy with and without AI assistance. 95\% CI in parentheses. $^{***}p=6.53 \times 10^{-8}$ compared to performance without AI assistance.}
\label{tab23}
\end{table}

\begin{table}
\footnotesize\centering
\setlength{\tabcolsep}{4pt}
\begin{tabular}{lllll}
\toprule
Class & Without AI & With AI & p-value & Corrected p-value \\
\midrule
AKIEC & 0.51 (0.44-0.59) & \textbf{0.67 (0.60-0.74)}$^{**}$ & 0.0036 & 0.0254 \\
BCC & 0.76 (0.69-0.82) & \textbf{0.84 (0.79-0.90)} & 0.0545 & 0.3812 \\
BKL & 0.57 (0.50-0.65) & \textbf{0.77 (0.71-0.84)}$^{***}$ & 0.0001 & 0.0007 \\
DF & 0.70 (0.63-0.77) & \textbf{0.81 (0.75-0.87)}$^{*}$ & 0.0151 & 0.1059 \\
MEL & 0.69 (0.64-0.74) & \textbf{0.83 (0.79-0.87)}$^{***}$ & $<$0.0001 & 0.0003 \\
NV & 0.83 (0.79-0.86) & \textbf{0.86 (0.82-0.89)} & 0.1948 & 1.0000 \\
VASC & 0.93 (0.90-0.97) & \textbf{0.95 (0.92-0.98)} & 0.4799 & 1.0000 \\
\bottomrule
\end{tabular}
\caption{\textbf{Class-specific performance comparison on human-AI collaboration study}. Comparison of accuracy with and without AI assistance for each class. 95\% CI in parentheses. $^{*}p<0.05$, $^{**}p<0.01$, $^{***}p<0.001$  compared to performance without AI assistance.}
\label{tab24}
\end{table}

\begin{table}
\footnotesize\centering
\setlength{\tabcolsep}{4pt}
\begin{tabular}{lllll}
\toprule
Experience Group & Without AI & With AI & p-value & Corrected p-value \\
\midrule
Low & 0.64 (0.58-0.70) & \textbf{0.83 (0.78-0.88)}$^{**}$ & 0.0082 & 0.0246 \\
Medium & 0.67 (0.62-0.72) & \textbf{0.79 (0.75-0.83)}$^{***}$ & $<$0.0001 & $<$0.0001 \\
High & 0.78 (0.75-0.81) & \textbf{0.84 (0.82-0.86) }$^{**}$& 0.0390 & 0.1170 \\
\bottomrule
\end{tabular}
\caption{\textbf{Performance by experience level on human-AI collaboration study}. Comparison of accuracy with and without AI assistance for each experience group. 95\% CI in parentheses. $^{*}p<0.05$, $^{**}p<0.01$, $^{***}p<0.001$ (Bonferroni corrected) compared to performance without AI assistance.}
\label{tab25}
\end{table}

\begin{table}
\footnotesize\centering
\setlength{\tabcolsep}{4pt}
\begin{tabular}{ll|ccc}
\toprule
Model & Metric & 3-year & 5-year & 7-year \\
\midrule\midrule
\multirow{2}{*}{PanDerm\_multi} & AUC & 0.9463 (0.9004-0.9922) & \textbf{0.9462 (0.9125-0.9799)} & \textbf{0.9303 (0.8953-0.9654)} \\
 & p-value & 0.0000 & 0.0000 & 0.0000 \\
\midrule
\multirow{2}{*}{Multi-clinical variables} & AUC & 0.8819 (0.7960-0.9279) & 0.9018 (0.8016-0.9420) & 0.8587 (0.7997-0.9177) \\
 & p-value & 0.1601 & 0.1521 & 0.1475 \\
\midrule
\multirow{2}{*}{Single clinical variables} & AUC & 0.8396 (0.7999-0.8793) & 0.8473 (0.7859-0.9086) & 0.8021 (0.7235-0.9007) \\
 & p-value & 0.0788 & 0.0225 & 0.0480 \\
\midrule
\multirow{2}{*}{PanDerm\_single} & AUC & \textbf{0.9501 (0.9095-0.9908)} & 0.9312 (0.8868-0.9755) & 0.9087 (0.8803-0.9371) \\
 & p-value & 0.4127 & 0.0871 & 0.0756 \\
\bottomrule
\end{tabular}
\caption{\textbf{Survival analysis performance comparison between PanDerm and clinical variables on CombinMel dataset}. Methods include PanDerm\_multi, Multi-clinical variables, Single clinical variables, and PanDerm\_single. Metrics: Area Under the Time-dependent ROC Curve (AUC) at 3, 5, and 7 years with corresponding p-values. The best-performing model for each time point is bolded. 95\% CI in parentheses.}
\label{tab26}
\end{table}

\begin{table}
\footnotesize\centering
\setlength{\tabcolsep}{4pt}
\begin{tabular}{l|ccc}
\toprule
Model & 3-year AUC & 5-year AUC & 7-year AUC \\
\midrule\midrule
PanDerm & \textbf{0.9501 (0.9095-0.9908)} & \textbf{0.9312 (0.8868-0.9755)} & \textbf{0.9087 (0.8803-0.9371)} \\
 & p = 0.0000 & p = 0.0000 & p = 0.0000 \\
\midrule
DINOv2 & 0.9276 (0.8903-0.9650) & 0.9014 (0.8273-0.9754) & 0.8836 (0.7981-0.9691) \\
 & p = 0.2901 & p = 0.3659 & p = 0.4613 \\
\midrule
SwavDerm & 0.9234 (0.8715-0.9754) & 0.8824 (0.7943-0.9704) & 0.8664 (0.7776-0.9552) \\
 & p = 0.2932 & p = 0.2066 & p = 0.2432 \\
\midrule
SL\_ImageNet & 0.9195 (0.8502-0.9889) & 0.8858 (0.7977-0.9738) & 0.8785 (0.7824-0.9747) \\
 & p = 0.3215 & p = 0.2370 & p = 0.4280 \\
\bottomrule
\end{tabular}
\caption{\textbf{Survival analysis performance comparison of different models on CombinMel dataset}. Models include PanDerm, DINOv2, SwavDerm, and SL\_ImageNet. Metrics: Area Under the Time-dependent ROC Curve (AUC) at 3, 5, and 7 years with corresponding p-values. The best-performing model for each time point is bolded. 95\% CI in parentheses.}
\label{tab:survival_d2}
\end{table}

\begin{table}
\footnotesize\centering
\setlength{\tabcolsep}{3pt}
\begin{tabular}{l|cccccccc|c}
\toprule
Model & ISIC19b & HAM\_c & BCN20000 & PAD & Derm7pt\_c & Dermnet & MCSI & TBP\_solar & Average \\
\midrule
CLIP\_base (teacher) & 88.07 & 96.13 & 91.09 & 91.03 & 77.10 & 89.92 & 99.87 & 94.29 & 90.94 \\
CLIP\_large (teacher) & 89.44 & 96.34 & 93.12 & 92.58 & 80.18 & 90.88 & 99.85 & 95.31 & 92.21 \\
MONET\_large (teacher) & 89.05 & 96.65 & 94.03 & 92.55 & 75.17 & 91.41 & 99.85 & 95.59 & 91.79 \\
BIOMED\_CLIP\_base (teacher) & 74.71 & 82.59 & 80.82 & 88.16 & 60.95 & 85.60 & 97.12 & 80.29 & 81.28 \\
\hline
Base model & 86.69 & 96.36 & 92.75 & 94.20 & 82.91 & 93.33 & \textbf{99.88} & 95.15 & 92.66 \\
+ BiomedCLIP\_base & 86.63 & 97.06 & 93.27 & 91.59 & 82.09 & 90.96 & 98.59 & 95.29 & 91.94 \\
+ CLIP\_base & 88.78 & 97.21 & 93.55 & 92.42 & 79.50 & 93.13 & 99.87 & 95.63 & 92.51 \\
\rowcolor{gray!10} + CLIP\_large & \textbf{89.26} & \textbf{98.03} & \textbf{95.60} & \textbf{94.61} & \textbf{87.85} & \textbf{94.97} & \textbf{99.88} & \textbf{96.80} & \textbf{94.63} \\
\bottomrule
\end{tabular}
\caption{\textbf{Ablation on target representation (teacher models) across various dermatology datasets}. Models include CLIP variants, MONET, BIOMED\_CLIP, and PanDerm with different pretraining strategies. Metrics represent accuracy percentages. The best-performing model for each dataset is bolded and highlighted. Datasets vary in modality and size: ISIC19b, HAM\_c, BCN20000 (derm, 20k/10k/12k), PAD, Derm7pt\_c, Dermnet, MCSI (clinic, 2k/839/19k/400), TBP\_solar (TBP, 6k). It shows that CLIP-large pretrained on the natural domain can outperform biomedical-specific CLIP (BiomedCLIP) and dermatology-specific CLIP (MONET). This can be attributed to the limited data scale of skin images in medical domain CLIP models. Thus, CLIP-large remains the best teacher model for creating target representations for masked image modeling in dermatology. When incorporating CLIP-large as the teacher model, it significantly improved the base model (+ 1.97 on average) and also outperformed the teacher model itself (+2.42 on average).}
\label{tab:target_rep}
\end{table}

\begin{table}
\footnotesize\centering
\setlength{\tabcolsep}{3pt}
\begin{tabular}{l|ccccccc|c|c}
\toprule
Model & HAM\_c & BCN20000 & PAD & Derm7pt\_c & Dermnet & MCSI & TBP\_solar & Average & Training time \\
\midrule
PanDerm (FT) & \textbf{98.03} & \textbf{97.65} & 93.59 & \textbf{86.68} & \textbf{95.21} & 98.10 & \textbf{96.38} & \textbf{95.09} & $\sim$ 80 min \\
PanDerm (LP) & 97.40 & 95.19 & \textbf{94.50} & 84.94 & 94.36 & \textbf{99.53} & 96.09 & 94.57 & $\sim$ 5 min \\
\rowcolor{gray!10} Performance difference & -0.63 & -2.46 & +0.91 & -1.74 & -0.85 & +1.43 & -0.29 & -0.52 & -75 min \\
\midrule
Modalities & derm & derm & clinic & clinic & clinic & clinic & TBP & & \\
Size & 10k & 12k & 2k & 839 & 19k & 400 & 6k & & \\
\#class & 7 & 9 & 6 & 2 & 23 & 4 & 3 & & \\
\bottomrule
\end{tabular}
\caption{\textbf{Performance comparison of PanDerm (FT) and PanDerm (LP) models across various dermatology datasets}. FT: Fine-Tuning, LP: Linear Probing. Metrics represent accuracy percentages. The best-performing model for each dataset is bolded. The performance difference row shows the change from FT to LP, with positive values indicating LP outperformed FT. Datasets vary in modality, size, and number of classes as shown in the bottom rows. It shows that PanDerm using simple linear probing can perform comparably with expensive full-parameter finetuning. This suggests that PanDerm’s features are already well-suited for diverse downstream multimodal skin-related tasks without requiring further training. All models are trained and evaluated using 4 $\times$ NVIDIA RTX 6000Ada GPUs.}
\label{tab:lp_ft}
\end{table}



\begin{table}[!htb]
\footnotesize
\centering
\renewcommand{\arraystretch}{1.2}
\begin{tabular}{l|l|l|l|c|c|c}
\toprule
Modality & Tasks & Dataset & Metric & ResNet50 & BioMedGPT & PanDerm \\
\midrule
TBP & Melanoma detection & ISIC2024 & AUROC & 0.8023$^{**}$ & 0.6840$^{***}$ & \textbf{0.893} \\
& (2 classes) & & & (0.7327-0.8723) & (0.5670-0.8009) & (0.839-0.940) \\
\midrule
Dermoscopic & Metastasis Prediction & CombinMel & AUROC & 0.9378$^{*}$ & 0.7774$^{***}$ & \textbf{0.964} \\
images & (2 classes) & & & (0.8979-0.9618) & (0.7094-0.8453) & (0.937-0.991) \\
\midrule
Dermoscopic& Skin cancer classification & HAM\_clean & Weighted & 0.8606$^{***}$ & 0.8309$^{***}$ & \textbf{0.926} \\
images& (7 classes) & & F1 & (0.8411-0.8771) & (0.8142-0.8476) & (0.912-0.940) \\
\midrule
Clinical & General(common) skin & DermNet & Weighted & 0.4181$^{***}$ & 0.3086$^{***}$ & \textbf{0.619} \\
images & condition classification & & F1 & (0.4039-0.4323) & (0.2952-0.3220) & (0.603-0.634) \\
& (23 classes) & & & & & \\
\midrule
Clinical& General skin condition & Fitzpatrick1 & Weighted & 0.2578$^{***}$ & 0.1350$^{***}$ & \textbf{0.4817} \\
images& classification & 7K\_clean & F1 & (0.2410-0.2747) & (0.1225-0.1475) & (0.4628-0.5005) \\
& (115 classes) & & & & & \\
\midrule
Dermato- & Tumor classification & PATCH16 & Weighted & 0.8304$^{***}$ & 0.7066$^{***}$ & \textbf{0.903} \\
pathology & (16 classes) & & F1 & (0.8242-0.8365) & (0.6991-0.7142) & (0.898-0.908) \\
\bottomrule
\end{tabular}
\caption{\textcolor{black}{\textbf{Performance comparison across different dermatological tasks and modalities.} Models are evaluated on various dermatological tasks spanning TBP, dermoscopic images, clinical photographs, and dermatopathology. Evaluation metrics include Area Under the Receiver Operating Characteristic curve (AUROC) for binary classification tasks and Weighted F1 score for multi-class classification. Performance is reported with 95\% confidence intervals in parentheses. The best performance for each task is bolded. $^{***}p<0.001$ compared to PanDerm.}}
\label{tab:GMAI}
\end{table}

\begin{table}[!htb]
\footnotesize
\centering
\renewcommand{\arraystretch}{1.2}
\begin{tabular}{l|cc}
\toprule
\textbf{Model} & \textbf{Accuracy} & \textbf{Weighted F1} \\
\midrule
BioMedGPT (finetune) & 0.7394 & 0.7740 \\
BioMedGPT (Linear probe) & 0.7149 & 0.6816 \\
\midrule
PanDerm (Finetune) & \textbf{0.8538} & \textbf{0.8538} \\
PanDerm (Linear probe) & 0.8306 & 0.8252 \\
\bottomrule
\end{tabular}
\caption{\textcolor{black}{\textbf{Performance comparison between BioMedGPT and PanDerm models on HAM10000.} Models are evaluated using both fine-tuning and linear probing (LP) approaches. Performance is reported using Accuracy and Weighted F1 score metrics. Best performance for each metric is bolded.}}
\label{tab:GAMI2}
\end{table}

%% file: tables/model_details.tex
\begin{table}[h]
  \centering
  \begin{tabular}{p{7.5cm}|p{3cm}}
    \toprule
    Hyper-parameter & Value \\
    \midrule
    Teacher model & CLIP \\
    First input size & 224 \\
    Second input size & 196 \\
    Second interpolation & bicubic \\
    Number of output dimensions & 768 \\
    Crop min size & 0.4 \\
    Crop max size & 1 \\
    Patch size & 16 \\
    Vocabulary size & 8000 \\
    Batch size & 480 \\
    Learning rate & 1.5e-3 \\
    Warmup epochs & 20 \\
    Total epochs & 500 \\
    Gradient clipping max norm & 3.0 \\
    Layer scale init value & 1e-5 \\
    Color jitter & 0.4 \\
    Drop path & 0.2 \\
    Mask generator & block \\
    Number of mask patches & 118 \\
    Decoder layer scale init value & 1e-5 \\
    Regressor depth & 4 \\
    Decoder depth & 0 \\
    Decoder embed dimension & 1024 \\
    Decoder number of heads & 16 \\
    Align loss weight & 0 \\
    Latent alignment loss weight & 1 \\
    \midrule
    Number of GPUs & 4 \\
    Distributed launch & torchrun \\
    Processes per node & 4 \\
    \bottomrule
  \end{tabular}
  \caption{\textbf{PanDerm hyperparameters used in pretraining}. 4 $\times$ 80GB NVIDIA H100 GPUs were used for pretraining.}
  \label{tab:pretrain}
\end{table}

\begin{table}[h]
  \centering
  \begin{tabular}{@{}p{4cm}|r@{}}
    \toprule
    Hyperparameter & Value \\
    \midrule
    Batch size & 256 \\
    Epochs & 50 \\
    learning rate & 5e-4 \\
    Layer decay & 0.75 \\
    Weight decay & 0.05 \\
    Drop path & 0.2 \\
    Reprob & 0.25 \\
    Mixup & 0.8 \\
    Cutmix & 1.0 \\
    \bottomrule
  \end{tabular}
  \caption{\textbf{PanDerm hyperparameters used in finetuning}. A single 49GB NVIDIA 6000Ada GPU was used for downstream finetuning.}
  \label{tab:finetune}
\end{table}